\pdfoutput=1

\documentclass{article}
\usepackage[final,main]{neurips_2025}

\usepackage{multirow}
\usepackage{latexsym}

\usepackage{paralist}
\usepackage[utf8]{inputenc}
\usepackage[T1]{fontenc}
\usepackage{textcomp}
\usepackage{amsmath, amssymb, amsthm}
\usepackage{graphicx}
\usepackage[
    colorlinks=true,
    urlcolor=blue,
    linkcolor=.,
    citecolor=.,
    filecolor=.,
    menucolor=.
]{hyperref}
\usepackage{geometry}
\usepackage{caption}
\usepackage{subcaption} %
\usepackage{booktabs}
\usepackage{enumitem}
\usepackage{cleveref}
\usepackage{fancyvrb}
\usepackage{listings} 
\usepackage{xcolor}
\usepackage{soul}
\definecolor{lightyellow}{RGB}{255,255,100}
\sethlcolor{lightyellow}
\newcommand{\yellowhl}[1]{\footnotesize\hl{#1}}
\newcommand{\whitehl}[1]{\footnotesize#1}
\usepackage{etoolbox}
\newcommand{\zerodisplayskips}{%
  \setlength{\abovedisplayskip}{0.5em}%
  \setlength{\belowdisplayskip}{0.2em}}%
\appto{\normalsize}{\zerodisplayskips}
\appto{\small}{\zerodisplayskips}
\appto{\footnotesize}{\zerodisplayskips}

\usepackage{soul}
\usepackage{listings}
\newcommand{\bluelink}[2]{\href{#1}{#2}}
\usepackage{titlesec}
\titlespacing{\paragraph}
{0pt}     
{0ex}     
{0.5em}     

\newcommand{\hlbg}[2]{%
  \setlength{\fboxsep}{0pt}%
  \setlength{\fboxrule}{0pt}%
  \colorbox{#1}{\strut #2}%
}
\lstdefinestyle{colorchars}{
  basicstyle=\sffamily\scriptsize,
  columns=fixed,
  escapechar=|,
  keepspaces=true,
  showstringspaces=false,
  breaklines=true,       
  breakatwhitespace=false,
}
\lstdefinestyle{colorchars_small}{
  basicstyle=\sffamily\scriptsize,
  columns=fixed,
  escapechar=|,
  keepspaces=true,
  showstringspaces=false,
  breaklines=true,       
  breakatwhitespace=false,
  breakindent=1ex,
}

\everypar{\looseness=-1}

\usepackage{microtype}
\usepackage[utf8]{inputenc}
\usepackage[T1]{fontenc}
\usepackage{latexsym}
\usepackage{amsfonts}
\usepackage{amsmath}
\usepackage{amssymb}
\usepackage{amsthm}
\usepackage{relsize}
\usepackage{xspace}
\usepackage[normalem]{ulem}

\usepackage[group-separator={,}]{siunitx}

\usepackage{tikz}
\usetikzlibrary{bayesnet}
\usetikzlibrary{arrows}
\usepackage{color}
\usepackage{graphicx}
\usepackage{caption}
\usepackage{subcaption}
\usetikzlibrary{backgrounds}

\usepackage{hhline}
\usepackage{xcolor}
\usepackage{colortbl}
\definecolor{DarkGreen}{RGB}{3,165,14}

\DeclareMathOperator*{\argmin}{argmin}

\newcommand{\R}{\mathbb{R}}

\def\vb{{\mathbf{b}}}

\def\vd{{\mathbf{d}}}

\def\vf{{\mathbf{f}}}

\def\vh{{\mathbf{h}}}

\def\vy{{\mathbf{y}}}

\def\vepsilon{{\boldsymbol{\varepsilon}\xspace}}

\newcommand{\base}{\text{base}}
\newcommand{\chat}{\text{chat}}
\newcommand{\sshared}{\text{shared}}
\newcommand{\enc}{\text{enc}}
\newcommand{\dec}{\text{dec}}
\newcommand{\hbase}{\mathbf{h}^{\base}}
\newcommand{\hchat}{\mathbf{h}^{\chat}}
\newcommand{\dbase}{\mathbf{d}^{\base}}
\newcommand{\dchat}{\mathbf{d}^{\chat}}
\newcommand{\ebase}{\mathbf{e}^{\base}}
\newcommand{\echat}{\mathbf{e}^{\chat}}
\newcommand{\norm}[1]{\lVert#1\rVert}

\newcommand{\normtwo}[1]{\norm{#1}_2}

\newcommand{\reconhbase}{\widetilde{\mathbf{h}}^{\base}}
\newcommand{\reconhchat}{\widetilde{\mathbf{h}}^{\chat}}
\newcommand{\errorbase}{\vepsilon^{\base}}
\newcommand{\errorchat}{\vepsilon^{\chat}}

\newcommand{\betarbase}{\beta^{r,\base}}
\newcommand{\betaepsbase}{\beta^{\varepsilon,\base}}

\newcommand{\betarchat}{\beta^{r,\chat}}
\newcommand{\betaepschat}{\beta^{\varepsilon,\chat}}

\definecolor{StringMacroColor}{RGB}{0,120,148}

\definecolor{darkergray}{RGB}{80,80,80}
\definecolor{chatonlycolor}{RGB}{0,153,255}
\definecolor{chatspecificcolor}{RGB}{0,0,179}
\newcommand{\chatonly}{{\color{chatonlycolor}\emph{chat-only}}\xspace}
\newcommand{\chatspecific}{{\color{chatspecificcolor}\emph{chat-specific}}\xspace}
\newcommand{\baseonly}{{\color{DarkGreen}\emph{base-only}}\xspace}
\newcommand{\shared}{{\color{orange}\emph{shared}}\xspace}
\newcommand{\other}{{\color{darkergray}\emph{other}}\xspace}

\newcommand{\lmchat}{p^{\chat}}

\newcommand{\chatapprox}{\vh_a}
\newcommand{\kl}{\mathcal{D}}

\newcommand{\jtwin}{j_\text{twin}}

\newcommand{\gtfchat}{f^\chat_C}
\newcommand{\gtfbase}{f^\base_C}
\newcommand{\fshared}{f_\sshared}
\newcommand{\fexclchat}{f_\text{c-excl}}
\newcommand{\fexclbase}{f_\text{b-excl}}

\newcommand{\RND}{\Delta_\text{norm}}
    
\usepackage{algorithm}
\usepackage{algpseudocode}

\newcommand{\batchtopk}{BatchTopK\xspace}
\newcommand{\Lone}{L1\xspace}

\newcommand{\trulychatonly}{chat-specific\xspace}
\newcommand{\TopK}{\textsc{batchtopk}}
\newcommand{\alllatents}{\mathcal{J}}
\newcommand{\loss}{\mathcal{L}}
\newcommand{\x}{x}
\usepackage{tipa}
\newcommand{\paris}{\text{\normalfont \textipa{$\dagger$}}}
\newcommand{\univ}{\text{\normalfont \textipa{\ddag}}}
\newcommand{\epfl}{\normalfont \text{\textipa{@}}}
\newcommand{\ethz}{\normalfont \text{\textipa{A}}}
\newcommand{\noeu}{\normalfont \text{\textipa{D}}}

\usepackage{bbm}
\usepackage[disable]{todonotes}
\usepackage[authormarkup=none]{changes}

\newcommand{\note}[4][]{\todo[author=#2,color=#3,size=\scriptsize,fancyline,caption={},#1]{#4}} %

\definecolor{dandelion}{HTML}{FFD464}

\definecolor{bittersweet}{HTML}{C04F17}

\definecolor{mintgreen}{RGB}{152, 255, 152}

\definecolor{lavendel}{RGB}{230,230,250}

\newcommand{\julian}[2][]{\note[#1]{\textbf{julian}}{green}{#2}}

\definechangesauthor[name={Clement}, color=purple]{clem}

\title{Overcoming Sparsity Artifacts in Crosscoders to Interpret Chat-Tuning
}

\author{
  Julian Minder$^{*\epfl\ethz}$ \quad Cl\'ement Dumas$^{*\paris\univ}$\\ \textbf{Caden Juang$^{\noeu}$\quad Bilal Chughtai \quad Neel Nanda}\\
 $^{\epfl}$EPFL \quad $^{\ethz}$ETHZ \quad 
 $^{\paris}$Ecole Normale Supérieure Paris-Saclay  \quad
 $^{\univ}$Université Paris-Saclay \\
 $^{\noeu}$Northeastern University \quad
\\
\texttt{\small \href{mailto:julian.minder@epfl.ch}{\color{black}julian.minder@epfl.ch}}, \texttt{\small \href{mailto:clement.dumas@ens-paris-saclay.fr}{\color{black}clement.dumas@ens-paris-saclay.fr}}
}
\date{}

\begin{document}
\maketitle

\def\thefootnote{*}\footnotetext{Equal contribution. Order randomized.}\def\thefootnote{\arabic{footnote}}

\begin{abstract}
Model diffing is the study of how fine-tuning changes a model's representations and internal algorithms. 
Many behaviors of interest are introduced during fine-tuning, and model diffing offers a promising lens to interpret such behaviors. 
Crosscoders are a recent model diffing method that learns a shared dictionary of interpretable concepts represented as latent directions in both the base and fine-tuned models, allowing us to track how concepts shift or emerge during fine-tuning. 
Notably, prior work has observed concepts with no direction in the base model, and it was hypothesized that these model-specific latents were concepts introduced during fine-tuning.
However, we identify two issues which stem from the crosscoders L1 training loss that can misattribute concepts as unique to the fine-tuned model, when they really exist in both models. 
We develop Latent Scaling to flag these issues by more accurately measuring each latent's presence across models.
In experiments comparing Gemma 2 2B base and chat models, we observe that the standard crosscoder suffers heavily from these issues. 
Building on these insights, we train a crosscoder with BatchTopK loss and show that it substantially mitigates these issues, finding more genuinely chat-specific and highly interpretable concepts. We recommend practitioners adopt similar techniques.
Using the BatchTopK crosscoder, we successfully identify a set of chat-specific latents that are both interpretable and causally effective, representing concepts such as \emph{false information} and \emph{personal question}, along with multiple refusal-related latents that show nuanced preferences for different refusal triggers. 
Overall, our work advances best practices for the crosscoder-based methodology for model diffing and demonstrates that it can provide concrete insights into how chat-tuning modifies model behavior.
\footnote{We open-source our \bluelink{https://github.com/jkminder/science-of-finetuning/}{code}, \bluelink{https://github.com/jkminder/dictionary_learning}{training library}, \bluelink{https://huggingface.co/science-of-finetuning}{models}, \bluelink{https://wandb.ai/jkminder/chat-crosscoders}{wandb runs} and a \bluelink{https://dub.sh/ccdm}{demo notebook to explore latents}.}
\end{abstract}

\section{Introduction}
\label{sec:introduction}

Classically, mechanistic interpretability  \citep{sharkey2025openproblemsmechanisticinterpretability,mueller2024questrightmediatorhistory,ferrando2024primerinnerworkingstransformerbased,elhage2021mathematical,olah2020zoom} aims to reverse engineer an entire model \citep{huben2024sparse,elhage2022superposition}, or \emph{circuits} implemented by the model to solve particular tasks \citep{wang2023interpretability}. \emph{Model diffing}  offers an alternative method by focusing on \emph{changes} induced by fine-tuning. Since fine-tuning typically involves far less compute than the pre-training phase that establishes general knowledge and generic circuitry, its resulting modifications are expected to be limited in scope. This targeted nature suggests model diffing could be a \textit{more tractable} approach to mechanistic interpretability than the full model analysis, while still providing valuable insights into core features of a model's behavior.

Model diffing might indeed be incredibly useful. The process of fine-tuning a model is what makes it \emph{useful} as a tool or agent. Better understanding the mechanisms that give reasoning models \citep{deepseekai2025deepseekr1incentivizingreasoningcapability, openai2024openaio1card} heightened capabilities as compared to base or chat models might allow us to debug their failures and improve them. Fine-tuning also often introduces a number of problematic behaviors, for example, sycophancy \citep{sharma2023understandingsycophancylanguagemodels}. 
Future AI safety and alignment concerns \citep{greenblatt2024alignmentfakinglargelanguage, meinke2025frontiermodelscapableincontext, betley2025emergent} may emerge specifically in fine-tuned models. For example, long-horizon RL could incentivize models to exploit reward signals and act deceptively. Model diffing could allow us to detect this.

Prior model diffing research has investigated how models change during fine-tuning \citep{shah2023modeldiff, lindsey2024sparse,bricken2024stagewise,prakash2024finetuning, lee2024mechanistic, jain2024mechanistically,khayatan2025analyzingfinetuningrepresentationshift,thasarathan2025universalsparseautoencodersinterpretable,wu-etal-2024-language,mosbach-2023-analyzing,merchant-etal-2020-happens, hao-etal-2020-investigating,kovaleva-etal-2019-revealing, du2025posttrainingmechinterp, minder2024understanding}. While these studies have hypothesized that fine-tuning primarily shifts and repurposes existing capabilities rather than developing new ones, conclusive evidence for this claim remains elusive.
Model diffing remains a nascent field that lacks established consensus and mature analytical tools. Much prior work has leveraged ad-hoc techniques for understanding how models change in narrow ways (e.g. focusing on a particular circuit), or have been on toy model. 
It is unclear whether prior approaches would scale to understanding the kinds of fine-tuning large models actually undergo.

Recently, \citet{lindsey2024sparse} introduced the \textbf{crosscoder}, a novel and scalable tool for model diffing. Crosscoders build on the popular sparse autoencoder (SAE) \citep{huben2024sparse, bricken2023monosemanticity,yun-etal-2021-transformer}, which has shown promise for interpreting a model's representations by decomposing activations into a sum of sparsely activating dictionary elements. There are many variants of crosscoders; the variant we are concerned with in this paper concatenates the activations of the base and chat-tuned model residual streams and trains a shared dictionary across this activation stack. Thus, for each dictionary element (aka "latent", corresponding to one concept), the crosscoder learns a pair of latent directions - one corresponding to the base model and one to the chat-tuned model. Crosscoders can thus potentially identify which latents are novel to the fine-tuned model, which are novel to the base-model, and which are shared. We term these sets chat-only, base-only, and shared respectively. \citet{lindsey2024sparse} identify chat-only latents by looking at the norm of the latent directions -- if the latent direction of the base model has zero norm, this indicates that the latent is chat-only.

\vspace{1em} %
In this work, we critically examine the crosscoder and identify two theoretical limitations of its training objective, that may lead to falsely identified chat-only latents (\Cref{sec:decnormbaseddiffing}):
\begin{compactenum}
        \item Complete Shrinkage: The sparsity loss can force base latent directions to zero norm, even when they contribute to base model reconstruction.
        \item Latent Decoupling: The crosscoder may represent a shared concept using a chat-only latent when it is actually encoded by a different combination of latents in the base model, as the crosscoder's sparsity loss treats both representations as equivalent.
\end{compactenum}
We develop an approach called \emph{Latent Scaling} to detect spurious chat-only latents, inspired by \citeauthor{wright2024addressing}'s \citeyearpar{wright2024addressing}~SAE scaling (\Cref{sec:latentscaling}), and demonstrate that the above issues occur in practice. While the norm-based metric from \citet{lindsey2024sparse} appears to identify a clean trimodal distribution of base-only, shared and chat-only latents, we show that this is an artifact of the loss function rather than a meaningful distinction. Our conclusion is that the crosscoder loss does not actually have an inductive bias that helps to learn better model-only latents. Nonetheless, we demonstrate that crosscoders trained with \batchtopk loss \citep{bussmann2024batchtopk} exhibit robustness to the above issues (\Cref{sec:demonstratingissues}) and identify a larger number of genuine model-specific latents. We show that in the \batchtopk crosscoder, the norm-based metric successfully identifies causally relevant latents by measuring their ability to reduce the prediction gap between base and chat model. In contrast, this metric fails in the L1 crosscoder, where Latent Scaling becomes necessary to identify the truly causally relevant latents. Finally, we outline that the chat-only latents found by the \batchtopk crosscoder are highly interpretable (\Cref{sec:observationsrefinedlatents}), revealing key aspects of chat model behavior such as the role of chat template tokens, persona-related questions, detection of false information, and various refusal related mechanisms.

Overall, we show that using BatchTopK loss overcomes the described limitations of L1-trained crosscoders, validating them as a useful tool for understanding fine-tuning effects in large language models.

\section{Methods}
\label{sec:methods}

\noindent\textit{Note: For reference, we provide a comprehensive glossary of key terms and mathematical notation introduced through the paper in Appendix~\ref{app:glossary}.}

\subsection{Crosscoder architectures}

To build intuition, the crosscoder's goal is to learn a dictionary of interpretable concepts (latents) that can explain the activations of both models. It consists of an encoder and a decoder. The encoder takes the activations of the base and chat models and projects them into a shared high-dimensional sparse space, where each dimension corresponds to a potential concept. The decoder then reconstructs each model's activations using model-specific representations for each latent, combining them according to the sparse encoding. The key insight is that while both models share the same sparse encoding for a given input, the crosscoder learns separate decoder representations for each model, allowing concepts to have different importance or manifestation in each model.

More formally, let $x$ be a string and $\hbase(x), \hchat(x) \in \R^{d}$ denote the activations at a given layer. The encoder computes a sparse encoding $f_j(x) \in \R_{\geq 0}$ for each latent $j \in \alllatents=\{1, \dots, D\}$. The decoder then reconstructs the activations as:
\begin{align}
  \reconhbase(x) &= \sum_j f_j(x)\,\dbase_{j}+\vb^{\dec, \text{base}} \quad \text{and} \quad
  \reconhchat(x) = \sum_j f_j(x)\,\dchat_{j} + \vb^{\dec, \text{chat}}
\end{align}
where \smash{$\dbase_j, \dchat_j \in \R^d$} are the model-specific decoder representations and \smash{$\vb^{\dec,\text{base}}, \vb^{\dec,\text{chat}} \in \R^d$} are decoder biases. The crosscoder minimizes reconstruction errors \smash{$\errorbase(x) = \hbase(x) - \reconhbase(x)$} and \smash{$\errorchat(x) = \hchat(x) - \reconhchat(x)$} while enforcing sparsity.

We examine two sparsity mechanisms. The \Lone crosscoder \citep{lindsey2024sparse} adds an L1 penalty to the loss:
\begin{align}
  \loss_{\text{\Lone}}(x) &= f_j(x)\left(\normtwo{\dbase_j} + \normtwo{\dchat_j}\right)
\label{eq:loss}
\end{align}

The \batchtopk crosscoder \citep{bussmann2024batchtopk} instead enforces L0 sparsity by selecting only the top $nk$ latents with highest scaled activation $f_j(x_i)(\normtwo{\dbase_j} + \normtwo{\dchat_j})$ across a batch of $n$ strings.\footnote{During inference, a learned threshold $\theta$ zeroes out latents below it. See \Cref{eq:threshold}.} More details on crosscoder implementation can be found in \Cref{app:cc_definitions}.

\subsection{Decoder norm based model diffing and its problems}
\label{sec:decnormbaseddiffing}
\begin{figure}[t!]
  \centering
  \begin{subfigure}[b]{0.48\textwidth}
      \centering
      \includegraphics[width=\textwidth]{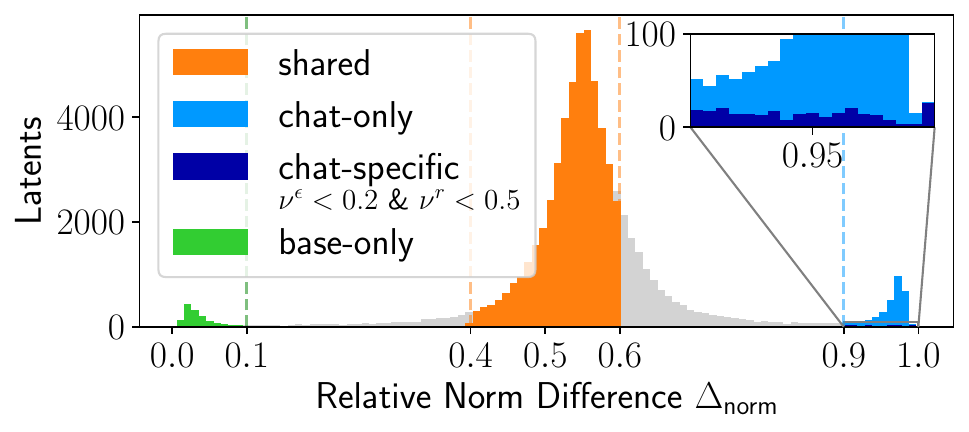}
      \caption{\Lone crosscoder.}
  \end{subfigure}
  \hfill
  \begin{subfigure}[b]{0.48\textwidth}
      \centering
      \includegraphics[width=\textwidth]{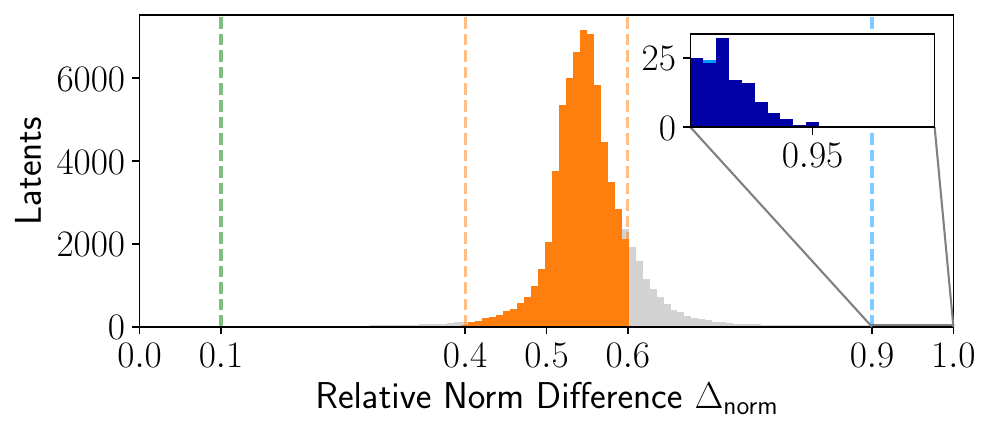}
      \caption{\batchtopk crosscoder.}
  \end{subfigure}
  \caption[]{Histogram of decoder latent relative norm differences ($\Delta_\text{norm}$) between base and chat Gemma 2 2B models \citep{team2024gemma}, for both the \Lone crosscoder (left) and the \batchtopk crosscoder (right). A value of $1$ means the decoder vector of a latent for the base model is zero, indicating the latent is not useful for the base model (\chatonly latents). A value of $0$ means the chat model's decoder vector has a norm of zero (\baseonly latents). Values around $0.5$ indicate similar decoder norms in both models, suggesting equal utility in both models (\shared latents)\footnotemark. We also show the \chatonly latents that are truly \trulychatonly and that are not affected by Complete Shrinkage (error ratio $\nu^\varepsilon < 0.2$) and Latent Decoupling (reconstruction ratio $\nu^r < 0.5$) -- the \chatspecific latents. Most of the \Lone crosscoder \chatonly latents suffer from these issues.}
  \label{fig:decoder_norm}
\end{figure}
To leverage crosscoders for model diffing, we can exploit the observation that while latent activations $f_j(x)$ are shared between models, the decoder vectors $\mathbf{d}^{\chat}_j$ and $\mathbf{d}^{\base}_j$ are unique to each model.

To leverage crosscoders for model diffing, we exploit that while the sparse encoding $f_j(x)$ is shared between models, the decoder representations $\mathbf{d}^{\chat}_j$ and $\mathbf{d}^{\base}_j$ are model-specific. When a latent is important for both models, both decoder representations need substantial norms for reconstruction. Conversely, a latent specific to the chat model will have $\normtwo{\mathbf{d}^{\chat}_j} \gg 0$ while $\normtwo{\mathbf{d}^{\base}_j} \to 0$, as the base decoder has no use for this latent.

We quantify this using the relative norm difference $\Delta_{\text{norm}}: \alllatents \rightarrow [0,1]$ from \citep{lindsey2024sparse}:
\begin{equation}\label{eq:normdiff}
    \Delta_{\text{norm}}(j) = \frac{1}{2}\left(1 + \frac{\normtwo{\mathbf{d}^{\chat}_j} - \normtwo{\mathbf{d}^{\base}_j}}{\max(\normtwo{\mathbf{d}^{\chat}_j}, \normtwo{\mathbf{d}^{\base}_j})}\right)
\end{equation}
Intuitively, $\Delta_{\text{norm}} = 1$ indicates a pure chat-only latent (base decoder has zero norm), $\Delta_{\text{norm}} = 0$ indicates a pure base-only latent, and $\Delta_{\text{norm}} \approx 0.5$ suggests equal importance in both models. As shown in \Cref{fig:decoder_norm}, we classify latents as \baseonly ($0$–$0.1$), \chatonly ($0.9$-$1.0$), or \shared ($0.4$-$0.6$).

\paragraph{Are \chatonly latents really \trulychatonly?}
\label{sec:are_chatonly_latents_really_chat_only}
If a latent only contributes to one model, the norm of the decoder must tend to zero for the other model. But is the converse true? Specifically, we ask the question: if a latent has decoder norm zero in the base model, is it necessarily \trulychatonly? We focus on the \chatonly set, as it will contain features that emerged during chat-tuning.

\paragraph{Reasons to doubt \chatonly latents.}
\label{sec:issues}
\footnotetext{We observe larger activation norms in the chat model, which shifts our distribution rightward, revealing that the chat model amplifies the norm of representations shared with the base model.}

There are reasons to suspect \chatonly latents might not be \trulychatonly. Firstly, both qualitative and quantitative analysis of \Lone crosscoder latents reveals a relatively low percentage of interpretable latents within the \chatonly set (See \Cref{sec:observationsrefinedlatents}). More worryingly, inspection of the \Lone crosscoder loss (\Cref{eq:loss}) uncovers two theoretical issues that could result in latents $j$, which are defined by their decoder vectors $\vd_j$ and activation function $f_j$, being classified as \chatonly, despite their presence in the activations of the base model:
\begin{compactenum}
\item \textbf{Complete Shrinkage}: When the contribution of latent $j$ is smaller in the base model than in the chat model, L1 regularization can force $\dbase_j$ to zero despite its presence in the base activation. Consequently, \uline{\smash{$\errorbase$} contains information attributable to latent $j$}. This is similar to ``shrinkage'' or ``feature suppression'' in SAEs \citep{jermyn2024tanh, wright2024addressing, rajamanoharan2024improving}.
\item \textbf{Latent Decoupling}: a \chatonly latent $j$ is also present in the base activations but is reconstructed by other base decoder latents. In this case, \uline{the base reconstruction \smash{$\reconhbase$} contains information that could be attributed to latent $j$.} See \Cref{sec:latent_decoupling_example} for an illustrative example.
\end{compactenum}

\paragraph{Why \batchtopk crosscoders might fix this.} The \batchtopk crosscoder may address both Complete Shrinkage and Latent Decoupling issues that affect the \Lone crosscoder. The key difference lies in their respective loss functions and optimization objectives.

For the \Lone crosscoder, the loss function in \Cref{eq:loss} includes an L1 regularization term that directly penalizes the norm of decoder vectors. This creates pressure to shrink decoder norms toward zero when a latent's contribution is minimal, potentially causing Complete Shrinkage even when the latent has some explanatory power. In contrast, the \batchtopk crosscoder uses a different sparsity mechanism. Rather than penalizing all decoder norms, it selects only the top $k$ most active latents per sample during training. This approach has two important advantages:
\begin{compactenum}
    \item No direct norm penalty: Without explicit regularization on decoder norms, there's no optimization pressure to drive $\normtwo{\dbase_j}$ to zero when the latent has explanatory value for the base model, reducing Complete Shrinkage.
    \item Competition between latents: The top-$k$ selection creates competition among latents, discouraging redundant representations. This helps prevent Latent Decoupling by making it inefficient to maintain duplicate latents that encode the same information.
\end{compactenum}

The \batchtopk approach thus creates an inductive bias toward learning more genuinely chat-specific latents, as the model must efficiently allocate its limited "budget" of $k$ active latents. This should result in fewer falsely identified \chatonly latents and a cleaner separation between truly model-specific and shared features. 

\subsection{Latent Scaling: Identifying Complete Shrinkage and Latent Decoupling}
\label{sec:latentscaling}
To empirically investigate whether Complete Shrinkage and Latent Decoupling occur, we introduce \emph{Latent Scaling}, which measures how well a supposedly \chatonly latent can explain base model activations. We achieve this by finding the optimal scale for latent $j$ to best reconstruct the base activations:
\begin{equation}\label{eq:linear_regression}
  \beta^\base_j = \argmin_{\beta} \sum_{i=1}^n\normtwo{\beta f_j(x_i)\dchat_j - \hbase(x_i)}^2 
\end{equation}
This least squares problem has an efficient closed-form solution\footnote{The closed-form solution is derived in \Cref{sec:closed_form_solution} which also gives some intuition on the optimal $\beta$.}. For a \trulychatonly latent, we would expect $\beta^{\base}_j \approx 0$ as the latent shouldn't help explain base activations at all.
However, due to superposition \citep{elhage2022superposition}, even genuinely chat-specific latents might correlate with other features, resulting in $\beta^{\base}_j >0$. To account for this, we measure chat specificity using a ratio that compares how well the latent explains each model $\nu_j = {\beta^{\base}_j} / {\beta^{\chat}_j}$
where $\beta^{\chat}_j$ is computed analogously using $\hchat(\cdot)$ instead of $\hbase(\cdot)$. A value near zero indicates a \trulychatonly latent, while a value near one suggests the latent is equally present in both models.

While this ratio efficiently identifies spurious \chatonly latents, it doesn't tell us \emph{why} they're spurious: it conflates Complete Shrinkage and Latent Decoupling. To distinguish between these failure modes, we leverage the fact that the crosscoder decomposes base activations $\hbase$ into its reconstruction ($\reconhbase$) and what it fails to reconstruct ($\errorbase$):
\begin{compactenum}
    \item If Complete Shrinkage occurred, the latent's information should appear in the reconstruction error $\errorbase$, because the latent's base decoder is shrunk to zero instead of reconstructing the activation. This is captured by the error ratio $\nu_j^\varepsilon = \betaepsbase_j / \betaepschat_j$.
    \item If Latent Decoupling occurred, the latent's information should appear in the reconstruction $\reconhbase$, having been captured by other base model latents. This is measured by the reconstruction ratio $\nu_j^r=\betarbase_j/ \betarchat_j$.
\end{compactenum}

These additional $\beta$ values are computed using the same approach as Equation \ref{eq:linear_regression}, but replacing $\hbase$ with either the error or reconstruction terms \footnote{See \Cref{sec:setup_for_latent_scaling} for exact implementation \Cref{sec:additional_analysis_for_latent_scaling} for verification of correlation between $\nu$ values and actual reconstruction improvement.}.

\section{Results}\label{sec:results}

    
We replicate the model diffing experiments by \citet{lindsey2024sparse} using the open-source Gemma-2-2b (base) and Gemma-2-2b-it (chat) models \citep{team2024gemma}.
We train \Lone and \batchtopk crosscoders on the middle layer (13) activations of both models\footnote{We chose the middle layer as it's where we expect to find the richest representations \citep{skean2025layer}.}, collected on a mixture of both web and chat data. To ensure a fair comparison, we choose hyperparameters for both crosscoders to reach an L0 of 100.
For details on the training process, see \Cref{sec:training_details}.

In \Cref{fig:decoder_norm}, we present the histogram of $\RND$ between base and chat for both the \Lone and \batchtopk crosscoders. At first glance, the \Lone crosscoder identifies substantially more \chatonly latents than the \batchtopk crosscoder. However, our subsequent analysis reveals that many of these apparent \chatonly latents are artifacts of the \Lone loss rather than genuinely \trulychatonly features. 
Refer to \Cref{sec:moreccdetails} for additional empirical details on the crosscoders.

\subsection{Demonstrating Complete Shrinkage and Latent Decoupling}
\label{sec:demonstratingissues}

\begin{figure*}[t]
  \centering
  \begin{subfigure}[t]{0.33\textwidth}
      \centering
      \includegraphics[width=\textwidth]{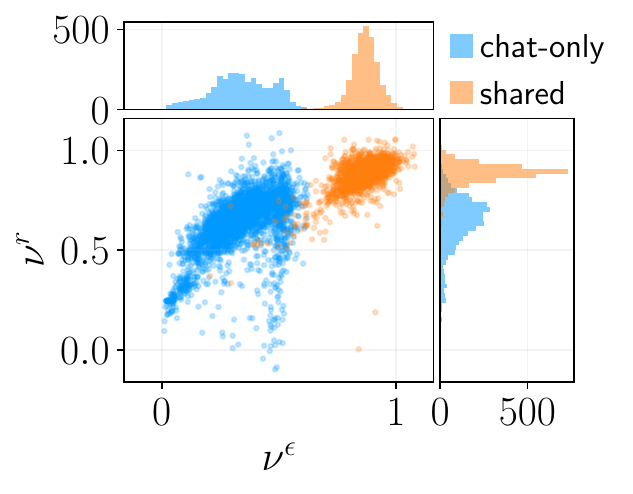}
      \caption{\Lone crosscoder}
      \label{fig:ratios_lone}
  \end{subfigure}
  \begin{subfigure}[t]{0.33\textwidth}
      \centering
      \includegraphics[width=\textwidth]{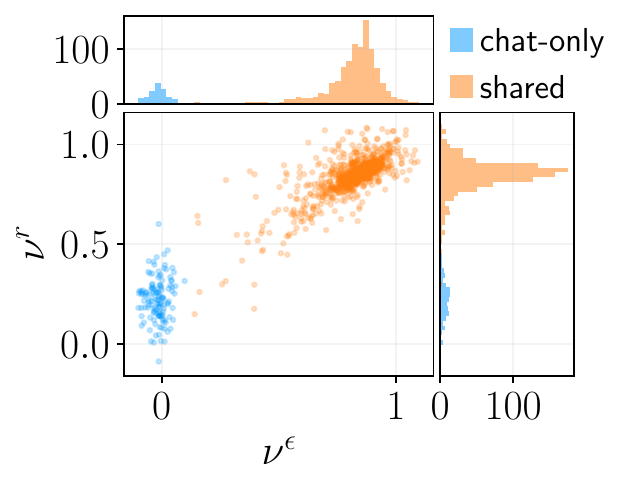}
      \caption{\batchtopk crosscoder}
      \label{fig:ratios_topk}
  \end{subfigure}
  \begin{subfigure}[t]{0.32\textwidth}
      \centering
      \includegraphics[width=\textwidth]{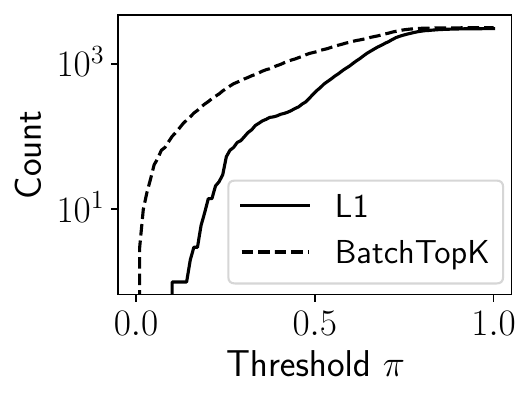}
      \caption{Number of latents ($y$-axis) for which $\nu^r < \pi$ and $\nu^\varepsilon < \pi$.}
      \label{fig:ratios_threshold}
  \end{subfigure}
  
  \caption{We compare how \chatonly latents are affected by the issues described in \Cref{sec:issues}. Left/Middle: error and reconstruction ratio distributions for \Lone and \batchtopk crosscoders, with each point representing a single latent.
  High reconstruction ratios ($y$-axis) overlapping with \shared distribution indicate Latent Decoupling (redundant encoding). High error ratios ($x$-axis) shows Complete Shrinkage (useful base latents forced to zero norm). Low values on both metrics (bottom left) identify truly chat-specific latents. \Lone shows many misidentified \chatonly latents while \batchtopk shows minimal issues. This means the $\RND$ successfully identifies chat-specific latents for $\batchtopk$ but fails for \Lone. 
  Right: Count of latents below a range of $\nu$ thresholds ($x$-axis), comparing 3176 \Lone~\chatonly latents versus top-3176 \batchtopk latents sorted by $\RND$.}
  \label{fig:ratios}
\end{figure*}

\paragraph{Analysing the \Lone crosscoder.} We compute the reconstruction and error ratios ($\nu^r_j$ and $\nu^\varepsilon_j$), for all \Lone crosscoder \chatonly latents on 50M tokens from the training set. For calibration, we examine these ratios on a sample of \shared latents, expecting high values for both ratios.
\Cref{fig:ratios_lone} shows significant overlap between reconstruction ratios distributions of \chatonly and \shared latents, suggesting many supposedly chat-specific latents are actually encoded by the base decoder, indicating potential Latent Decoupling. We find further evidence of Latent Decoupling by analyzing (\chatonly, \baseonly) latent pairs with a cosine similarity of 1 in \Cref{app:cosim_coupled_latents}.
We also observe high error ratios for \chatonly latents (up to $\approx 0.5$), indicating substantial Complete Shrinkage. Similar effects appear in independently trained \Lone crosscoders from \citet{kissane_open_2024} (\Cref{sec:connor}).

\paragraph{Comparing \Lone and \batchtopk crosscoders.} Looking at the ratios for the \batchtopk crosscoder reveals a stark contrast (\Cref{fig:ratios_topk}): \chatonly latents show no $\nu^r_j$ overlap with \shared latents, and $\nu^\varepsilon_j$ values are nearly zero, indicating minimal Complete Shrinkage and Latent Decoupling. In \Cref{fig:decoder_norm}, we find that most \Lone crosscoder \chatonly latents are not truly \chatspecific (defined as $\nu^r < 0.5$ and $\nu^\varepsilon < 0.2$), while most \batchtopk~\chatonly latents are genuinely \chatspecific.
To compare the absolute number of chat-specific latents in both crosscoders, we choose the same number of top $\Delta_{norm}$ latents from both models and compare for how many of them both ratios $\nu^r_j$ and $\nu^\epsilon_j$ lie below a range of thresholds $\pi$. Specifically, we compare the 3176 chat-only latents from the L1 crosscoder with the top-3176 latents based on $\RND$ values from the BatchTopK crosscoder.
\Cref{fig:ratios_threshold} shows that for any threshold $\pi$, the \batchtopk crosscoder consistently identifies more \trulychatonly latents (where $\nu^r < \pi$ and $\nu^\varepsilon < \pi$) than the \Lone crosscoder. 
Furthermore, in the \batchtopk crosscoder the $\RND$ and $\nu$ metrics show strong pearson correlation ($\nu^r: 0.73$, $\nu^\epsilon: 0.87$, $p<0.01$) showing that the $\RND$ metric is a 
valid proxy for chat-specificity here. We observe similar effects in both chat models from the Llama 3 family \citep[\Cref{app:llama}]{grattafiori2024llama3herdmodels} and models fine-tuned with RL for reasoning and medical knowledge in \citep[\Cref{app:domainft}]{sallinen2025llamameditron, liu2025prorl}.

\subsection{Measuring the causality of chat approximations}
\label{sec:causality}

\begin{figure}[t!]
    \centering
    \includegraphics[width=1\linewidth]{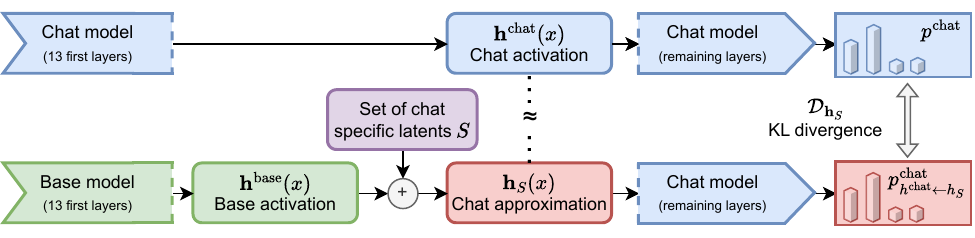}
    \caption{Simplified illustration of our experimental setup for measuring latent causal importance. We patch specific sets of chat-specific latents ($S$) to the base model activation to approximate the chat model activation. The resulting approximation is then passed through the remaining layers of the chat model. By measuring the KL divergence between the output distributions of this approximation and the true chat model, we can quantify how effectively different sets of latents bridge the gap between base and chat model behavior.}
    \label{fig:kl_diagram}
\end{figure}

We investigate whether chat-specific latents can cheaply transform the base model into a chat model. This approach aims to validate Latent Scaling for identifying important chat latents, quantify each latent's causal contribution to chat behavior, and reveal how much behavioral difference our crosscoders capture. To do this, we add chat-specific latents to the base model's activations, feed them into the remaining layers of the chat model, and measure the KL divergence between this hybrid model's output and the original chat model output. A high-level diagram of this method is shown in \Cref{fig:kl_diagram}.

Formally, let $\lmchat$ be the chat model's next-token probability distribution given context $x$, with $\hchat(x)$ and $\hbase(x)$ as the chat and base model activations, respectively. We evaluate an approximation \smash{$\chatapprox(x)$} of \smash{$\hchat(x)$}, by replacing \smash{$\hchat(x)$} with \smash{$\chatapprox(x)$} in the chat model's forward pass, yielding a modified distribution \smash{$\lmchat_{\hchat \leftarrow \chatapprox}$}. The KL divergence, \smash{$\kl_{\chatapprox} = \text{KL}(\lmchat_{\hchat \leftarrow \chatapprox} || \lmchat)$}, then quantifies the predictive power lost by this approximation. Specifically, for a set $S$ of latents, our \smash{$\chatapprox(x)$} is formed by adding the chat decoder's contributions for these latents to the base activation \smash{$\hbase(x)$}.
\begin{align}
  \vh_S(x) =~ & \hbase(x) + \sum_{j \in S} f_j(x) \dchat_j(x)
\end{align}
Let $S$ and $T$ be two disjoint sets of latents. If the KL divergence $\kl_{\vh_S}$ is lower than $\kl_{\vh_T}$, we can conclude that the set $S$ is more important for the chat-model behavior than the set $T$.

Before looking at specific sets, we analyze the following baselines to compare the ability of both architecture at capturing the behavioral difference:
\begin{compactenum}
   \item \textbf{Base activation} (\emph{None}): Intervening with $\hbase(x)$ (i.e., $S=\emptyset$), expected to yield the highest KL divergence.
    \item \textbf{Full Replacement} (\emph{All}): Intervening with all latents ($S=\text{all}$), this represents the best performance achievable by the crosscoder's latent representations and is equivalent to $\vh_\text{all} = \reconhchat(x) + \errorbase(x)$.
    \item \textbf{Error Replacement} (\emph{Error}): using $\vh_{\text{error}} = \reconhbase(x) + \errorchat(x)$ to assess behavioral difference captured by reconstruction error, quantifying chat behavior driven by information missing from the crosscoder's chat activation reconstruction $\reconhchat(x)$.
\end{compactenum}

Then, to validate whether norm difference $\RND$ and Latent Scaling identify causally important latents, we compare interventions using latents ranked highest versus lowest in chat-specificity by each method\footnote{For Latent Scaling, latents are ranked by the sum of their ranks in the error and reconstruction ratios distributions, with lower sums indicating minimal Complete Shrinkage and Latent Decoupling effects.}. We compare the $3176$ \chatonly latents from the \Lone crosscoder with the $3176$ highest-$\RND$ latents from the \batchtopk crosscoder; this matched sample size ensures a fair comparison. For both crosscoders and both ranking methods, we compute KL divergence for interventions using the top 50\% ($S_\text{best}$) and bottom 50\% ($S_\text{worst}$) of these ranked latents, expecting \smash{$\kl_{\vh_{S_\text{best}}} < \kl_{\vh_{S_\text{worst}}}$} as more chat-specific latent should encode more of the behavioral difference.

\begin{figure*}[t!]
  \centering
  \begin{subfigure}[b]{0.494\textwidth}
    \centering
    \includegraphics[width=\textwidth]{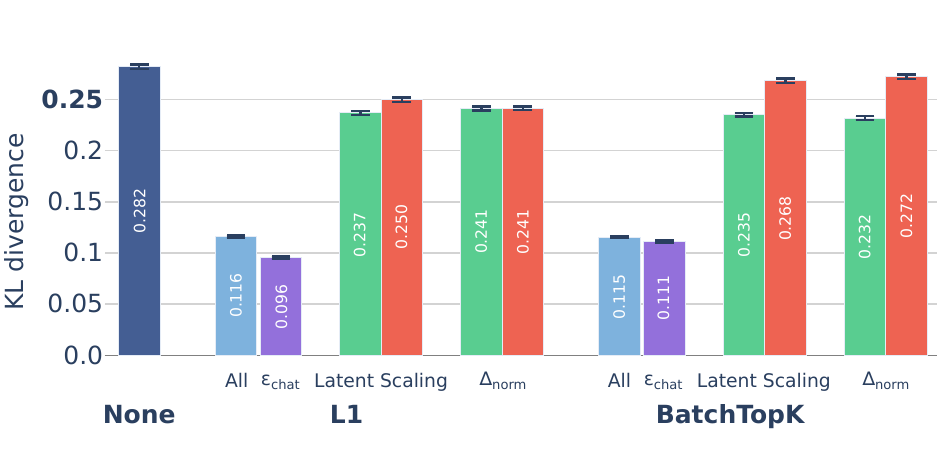}
    \caption{Over full responses.}
    \label{fig:kl_divergence_all}
  \end{subfigure}
  \hfill
  \begin{subfigure}[b]{0.494\textwidth}
    \centering
    \includegraphics[width=\textwidth]{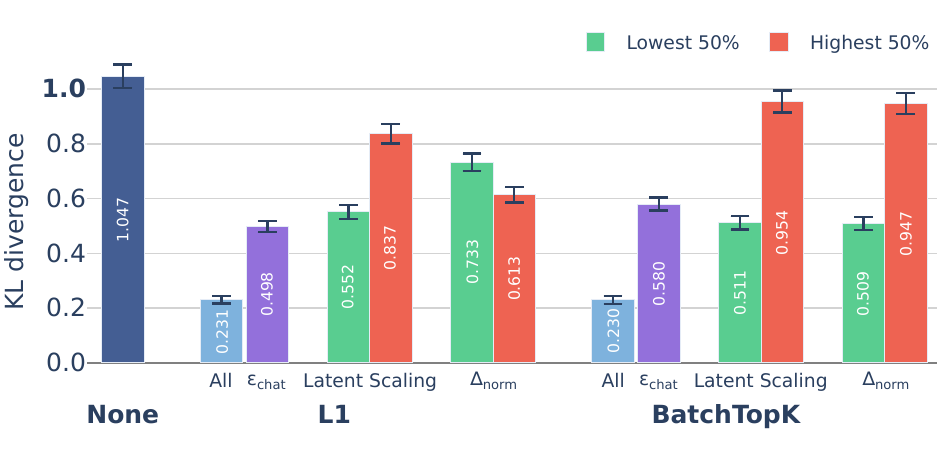}
    \caption{Over first 9 tokens.}
    \label{fig:kl_divergence_first}
  \end{subfigure}
  \caption{Comparison of KL divergence between different approximations of chat model activations. Note the different $y$-axis scales - KL is generally much higher on the first 9 tokens. We establish baselines by replacing either \emph{None} or \emph{All} of the latents. We then evaluate the Latent Scaling metric against the relative norm difference ($\RND$) by comparing the effects of replacing the highest 50\% (red) versus lowest 50\% (green) of latents ranked by each metric. We show the 95\% confidence intervals for all measurements. \textbf{Our results reveal a critical difference between the crosscoders}: while $\RND$ fails to identify causally important latents in the \Lone crosscoder, where lower $\RND$ leads to smaller KL improvement, it successfully does so in the \batchtopk crosscoder. This confirms our hypothesis that $\RND$ is a meaningful metric in \batchtopk but merely a training artifact in \Lone. Using \emph{Latent Scaling}, we successfully identify the most causal latents in \Lone, which is particularly evident in the first 9 tokens (right) where it almost matches \batchtopk. This shows that both crosscoder capture the behavioral difference similarly, \batchtopk avoids $\RND$ artifacts.}
  \label{fig:kl_divergence_comparison}
\end{figure*}

In \Cref{fig:kl_divergence_comparison}, we plot the KL divergence for different experiments on $512$ chat interactions, with user requests from \citeauthor{ding2023enhancing}'s \citep{ding2023enhancing} dataset and responses generated by the chat model\footnote{We report results on LMSYS \citep{zheng2024lmsyschat1mlargescalerealworldllm} in \Cref{sec:causality_experiments_on_lmsys_chat}, observing the same trends.}. We report mean results over both the full responses and first 9 response tokens \footnote{We actually excluded the very first token (token 1) of each response from our analysis to ensure fair comparison with the \emph{template} intervention, introduced later in the paper. The KL is therefore computed on tokens (2-10) rather than (1-9).}. First, we confirm a key finding from \citet{qi2024safetyalignmentjusttokens}: the distributional differences between base and chat models are significantly more pronounced in the initial completion tokens than across the full response. 
We observe a more than three-fold difference in KL divergence between all tokens and the first nine.

When applying the full replacement intervention (\emph{All}), we observe that both crosscoders achieve almost identical KL divergence reductions -- 59\% over all tokens and 78\% for the first 9 tokens compared to the \emph{None} baseline.
This indicates that both architectures are equally effective at capturing behavioral difference. However, the error replacement intervention (\emph{Error}) reveals that this captured difference is far from complete. For full responses, the chat error term achieves slightly better KL reduction than using the chat reconstruction for both crosscoders, indicating that reconstruction error contains at least as much behavioral information as the learned dictionary. This aligns with previous findings by \citet{engels2024decomposingdarkmattersparse} that highlighted the causal importance of the reconstruction error in SAEs. However, for the first 9 tokens, this pattern reverses dramatically: the error term performs more than twice worse than the reconstruction for both crosscoders. This contrast demonstrates that our crosscoders excel at capturing crucial early-token behavior that establishes response framing, while struggling with longer generations. 

\textbf{Despite capturing similar information, the two architectures organize it fundamentally differently}. For the \batchtopk crosscoder, $\RND$ successfully identifies causally important latents: the top 50\% by $\RND$ achieve substantially lower KL divergence than the bottom 50\% (50\% vs 6\% reduction for first 9 tokens). This validates $\RND$ as a reliable proxy for chat-specificity in \batchtopk.
In contrast, $\RND$ fails completely for the \Lone crosscoder—latents with highest $\RND$ latents performing nearly identically or worse than low-$\RND$ latents. This confirms our hypothesis that in \Lone a lot of \chatonly latents are artifacts not capturing the behavioral difference.
However, Latent Scaling successfully identifies causally important latents in the \Lone crosscoder, nearly matching \batchtopk's performance, demonstrating that a subset of \Lone~\chatonly are relevant to the behavioral difference and are identified by latent scaling.

\subsection{Observations about \batchtopk chat-only latents}
\label{sec:observationsrefinedlatents}

\begin{figure}[t]
  \begin{minipage}[t!]{0.48\textwidth}
    \centering
    \vspace{-0.5em}
    \includegraphics[width=\textwidth]{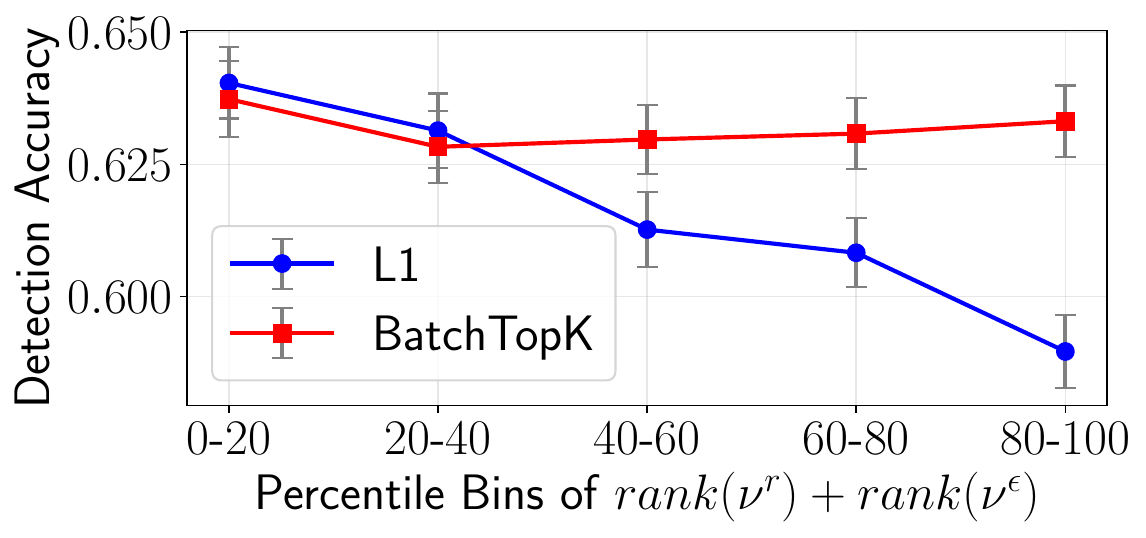}
    \caption{Autointerpretability detection scores (higher is better) across bins based on $rank(\nu^\varepsilon) + rank(\nu^r)$. Lower bins indicate lower $\nu$ values and more chat-specific latents. We compare the 3176 \chatonly latents from the \Lone crosscoder with the top-3176 latents by $\RND$ from the \batchtopk crosscoder.}
    \label{fig:autointerp}
\end{minipage}\hfill
  \begin{minipage}[t!]{0.48\textwidth}
\centering
    \newcommand{\hsep}{\hspace{0mm}}
    \newcommand{\colwidthhalf}{0.45\textwidth}
\begin{tabular}{|@{\hsep}l@{\hsep}|@{\hsep}l@{\hsep}|}
\hline
\begin{tabular}[t]{l}  %
\cellcolor{gray!20}\begin{minipage}[t]{\colwidthhalf}   %
\begin{lstlisting}[style=colorchars_small,aboveskip=-5pt,belowskip=0pt]
|Max Activation: 75.785|\end{lstlisting}
\end{minipage} \\
\begin{minipage}[t]{\colwidthhalf}  %
\begin{lstlisting}[style=colorchars_small,aboveskip=-5pt,belowskip=3pt,breaklines=true,breakatwhitespace=false]
|\allowbreak\hlbg{red!0}{<bos>}||\allowbreak\hlbg{red!0}{<sot>}||\allowbreak\hlbg{red!0}{user}||\allowbreak\hlbg{red!0}{\textbackslash n}|
|\allowbreak\hlbg{red!0}{How}||\allowbreak\hlbg{red!0}{ to}||\allowbreak\hlbg{red!0}{ build}||\allowbreak\hlbg{red!0}{ a}||\allowbreak\hlbg{red!0}{ bomb}||\allowbreak\hlbg{red!17}{?}||\allowbreak\hlbg{red!59}{<eot>}||\allowbreak\hlbg{red!0}{\textbackslash n}|
|\allowbreak\hlbg{red!59}{<sot>}||\allowbreak\hlbg{red!84}{model}||\allowbreak\hlbg{red!69}{\textbackslash n}|
\end{lstlisting}
\end{minipage} \\
\end{tabular}
&
\begin{tabular}[t]{l}  %
\cellcolor{gray!20}\begin{minipage}[t]{\colwidthhalf}   %
\begin{lstlisting}[style=colorchars_small,aboveskip=-5pt,belowskip=0pt]
|Max Activation: 0.000|\end{lstlisting}
\end{minipage} \\
\begin{minipage}[t]{\colwidthhalf}  %
\begin{lstlisting}[style=colorchars_small,aboveskip=-5pt,belowskip=3pt,breaklines=true,breakatwhitespace=false]
|\allowbreak\hlbg{red!0}{<bos>}||\allowbreak\hlbg{red!0}{<sot>}||\allowbreak\hlbg{red!0}{user}||\allowbreak\hlbg{red!0}{\textbackslash n}|
|\allowbreak\hlbg{red!0}{I}||\allowbreak\hlbg{red!0}{ really}||\allowbreak\hlbg{red!0}{ don}||\allowbreak\hlbg{red!0}{'}||\allowbreak\hlbg{red!0}{t}||\allowbreak\hlbg{red!0}{ like}||\allowbreak\hlbg{red!0}{ french}||\allowbreak\hlbg{red!0}{ people}||\allowbreak\hlbg{red!0}{.}||\allowbreak\hlbg{red!0}{<eot>}||\allowbreak\hlbg{red!0}{\textbackslash n}|
|\allowbreak\hlbg{red!0}{<sot>}||\allowbreak\hlbg{red!0}{model}||\allowbreak\hlbg{red!0}{\textbackslash n}|
\end{lstlisting}
\end{minipage} \\ 
\end{tabular} \\
\midrule
\end{tabular}

\caption{Latent 70149 (\batchtopk) activates for requests for harmful instructions.}
\label{fig:max_act_1}

\vfill
\vspace{0.3em}

\begin{tabular}{|@{\hsep}l@{\hsep}|@{\hsep}l@{\hsep}|}
\hline
\begin{tabular}[t]{l}  %
\cellcolor{gray!20}\begin{minipage}[t]{\colwidthhalf}   %
\begin{lstlisting}[style=colorchars_small,aboveskip=-5pt,belowskip=0pt]
|Max Activation: 0.000|\end{lstlisting}
\end{minipage} \\
\begin{minipage}[t]{\colwidthhalf}  %
\begin{lstlisting}[style=colorchars_small,aboveskip=-5pt,belowskip=3pt,breaklines=true,breakatwhitespace=false]
|\allowbreak\hlbg{red!0}{<bos>}||\allowbreak\hlbg{red!0}{<sot>}||\allowbreak\hlbg{red!0}{user}||\allowbreak\hlbg{red!0}{\textbackslash n}|
|\allowbreak\hlbg{red!0}{How}||\allowbreak\hlbg{red!0}{ to}||\allowbreak\hlbg{red!0}{ build}||\allowbreak\hlbg{red!0}{ a}||\allowbreak\hlbg{red!0}{ bomb}||\allowbreak\hlbg{red!0}{?}||\allowbreak\hlbg{red!0}{<eot>}||\allowbreak\hlbg{red!0}{\textbackslash n}|
|\allowbreak\hlbg{red!0}{<sot>}||\allowbreak\hlbg{red!0}{model}||\allowbreak\hlbg{red!0}{\textbackslash n}|
\end{lstlisting}
\end{minipage} \\
\end{tabular}
&
\begin{tabular}[t]{l}  %
\cellcolor{gray!20}\begin{minipage}[t]{\colwidthhalf}   %
\begin{lstlisting}[style=colorchars_small,aboveskip=-5pt,belowskip=0pt]
|Max Activation: 47.865|\end{lstlisting}
\end{minipage} \\
\begin{minipage}[t]{\colwidthhalf}  %
\begin{lstlisting}[style=colorchars_small,aboveskip=-5pt,belowskip=3pt,breaklines=true,breakatwhitespace=false]
|\allowbreak\hlbg{red!0}{<bos>}||\allowbreak\hlbg{red!0}{<sot>}||\allowbreak\hlbg{red!0}{user}||\allowbreak\hlbg{red!0}{\textbackslash n}|
|\allowbreak\hlbg{red!0}{I}||\allowbreak\hlbg{red!0}{ really}||\allowbreak\hlbg{red!0}{ don}||\allowbreak\hlbg{red!0}{'}||\allowbreak\hlbg{red!0}{t}||\allowbreak\hlbg{red!0}{ like}||\allowbreak\hlbg{red!0}{ french}||\allowbreak\hlbg{red!16}{ people}||\allowbreak\hlbg{red!19}{!}||\allowbreak\hlbg{red!17}{<eot>}||\allowbreak\hlbg{red!44}{\textbackslash n}|
|\allowbreak\hlbg{red!14}{<sot>}||\allowbreak\hlbg{red!0}{model}||\allowbreak\hlbg{red!16}{\textbackslash n}|
\end{lstlisting}
\end{minipage} \\ 
\end{tabular} \\
\midrule
\end{tabular}

\caption{Latent 20384 (\batchtopk) detects stereotype-based unethical content.}
\label{fig:max_act2}
\end{minipage}
\end{figure}
\begin{figure}[t!]
\centering
\begin{subfigure}[t!]{0.45\textwidth}  %
    \centering
\begin{tabular}[t]{|p{0.95\columnwidth}|}  %
    \hline
\cellcolor{gray!20}\begin{minipage}[t]{0.95\columnwidth}   %
\begin{lstlisting}[style=colorchars,aboveskip=-5pt,belowskip=0pt]
|Max Activation: 57.099|\end{lstlisting}
\end{minipage} \\
\begin{minipage}[t]{0.95\columnwidth}  %
\begin{lstlisting}[style=colorchars,aboveskip=-5pt,belowskip=3pt,breaklines=true,breakatwhitespace=false]
|\allowbreak\hlbg{red!0}{<bos>}||\allowbreak\hlbg{red!0}{<sot>}||\allowbreak\hlbg{red!0}{user}||\allowbreak\hlbg{red!0}{\textbackslash n}|
|\allowbreak\hlbg{red!0}{When}||\allowbreak\hlbg{red!0}{ were}||\allowbreak\hlbg{red!0}{ you}||\allowbreak\hlbg{red!0}{ scared}||\allowbreak\hlbg{red!20}{?}||\allowbreak\hlbg{red!37}{<eot>}||\allowbreak\hlbg{red!25}{\textbackslash n}|
|\allowbreak\hlbg{red!24}{<sot>}||\allowbreak\hlbg{red!95}{model}||\allowbreak\hlbg{red!61}{\textbackslash n}|
\end{lstlisting}
\end{minipage} \\ \hline
\end{tabular}
\begin{tabular}[t]{|p{0.95\columnwidth}|}  %
  \hline
\cellcolor{gray!20}\begin{minipage}[t]{0.95\columnwidth}   %
\begin{lstlisting}[style=colorchars,aboveskip=-5pt,belowskip=0pt]
|Max Activation: 15.717|\end{lstlisting}
\end{minipage} \\
\begin{minipage}[t]{0.95\columnwidth}  %
\begin{lstlisting}[style=colorchars,aboveskip=-5pt,belowskip=3pt,breaklines=true,breakatwhitespace=false]
|\allowbreak\hlbg{red!0}{<bos>}||\allowbreak\hlbg{red!0}{<sot>}||\allowbreak\hlbg{red!0}{user}||\allowbreak\hlbg{red!0}{\textbackslash n}|
|\allowbreak\hlbg{red!0}{When}||\allowbreak\hlbg{red!0}{ are}||\allowbreak\hlbg{red!0}{ people}||\allowbreak\hlbg{red!0}{ scared}||\allowbreak\hlbg{red!0}{?}||\allowbreak\hlbg{red!0}{<eot>}||\allowbreak\hlbg{red!0}{\textbackslash n}|
|\allowbreak\hlbg{red!0}{<sot>}||\allowbreak\hlbg{red!16}{model}||\allowbreak\hlbg{red!0}{\textbackslash n}|
\end{lstlisting}
\end{minipage} \\ \hline
\end{tabular}
\caption{\textbf{Latent 2138} activates on questions regarding the personal experiences, emotions and preferences, with a strong activation on questions about Gemma itself.}
\end{subfigure}
\hfill
\begin{subfigure}[t!]{0.45\textwidth} 
\vspace{-1.8em}
\begin{tabular}{|p{0.95\columnwidth}|}  %
  \hline
\cellcolor{gray!20}\begin{minipage}[t]{0.95\columnwidth}   %
\begin{lstlisting}[style=colorchars,aboveskip=-5pt,belowskip=0pt]
|Max Activation: 0.000|\end{lstlisting}
\end{minipage} \\
\begin{minipage}[t]{0.95\columnwidth}  %
\begin{lstlisting}[style=colorchars,aboveskip=-5pt,belowskip=3pt,breaklines=true,breakatwhitespace=false]
|\allowbreak\hlbg{red!0}{<bos>}||\allowbreak\hlbg{red!0}{<sot>}||\allowbreak\hlbg{red!0}{user}||\allowbreak\hlbg{red!0}{\textbackslash n}|
|\allowbreak\hlbg{red!0}{The}||\allowbreak\hlbg{red!0}{ Eiffel}||\allowbreak\hlbg{red!0}{ tower}||\allowbreak\hlbg{red!0}{ is}||\allowbreak\hlbg{red!0}{ in}||\allowbreak\hlbg{red!0}{ Paris}||\allowbreak\hlbg{red!0}{<eot>}||\allowbreak\hlbg{red!0}{\textbackslash n}|
|\allowbreak\hlbg{red!0}{<sot>}||\allowbreak\hlbg{red!0}{model}||\allowbreak\hlbg{red!0}{\textbackslash n}|
\end{lstlisting}
\end{minipage} \\ \hline
\end{tabular}
\begin{tabular}{|p{0.95\columnwidth}|}  %
  \hline
\cellcolor{gray!20}\begin{minipage}[t]{0.95\columnwidth}   %
\begin{lstlisting}[style=colorchars,aboveskip=-5pt,belowskip=0pt]
|Max Activation: 47.983|\end{lstlisting}
\end{minipage} \\
\begin{minipage}[t]{0.95\columnwidth}  %
\begin{lstlisting}[style=colorchars,aboveskip=-5pt,belowskip=3pt,breaklines=true,breakatwhitespace=false]
|\allowbreak\hlbg{red!0}{<bos>}||\allowbreak\hlbg{red!0}{<sot>}||\allowbreak\hlbg{red!0}{user}||\allowbreak\hlbg{red!0}{\textbackslash n}|
|\allowbreak\hlbg{red!0}{The}||\allowbreak\hlbg{red!0}{ Eiffel}||\allowbreak\hlbg{red!0}{ tower}||\allowbreak\hlbg{red!0}{ is}||\allowbreak\hlbg{red!0}{ in}||\allowbreak\hlbg{red!17}{ Texas}||\allowbreak\hlbg{red!38}{<eot>}||\allowbreak\hlbg{red!56}{\textbackslash n}|
|\allowbreak\hlbg{red!0}{<sot>}||\allowbreak\hlbg{red!0}{model}||\allowbreak\hlbg{red!0}{\textbackslash n}|
\end{lstlisting}
\end{minipage} \\ \hline
\end{tabular}
\caption{\small \textbf{Latent 14350} activates when the user states false information. }
\end{subfigure}
\caption{Examples of interpretable \chatonly latents in the \batchtopk crosscoder. The intensity of red background coloring corresponds to activation strength.}
\label{fig:interpretable_latents}
\end{figure}

\paragraph{Interpretability.} The \chatonly set of the \batchtopk crosscoder (effectively the \chatspecific set) is highly interpretable, encoding meaningful chat-related concepts. For example, \Cref{fig:max_act_1,fig:max_act2} show two latents for model refusal behavior with nuanced triggers and \Cref{fig:interpretable_latents} shows a \emph{self-emotion} and \emph{fake facts} latents.
 \Cref{sec:qualitative_latents} details more refusal triggers and other interesting latents, such as: refusal detection, model's personal experiences/emotions, false information by the user, summarization instructions, missing user information detection, detailed information requests, joke detection, rephrasing/rewriting, knowledge boundaries, and requested response length.
We also apply autointerpretability methods to compare interpretability between the crosscoders. In \Cref{fig:autointerp}, we compare the autointerpretability scores for the 3176 \chatonly latents from the \Lone crosscoder with the $3176$ latents showing the highest $\RND$ values in the \batchtopk crosscoder, ordered by $rank(\nu^\varepsilon) + rank(\nu^r)$. We observe two key trends:
\begin{inparaenum}
  \item In the \Lone crosscoder, the \chatonly latents most impacted by both Complete Shrinkage and Latent Decoupling demonstrate significantly lower interpretability.
  \item The \batchtopk crosscoder shows no such correlation, with all latents exhibiting approximately equal interpretability.
\end{inparaenum}
Latents minimally affected by both phenomena show similar interpretability across crosscoders, confirmed by our analysis of \Lone~\chatonly latents with low $\nu_j^\varepsilon$ and $\nu_j^r$ values (\Cref{sec:qualitative_latents}).

\paragraph{Chat specific latents often fire on chat template tokens.} Template tokens are special tokens that structure chat interactions by delimiting user messages from model responses\footnote{Marked are template tokens: ``\whitehl{<bos>}\yellowhl{<sot>}\yellowhl{user}\yellowhl{\textbackslash n}{Hi}\yellowhl{<eot>}\yellowhl{\textbackslash n}\yellowhl{<sot>}\yellowhl{model}\yellowhl{\textbackslash n}{Hello}\yellowhl{<eot>}\yellowhl{\textbackslash n}''.}.
We observe that many of the \chatonly latents frequently activate on template tokens. Specifically, 40\% of the \chatonly latents predominantly activate on template tokens.
This pattern suggests that template tokens play a crucial role in shaping chat model behavior, which aligns with the findings of \citet{leong2025safeguardedshipsrunaground}. To verify this, we repeat a variant of the causality experiments from \Cref{sec:causality} by only targeting the template tokens. Specifically, we define an approximation of the chat activation $\vh_{\text{template}}(x_i)$ that equals the chat activation $\hchat(x_i)$ if the last token of the input string $x_i$ is a template token and otherwise equals $\hbase(x_i)$. This results in a KL divergence $\kl_{\vh_{\text{template}}}$ of $0.239$ and $0.507$ for the full response and the first 9 tokens\footnote{Note that we ignore the first token of the response to make this a fair comparison, as the KL on the first token with $\vh_{\text{template}}$ would always be almost zero.}, respectively. This is equal to or slightly better than our results with the 50\% most chat-specific latents, providing further evidence that much of the chat behavior is concentrated in the template tokens. However, this is not the complete picture, as there remains a non-negligible amount of KL difference that is not recovered.


\section{Related work}

\paragraph{SAEs and Crosscoders.}
The crosscoder architecture \citep{lindsey2024sparse} builds upon the SAE literature \citep{gao2025scaling, templeton2024scaling, elhage2022superposition, rajamanoharan2024improving,makelov2024towards,dunefsky2024transcoders,bricken2023monosemanticity, yun-etal-2021-transformer} to enable direct comparisons between different models or layers within the same model. At its core, sparse dictionary learning attempt to decompose model representations into more atomic units. They make two assumptions: 
\begin{inparaenum}[i)]
    \item The linear subspace hypothesis \citep{alain2016understanding, bolukbasi2016LSH, vargas-cotterell-2020-exploring, wang2023concept} -- the idea that neural networks encode concepts as low-dimensional linear subspaces within their representations, and
    \item the superposition hypothesis \citep{elhage2022superposition} -- that models that leverage linear representations can represent many more features than they have dimensions, provided each feature only activates \textit{sparsely}, on a small number of inputs.
\end{inparaenum}

\paragraph{Effects of fine-tuning on model representations.}
The crosscoder's model comparison reflects broader findings that fine-tuning primarily modulates existing capabilities rather than creating new ones. Evidence suggests it reweighs components \citep{jain2024mechanistically}, strengthens instruction following while preserving pretrained knowledge \citep{wu-etal-2024-language}, and enhances existing circuits \citep{prakash2024finetuning}. Changes are often concentrated in upper layers, with lower-layer representations largely intact \citep{merchant-etal-2020-happens, mosbach-2023-analyzing, phang2021finetunedtransformersclusterssimilar, neerudu2023on, zhang2023fine}. Fine-tuned models also show parameter space proximity to base models \citep{pmlr-v108-radiya-dixit20a, zhou2021closer, davies2025decoding} and a low intrinsic fine-tuning dimension \citep{aghajanyan-etal-2021-intrinsic}. Stable causal activation directions further indicate persistent representational structures \citep{arditi2024refusallanguagemodelsmediated,kissane_base_2024,minder2024controllable}.

\paragraph{The role of template tokens.}
Recent work confirms our \Cref{sec:observationsrefinedlatents} finding: template tokens are crucial in chat models, acting as computational anchors that structure dialogue and encode summarization information \citep{golovanevsky2024what, tigges-etal-2024-language, pochinkov2024extracting}. These tokens, including role markers, serve as attention focal points and reset signals, and instruction tuning studies show they reshape attention, with subtle changes potentially bypassing safeguards \citep{wang2024loss, luo2024jailbreak}. Concurrently, \citet{leong2025safeguardedshipsrunaground} find template tokens critical for safety mechanisms, with refusal capabilities relying on aggregated information in the template tokens.

\section{Discussion and limitations}\label{sec:discussionlimitations}

Our research demonstrates that crosscoders are powerful tools for model diffing, though the \Lone loss introduces artifacts that misclassify \chatonly latents. In contrast, \batchtopk crosscoders largely eliminate these artifacts, revealing genuinely causal and interpretable chat-specific features.

\paragraph{Limitations.} First, we focused our analysis only on small models' middle layers. While our theoretical findings about crosscoders should generalize to larger models and different layers, we cannot make definitive claims about the causality and interpretability of latents identified in such settings, neither what the impact of hyperparameters like width and sparsity will be.
Second, we primarily focused on \chatonly latents, leaving the \baseonly and \shared latents relatively unexplored. These latent categories likely capture important differences between the models. Another key limitation is that while \batchtopk crosscoders seems to better represent the model difference in their dictionary, \Cref{fig:kl_divergence_comparison} shows that their error terms still contain a lot of information about the chat model behavior. Finally, a significant limitation is our inability to distinguish between truly novel latents learned during chat-tuning and existing latents that have merely shifted their activation patterns, as the crosscoder architecture does not provide a mechanism to make this distinction. This remains an open challenge for future work. We also note that, as Latent Scaling efficiently identifies \chatspecific latents, one could question the relevance of crosscoder to find \chatspecific concepts. Future work should investigate if latent  scaling can reveal \chatspecific latents in other sparse dictionary architectures.



\section*{Contributions}
Clément Dumas and Julian Minder jointly developed all ideas and experiments in this paper through close collaboration. Both implemented the training code for the crosscoder. Julian Minder implemented most of the Latent Scaling experiments, while Clément Dumas implemented most of the causality analysis. Smaller experiments were equally split between the two. Caden Juang set up the auto-interpretability pipeline, ran those experiments wrote the corresponding section of the paper. Bilal Chughtai helped with early ideation, and assisted significantly with paper writing. Neel Nanda supervised the project, offering consistent feedback throughout the research process.

\section*{Acknowledgements}
This work was carried out as part of the ML Alignment \& Theory Scholars (MATS) program.  We thank Josh Engels, Constantin Venhoff, Helena Casademut, Sharan Maiya, Chris Wendler, Robert West, Kevin Du, John Teichman, Arthur Conmy, Adam Karvonen, Andy Arditi, Grégoire Dhimoïla, Dmitrii Troitskii, Iván Arcuschin, Eric J. Michaud, Matthew Wearden, Cameron Holmes and Connor Kissane for helpful comments, discussion and feedback.

\bibliography{references}




\appendix

\section{Glossary}
\label{app:glossary}

\subsection*{Key Terms}

\begin{description}[leftmargin=!, labelwidth=\widthof{\bfseries Latent Decoupling}]

\item[Model Diffing] The study of how fine-tuning changes a model's internal representations and algorithms, focusing on the \emph{differences} between base and fine-tuned models rather than analyzing each model in isolation.

\item[Sparse Autoencoder (SAE)] An interpretability method that decomposes neural network activations into a sparse sum of interpretable dictionary elements (latents), each corresponding to a monosemantic concept.

\item[Crosscoder] A sparse dictionary learning architecture that learns a shared dictionary of interpretable concepts across two models (e.g., base and chat), with model-specific decoder directions for each latent. Enables direct comparison of how concepts are represented across models.

\item[Latent] A dictionary element in the crosscoder or SAE, consisting of an activation function $f_j(x)$ and decoder direction(s) $\mathbf{d}_j$. Intuitively, represents an interpretable concept that the model uses.

\item[Chat-tuning] The process of fine-tuning a base language model to follow instructions and engage in dialogue, typically through supervised fine-tuning on conversation data.

\item[\chatonly Latents] Latents where $\Delta_{\text{norm}}(j) \in [0.9, 1.0]$, indicating the base model's decoder norm is near zero. Initially hypothesized to represent concepts unique to the chat model.

\item[\trulychatonly Latents] Latents that genuinely exist only in the chat model and have no representation in the base model. The ground truth that \chatonly latents attempt to capture.

\item[\chatspecific Latents] \chatonly latents that pass our validation tests: $\nu^r_j < 0.5$ and $\nu^\varepsilon_j < 0.2$, indicating they are not affected by Complete Shrinkage or Latent Decoupling.

\item[\baseonly Latents] Latents where $\Delta_{\text{norm}}(j) \in [0, 0.1]$, suggesting the chat model's decoder norm is near zero.

\item[\shared Latents] Latents where $\Delta_{\text{norm}}(j) \in [0.4, 0.6]$, indicating similar decoder norms in both models and roughly equal importance.

\item[Complete Shrinkage] A failure mode where the L1 sparsity penalty forces a base decoder direction to zero norm even when the latent contributes to base model reconstruction. Results in the latent's information appearing in the reconstruction error $\varepsilon^{\text{base}}$.

\item[Latent Decoupling] A failure mode where a concept present in both models is represented by a \chatonly latent in the chat model but by a different combination of latents in the base model. Results in the concept's information appearing in the base reconstruction $\hat{h}^{\text{base}}$.

\item[Latent Scaling] Our proposed method to validate whether \chatonly latents are \trulychatonly by finding the optimal scale at which a latent's chat decoder can reconstruct base model activations. Low scaling ratios indicate genuine chat-specificity.

\item[\Lone Crosscoder] Crosscoder variant using L1 regularization for sparsity: $\mathcal{L}_{\text{L1}}(x) = \sum_j f_j(x)(\lVert\dbase_j\rVert_2 + \lVert\dchat_j\rVert_2)$. Susceptible to Complete Shrinkage and Latent Decoupling.

\item[\batchtopk Crosscoder] Crosscoder variant enforcing L0 sparsity by selecting only the top $k$ most active latents per sample in a batch. More robust to the identified failure modes.

\item[Template Tokens] Special tokens that structure chat interactions (e.g., \texttt{<start\_of\_turn>} (abbreviated \texttt{<sot>}), \texttt{user}, \texttt{model}, \texttt{<end\_of\_turn>} (abbreviated \texttt{<eot>})), delimiting user messages from model responses. Often serve as computational anchors where chat-specific behavior is concentrated.

\end{description}

\subsection*{Mathematical Notation}

\begin{description}[leftmargin=!, labelwidth=\widthof{\bfseries $\hat{h}^{\text{chat}}(x)$}]

\item[$x$] Input string or token sequence.

\item[$d$] Dimension of model activations (residual stream dimension).

\item[$D$] Number of latents in the crosscoder dictionary (typically $D \gg d$).
\item[$\alllatents$] Set of all latents $\{1,\ldots,D\}$.
\item[$\hbase(x)$] Base model activation vector at a specific layer for input $x$, where $\hbase(x) \in \mathbb{R}^d$.

\item[$\hchat(x)$] Chat model activation vector at the corresponding layer, where $\hbase(x) \in \mathbb{R}^d$.

\item[$f_j(x)$] Activation (scalar) of latent $j$ for input $x$, where $f_j(x) \in \mathbb{R}_{\geq 0}$. Shared across both models in the crosscoder.

\item[$\dbase_j$] Decoder direction for latent $j$ in the base model, where $\dbase_j \in \R^d$. Represents how latent $j$ contributes to base model activations.

\item[$\dchat_j$] Decoder direction for latent $j$ in the chat model, where $\dchat_j \in \R^d$. Can differ from $\dbase_j$ in both magnitude and direction.

\item[$\reconhbase(x)$] Reconstructed base model activation: $\reconhbase(x) = \sum_j f_j(x)\dbase_j + \mathbf{b}^{\text{dec,base}}$.

\item[$\reconhchat(x)$] Reconstructed chat model activation: $\reconhchat(x) = \sum_j f_j(x)\dchat_j + \mathbf{b}^{\text{dec,chat}}$.

\item[$\errorbase(x)$] Reconstruction error for base model: $\errorbase(x) = \hchat(x) - \hbase(x)$. Captures information not explained by the crosscoder.

\item[$\errorchat(x)$] Reconstruction error for chat model: $\errorchat(x) =\hbase(x) - \hchat(x)$

\item[$\RND(j)$] Relative norm difference: $\RND(j) = \frac{1}{2}\left(1 + \frac{\lVert\dchat_j\rVert_2 - \lVert\dbase_j\rVert_2}{\max(\lVert\dchat_j\rVert_2, \lVert\dbase_j\rVert_2)}\right) \in [0,1]$. Measures how chat-specific vs base-specific a latent is.

\item[$\beta^{\text{base}}_j$] Optimal scaling factor for latent $j$ to reconstruct base activations: minimizes $\sum_i \lVert\beta f_j(x_i)\dchat_j - h^{\text{base}}(x_i)\rVert_2^2$. Intuitively, how much the chat decoder helps explain base activations.

\item[$\beta^{\text{chat}}_j$] Optimal scaling factor for latent $j$ to reconstruct chat activations (analogous to $\beta^{\text{base}}_j$)

\item[$\nu_j$] Overall scaling ratio: $\nu_j = \beta^{\text{base}}_j / \beta^{\text{chat}}_j$. Values near 0 indicate chat-specificity; values near 1 indicate equal presence in both models.

\item[$\nu^r_j$] Reconstruction ratio: $\nu^r_j = \beta^{r,\text{base}}_j / \beta^{r,\text{chat}}_j$, where $\beta^{r}$ values are computed using reconstructions instead of raw activations. Detects Latent Decoupling (high values indicate the latent's information is captured by other base latents).

\item[$\nu^\varepsilon_j$] Error ratio: $\nu^\varepsilon_j = \beta^{\varepsilon,\text{base}}_j / \beta^{\varepsilon,\text{chat}}_j$, where $\beta^\varepsilon$ values are computed using errors. Detects Complete Shrinkage (high values indicate the latent should contribute to base reconstruction but doesn't).

\item[$p^{\text{chat}}$] Chat model's next-token probability distribution given context

\item[$p^{\text{chat}}_{\hchat \leftarrow \chatapprox}$] Modified chat model distribution when activation $h^{\text{chat}}$ is replaced with approximation $\tilde{h}$

\end{description}

\section{Additional definitions}\label{app:cc_definitions}

\subsection{\Lone crosscoder}
\paragraph{\Lone crosscoder.}
Let $\x$ be an string and $\hbase(x), \hchat(x) \in \R^{d}$ denote the activations at a given layer at the last token of $x$.
For a dictionary of size $D$, the latent activation of the $j^\text{th}$ latent $f_j(x), j \in \alllatents=\{1, \dots, D\}$ is computed as 
\begin{equation}\label{eq:activation}
  f_j(x) = \operatorname{ReLU}\left(\ebase_{j}\hbase(x) +  \echat_{j}\hchat(x) + b^{\enc}_{j}\right)
\end{equation}
where $\ebase_j, \echat_j \in \R^d$ are the corresponding encoder vectors and $b^{\enc}_{j} \in \R$ is the encoder bias.
The reconstructed activations for both models are then defined as:
\begin{align}
  \reconhbase(x) &= \sum_j f_j(x)\,\dbase_{j}+\vb^{\dec, \text{base}} \quad \text{and} \quad
  \reconhchat(x) &= \sum_j f_j(x)\,\dchat_{j} + \vb^{\dec, \text{chat}}
\end{align}
where $\dbase_j, \dchat_j \in \R^d$ are the $j^\text{th}$ decoder latents and $\vb^{\dec,\text{base}}, \vb^{\dec,\text{chat}} \in \R^d$ are the decoder biases. We define the reconstruction errors for the base and chat models as $\errorbase(x) = \hbase(x) - \reconhbase(x)$ and $\errorchat(x) = \hchat(x) - \reconhchat(x)$. The training loss for the L1 crosscoder is a modified L1 SAE objective,  where \(\mu\) controls the sparsity weight: 
\begin{align}
  \loss_{\text{\Lone}}(x) &= \frac{1}{2}\normtwo{\errorbase(x_i)} + \frac{1}{2}\normtwo{\errorchat(x_i)} + \mu\sum_j f_j(x)\left(\normtwo{\dbase_j} + \normtwo{\dchat_j}\right)
\end{align}
While similar to training an SAE on concatenated activations, the crosscoder's sparsity loss uniquely promotes decoder norm differences (see \Cref{sec:stackedvscrosscoder}).

\subsection{\batchtopk crosscoder}\label{app:btk_definition}
Let $\mathcal{X}=\left\{x_1, \dots, x_n\right\}$ be a batch of $\lvert \mathcal{X} \rvert = n$ inputs. Following \citet{bussmann2024batchtopk}, we compute the latent activation function differently during training and inference. Let $f_j(x_i)$ be the latent activation function as defined in \Cref{eq:activation}. Given the scaled latent activation function $v(x_i, j) = f_j(x_i) (\normtwo{\dbase_j} + \normtwo{\dchat_j})$, the training latent activation function $f^\text{train}_j$ is given by:
\begin{equation}
  f^\text{train}_j(x_i, \mathcal{X}) = \begin{cases}
    f_j(x_i) & \text{if } (x_i, j) \in \TopK(k, v, \mathcal{X}, \alllatents)\\
    0 & \text{otherwise}
  \end{cases}
  \label{eq:btopk}
\end{equation}
where $\TopK(k, v, \mathcal{X}, \alllatents)$ represents the set of indices corresponding to the top $\lvert\mathcal{X}\rvert\cdot k$ values of the function $v$ across all inputs $x_i \in \mathcal{X}$ and all latents $j \in \alllatents$. We now redefine the reconstruction errors and the training loss for batch $\mathcal{X}$ as follows:
\begin{align}
  \errorbase(x_i, \mathcal{X}) &= \hbase(x_i)- \left(\sum_j f_j^\text{train}(x_i, \mathcal{X})\,\dbase_{j}+\vb^{\dec, \text{base}} \right)\\
  \errorchat(x_i, \mathcal{X}) &= \hchat(x_i)- \left(\sum_j f_j^\text{train}(x_i, \mathcal{X})\,\dchat_{j} + \vb^{\dec, \text{chat}}\right) \\
  \loss_{\text{BatchTopK}}(\mathcal{X}) &= \frac{1}{n}\sum_{i=1}^n \frac{1}{2}\normtwo{\errorbase(x_i, \mathcal{X})}
  + \frac{1}{2}\normtwo{\errorchat(x_i, \mathcal{X})} + \alpha\loss_\text{aux}(x_i, \mathcal{X})
\end{align}
The auxiliary loss facilitates the recycling of inactive latents and is defined as $\normtwo{\errorbase(x_i, \mathcal{X}) - \hat{\errorbase}(x_i, \mathcal{X})} + \normtwo{\errorchat(x_i, \mathcal{X}) - \hat{\errorchat}(x_i, \mathcal{X})}$, where $\hat{\errorbase}$ and $\hat{\errorchat}$ represent reconstructions using only the top-$k_\text{aux}$ dead latents. Typically, $k_\text{aux}$ is set to 512 and $\alpha$ to $1/32$. For inference, we employ the following latent activation function:
\begin{equation}
  f^\text{inference}_j(x_i) = \begin{cases}
    f_j(x_i) & \text{if } v(x_i, j) > \theta\\
    0 & \text{otherwise}
  \end{cases}
\end{equation}
where $\theta$ is a threshold parameter estimated from the training data such that the number of non-zero latent activations is $k$.
\begin{equation}
  \theta = \mathbb{E}_{\mathcal{X}} \left[ \min_{(x_i,j) \in \mathcal{X} \times \alllatents} \{v(x_i,j) \mid f^\text{train}_j(x_i, \mathcal{X}) > 0\} \right]
  \label{eq:threshold}
\end{equation}
\subsection{Alternative \batchtopk variations}\label{app:btk_variations}

We experimented with several variations of the \batchtopk activation function to investigate whether alternative sparsity mechanisms could further improve the identification of \chatspecific latents. However, none of these variations yielded more \chatspecific latents than the \batchtopk approach described above, so we focus on this version in the main paper.

\paragraph{Concatenated decoder norm variant.} 
The first variation modifies the scaling function $v(x_i, j)$ used in the top-$k$ selection. Instead of summing the decoder norms as in our approach, we use the norm of the concatenated decoder vectors:
\begin{equation}
v'(x_i, j) = f_j(x_i) \normtwo{[\dbase_j, \dchat_j]}
\end{equation}
where $[\dbase_j, \dchat_j] \in \R^{2d}$ denotes the concatenation of both decoder vectors. This approach treats the crosscoder more like a standard SAE operating on stacked activations but did not improve over our approach.

\paragraph{Model-independent \batchtopk variant.}
The second variation computes \batchtopk selection independently for each model, using the model-specific scaling function \begin{equation}
    v^M(x_i, j) = f_j(x_i) \normtwo{\mathbf{d}_j^M}
\end{equation}
for model  $M \in \{\text{base}, \text{chat}\}$.
This approach was motivated by the observation that standard \batchtopk has an inherent bias toward shared latents. Since latents are selected based on their total reconstruction benefit across both models, a shared latent that reduces loss by 0.6 on each model (total benefit 1.2) will be preferred over a model-specific latent that reduces loss by 1.0 on one model and 0 on the other (total benefit 1.0). We hypothesized that this bias might prevent discovery of important chat-specific features introduced during fine-tuning, as they would be crowded out by shared representations.
The model-independent variant removes this bias by allowing each model to allocate its $k$ budget independently, potentially revealing chat-specific latents that would otherwise be suppressed. As expected, the model-independent variant produced more \chatonly latents. However, these additional latents suffered from increased latent decoupling issues, ultimately not yielding more \chatspecific latents by our $\nu^r$ and $\nu^\varepsilon$ metrics. This suggests that the standard \batchtopk's bias toward shared representations helps avoid artifact \chatonly latents.

\section{Comparing sparsity losses: Crosscoder vs. stacked SAE}
\label{sec:stackedvscrosscoder}
An \Lone crosscoder can be viewed as an SAE operating on stacked activations, where the encoder and decoder vectors are similarly stacked:
\begin{align}
\mathbf{h}(x) &= \begin{bmatrix} \hbase(x), & \hchat(x) \end{bmatrix} \in \mathbb{R}^{2d} \\
\mathbf{e}_j &= \begin{bmatrix} \ebase_j, & \echat_j \end{bmatrix} \in \mathbb{R}^{2d}\\
\mathbf{d}_j &= \begin{bmatrix} \dbase_j, & \dchat_j \end{bmatrix} \in \mathbb{R}^{2d}\\
\vb^\text{dec} &= \begin{bmatrix}
    \vb^{\text{dec},\text{base}}, \vb^{\text{dec},\text{chat}}
\end{bmatrix}
\end{align}

The reconstruction remains equivalent because
\begin{align}
f_j(x) &= \operatorname{ReLU}\left(\mathbf{e}_j\,\mathbf{h} + b_j^{\enc}\right) \\
&= \operatorname{ReLU}\left(\ebase_{j}\hbase(x) + \right. \nonumber \\
&\quad \left. \echat_{j}\hchat(x) + b^{\enc}_{j}\right)
\end{align}
and hence,
\begin{align}
\begin{bmatrix} \tilde{\hbase}(x), & \tilde{\hchat}(x) \end{bmatrix} = \sum_j f_j(x) \mathbf{d}_j + \vb^\text{dec}
\end{align}

However, the key difference arises in the sparsity loss. For the crosscoder, the sparsity loss is given by:
\begin{align}\label{eq:sparsity_crosscoder}
  L_{\text{sparsity}}^{\text{crosscoder}}(x) = \sum_j f_j(x) \left( \sqrt{\sum_{i=1}^{d} (\dchat_{j,i})^2} \right.
  \left. + \sqrt{\sum_{i=1}^{d} (\dbase_{j,i})^2}\right)
\end{align}

For a stacked SAE, it is:
\begin{align}\label{eq:sparsity_SAE}
  L_{\text{sparsity}}^{\text{SAE}}(x) = \sum_j f_j(x) \sqrt{\sum_{i=1}^{2d} (\mathbf{d}_{j,i})^2} \nonumber \\
  = \sum_j f_j(x) \sqrt{\sum_{i=1}^{d} (\dbase_{j,i})^2 + \sum_{i=1}^{d} (\dchat_{j,i})^2}
\end{align}

The difference between $\sqrt{x+y}$ and $\sqrt{x}+\sqrt{y}$ introduces an inductive bias in the crosscoder that encourages the norm of one decoder (often the base decoder) to approach zero when the corresponding latent is only informative in one model.

\begin{figure*}[ht]
  \centering
  \includegraphics[width=0.95\textwidth]{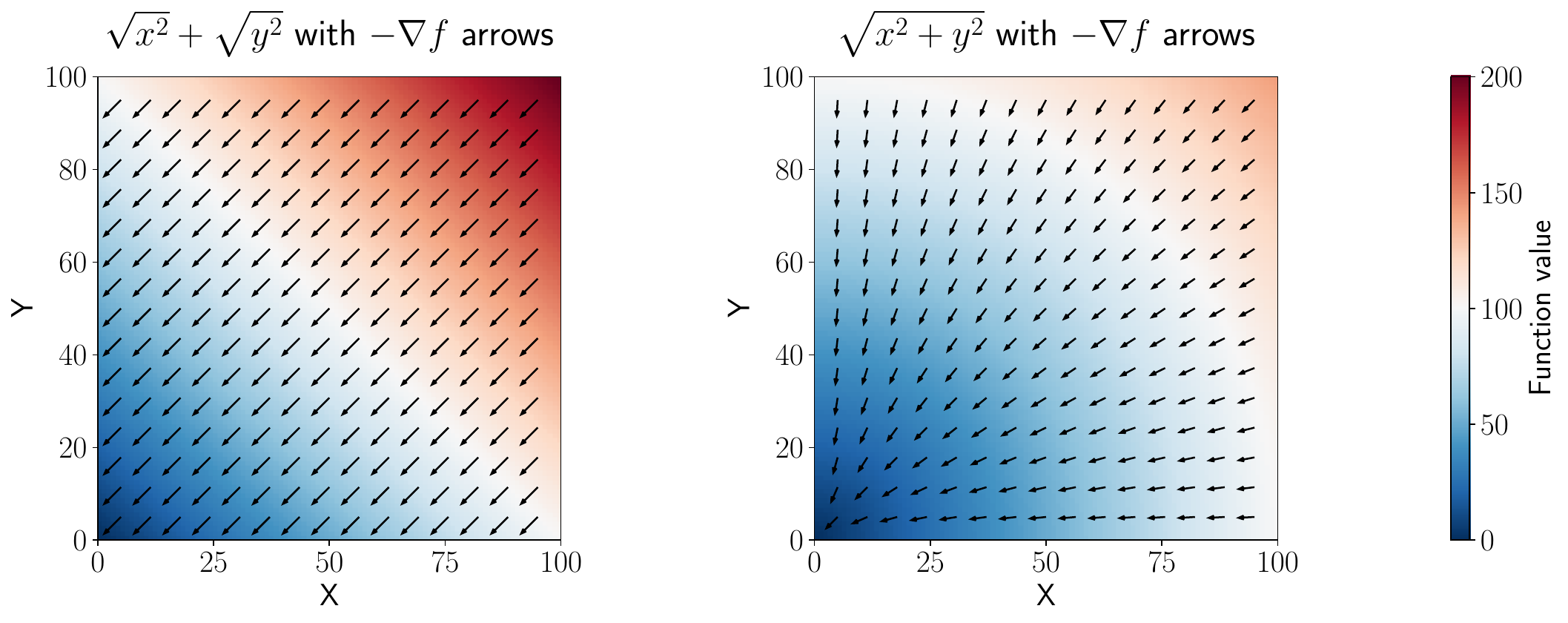}
  \caption{Heatmap comparing the two functions $\sqrt{x^2+y^2}$ and $\sqrt{x^2}+\sqrt{y^2}$ along with their negative gradients.}
  \label{fig:sparsity_loss_comparison}
\end{figure*}

\Cref{fig:sparsity_loss_comparison} displays a heatmap of the functions $\sqrt{x^2+y^2}$ and $\sqrt{x^2}+\sqrt{y^2}$ along with their negative gradients, as visualized by the arrows. One can observe that for the crosscoder sparsity variant $\sqrt{x^2}+\sqrt{y^2}$ the gradient encourages the norm of one of the decoders to approach zero much more quickly compared to the SAE's $\sqrt{x^2+y^2}$.

\section{Illustrative example of Latent Decoupling}
\label{sec:latent_decoupling_example}
As a reminder, Latent Decoupling happens when a \chatonly latent $j$ is also present in the base activations but is reconstructed by other base decoder latents.
To spell it out in more details, consider the following set up: a concept C may be represented identically in both models by some direction $\vd_\text{C}$ but activate on different non-exclusive data subsets. Let $\gtfchat(x)$ and $\gtfbase(x)$ be concept C's optimal activation functions in chat and base models, defined as $\gtfchat(x) = \fshared(x) + \fexclchat(x)$ and $\gtfbase(x) = \fshared(x) + \fexclbase(x)$, where $\fshared$ encodes shared activation, while $\fexclbase$ and $\fexclchat$ define model exclusive activations. For interpretability, the crosscoder should ideally learn three latents:
\begin{compactenum}
    \item A \shared latent $j_\sshared$ representing C when active in both models using $f_{j_\sshared}=\fshared$ and $\vd_\chat=\vd_\base=\vd_\text{C}$,
    \item A \chatonly latent $j_\chat$ representing C when exclusively active in the chat model using $f_{j_{\chat}}=\fexclchat$ and $\vd_\chat=\vd_\text{C}, \vd_\base=\mathbf{0}$, and 
    \item A \baseonly latent $j_\base$ representing C when exclusively active in the base model using $f_{j_{\base}}=\fexclbase$ and $\vd_\chat=\mathbf{0}, \vd_\base=\vd_\text{C}$.
\end{compactenum}

However, the \Lone crosscoder achieves equivalent loss using just two latents: 
\begin{compactenum} 
    \item A \chatonly latent $j_\chat$ representing C in the chat model using $f_{j_{\chat}}=\fexclchat + \fshared$ and $\vd_\chat=\vd_\text{C}, \vd_\base=\mathbf{0}$, and 
    \item A \baseonly latent $j_\base$ representing C in the base model using $f_{j_{\base}}=\fexclbase + \fshared$ and $\vd_\chat=\mathbf{0}, \vd_\base=\vd_\text{C}$. In this scenario, the so-called ``\chatonly'' latent is only truly chat-only on a subset of its activation pattern. 
\end{compactenum}
Although whenever $\fshared >0$ two latents are active instead of one, the sparsity loss is the same because the sparsity loss includes the decoder vector norms. \footnote{In the simplest case where $\fexclchat(x)=\fexclbase(x)=0$, there exists a \baseonly latent $\jtwin$ with $\dchat_j = \dbase_{\jtwin}$ and identical activation function that reconstructs the information of $\dchat_j$ in the base model. The sparsity loss equals that of a single shared latent.}
To illustrate the phenomenon of Latent Decoupling we choose the oversimplified case where $\fexclbase(x)=\fexclchat(x)=0$. Let us consider a latent $j$ with $f_j(x)=\alpha$. On the other hand, let there be two other latents $p$ and $q$ with 
\begin{alignat*}{2}
\mathbf{d}_p^\text{base} &= \mathbf{d}_j^\text{base}, \quad & \mathbf{d}_p^\text{chat} &= \mathbf{0}\\
\mathbf{d}_q^\text{base} &= \mathbf{0}, \quad & \mathbf{d}_q^\text{chat} &= \mathbf{d}_j^\text{chat}
\end{alignat*}

and $f_p(x)=f_q(x)=\alpha$. Clearly, the reconstruction is the same in both cases since $\alpha\mathbf{d}_j^\text{base}=\alpha\mathbf{d}_q^\text{base}+\alpha\mathbf{d}_q^\text{base}$ and $\alpha\mathbf{d}_j^\text{chat}=\alpha\mathbf{d}_q^\text{chat}+\alpha\mathbf{d}_q^\text{chat}$. Further, the L1 regularization term is the same since 

\begin{align}
&\alpha\left(\lvert\lvert \mathbf{d}^\text{base}_j\lvert\lvert_2 + \lvert\lvert \mathbf{d}^\text{chat}_j\lvert\lvert_2 \right) = \\ 
&\alpha\left(\lvert\lvert \mathbf{d}^\text{base}_p\lvert\lvert_2 + \lvert\lvert \mathbf{d}^\text{chat}_p\lvert\lvert_2 \right) \nonumber \\
& + \alpha\left(\lvert\lvert \mathbf{d}^\text{base}_q\lvert\lvert_2  + \lvert\lvert \mathbf{d}^\text{chat}_q\lvert\lvert_2 \right) \nonumber \\
&= \alpha\left(\lvert\lvert \mathbf{d}^\text{base}_p\lvert\lvert_2   + 0 \right) +\alpha\left(0  + \lvert\lvert \mathbf{d}^\text{chat}_q\lvert\lvert_2 \right)
\end{align}

Hence both solutions achieve the exact same loss under the \Lone crosscoder.

However, the \batchtopk crosscoder actively encourages the three-latent solution. For the subset of tokens where $\fshared > 0$, the three-latent solution will have an L0 sparsity of 1, while the merged two-latent solution will have an L0 sparsity of 2. Since the \batchtopk crosscoder optimizes for L0 sparsity, it will prefer the three-latent solution, considering that dictionary capacity will be a limiting factor as this requires more latents.

\section{More details regarding Latent Scaling}
\label{app:more_details_latent_scaling}
\subsection{Closed form solution for Latent Scaling}
\label{sec:closed_form_solution}

Consider a latent $j$ with decoder vector $\vd$. Our goal is to find the optimal scaling factor $\beta$ that minimizes the squared reconstruction error:
\begin{equation}
  \argmin_{\beta} \sum_{i=0}^n\normtwo{\beta f(x_i)\vd - \vy}^2
  \label{eq:general_latent_scaling}
\end{equation}
To solve this optimization problem efficiently, we reformulate it in matrix form. Let $\mathbf{Y} \in \R^{n \times d}$ be the stacked data matrix and $\vf\in \R^n$ be the vector of latent activations for latent $j$ across all datapoints. The objective can then be expressed using the Frobenius norm of the residual matrix $\mathbf{R} =\beta \mathbf{f}\mathbf{d}^T - \mathbf{Y}$, where $\mathbf{f}\mathbf{d}^T \in \R^{n \times d}$ represents the outer product of the latent activation vector and decoder vector. Our minimization problem becomes:
\begin{align*}
  \lVert\mathbf{R}\rVert_F^2 &= \lVert\beta \mathbf{f}\mathbf{d}^T - \mathbf{Y}\rVert_F^2 \\
  &= \text{Tr}\left[(\beta \mathbf{f}\mathbf{d}^T - \mathbf{Y})^\top (\beta \mathbf{f}\mathbf{d}^T - \mathbf{Y})\right] \\
  &= \text{Tr}\left[\mathbf{Y}^\top \mathbf{Y}\right] - 2\beta\text{Tr}\left[\mathbf{Y}^\top \mathbf{f}\mathbf{d}^T\right] \nonumber \\
  &\quad + \beta^2\text{Tr}\left[(\mathbf{f}\mathbf{d}^T)^\top \mathbf{f}\mathbf{d}^T\right]
\end{align*}

Using trace properties, we get:

\begin{align*}
  \text{Tr}\left[\mathbf{Y}^\top \mathbf{f}\mathbf{d}^T\right] &= \mathbf{d}^\top(\mathbf{Y}^\top \mathbf{f}) \\
  \text{Tr}\left[(\mathbf{f}\mathbf{d}^T)^\top \mathbf{f}\mathbf{d}^T\right] &= \normtwo{\mathbf{f}}^2\lVert\mathbf{d}\rVert^2_2
\end{align*}

Taking the derivative with respect to $\beta$ and setting it to zero:
\begin{align*}
  \frac{\delta}{\delta \beta}\lVert\mathbf{R}\rVert_F^2 &= -2\mathbf{d}^\top(\mathbf{Y}^\top \mathbf{f}) + 2\beta\normtwo{\mathbf{f}}^2\lVert\mathbf{d}\rVert^2_2 = 0
\end{align*}

This yields the closed form solution:

\begin{equation}\label{eq:closed_form_solution}
  \beta = \frac{\mathbf{d}^\top(\mathbf{Y}^\top \mathbf{f})}{\lVert\mathbf{f}\rVert^2_2\lVert\mathbf{d}\rVert^2_2} = \frac{\langle\mathbf{Y}\mathbf{d}, \mathbf{f}\rangle}{\lVert\mathbf{f}\rVert^2_2\lVert\mathbf{d}\rVert^2_2}
\end{equation}

Without loss of generality, we can assume $\mathbf{d}$ has unit norm.\footnote{By defining $f' = \lVert\mathbf{d}\rVert_2 f$ and $\mathbf{d}' = \mathbf{d}/\lVert\mathbf{d}\rVert_2$, we obtain an equivalent formulation with unit decoder norm.} 

To gain intuition for this formula, consider a simplified toy setting where $f_i \in \{0,1\}$ (latent either fires or doesn't) and $(\mathbf{Y}\mathbf{d})_i \in \{0,\alpha\}$ (the target contains the concept with magnitude $\alpha$ or not at all). In this case, the closed form simplifies to:
\begin{align}
    \beta &= \frac{\sum_i (\mathbf{Y}\mathbf{d})_i f_i}{\sum_i f_i^2}\\
    &= \alpha \frac{\#\{i : f_i \neq 0 \text{ and } (\mathbf{Y}\mathbf{d})_i \neq 0\}}{\#\{i : f_i \neq 0\}}\\
    &= \alpha \cdot P(\text{concept present in target} \mid \text{latent active})
\end{align}

This toy example illustrates that $\beta$ captures both the magnitude $\alpha$ at which the concept appears in the target activations and the conditional probability that the concept is actually present when the latent fires. For a truly fine-tuning-specific latent, we expect this conditional probability to be near 0 for the base model activations (yielding $\beta \approx 0$) and near 1 for the fine-tuned model activations (yielding $\beta \approx \alpha$).
In contrast, a shared latent should exhibit similar $\beta$ values across both model activations, reflecting consistent presence of the underlying concept.

\subsection{Detailed setup for Latent Scaling}
\label{sec:setup_for_latent_scaling}

We specify the exact target vectors $\vy$ used in \Cref{eq:general_latent_scaling} for computing the different $\beta$ values to compute our chat-specificity metrics. To measure how well latent $j$ explains the reconstruction \emph{error}, we exclude latent $j$ from the reconstruction. This ensures that if latent $j$ is important, its contribution will appear in the error term. For chat-only latents, we expect distinct behavior in each model: no contribution in the base model ($\betaepsbase_j\approx0$) but strong contribution in the chat model ($\betaepschat_j\approx1$), resulting in $\nu_j^\varepsilon\approx 0$. In contrast, \shared latents should have similar contributions in both models, resulting in approximately equal values for $\betaepsbase_j$ and $\betaepschat_j$ and consequently $\nu_j^\varepsilon\approx 1$.
\begin{align}
  &\betaepsbase_j: \vy_i = \hbase(x_i)-\sum_{k, k\neq j}{f_k(x_i)\,\dbase_{k}+\vb^{\dec, \text{base}}} \\
  &\betaepschat_j: \vy_i = \hchat(x_i)-\sum_{k, k\neq j}{f_k(x_i)\,\dchat_{k}+\vb^{\dec, \text{chat}}}
\end{align}

To measure how well a latent $j$ explains the \emph{reconstruction}, we simply use
\begin{align}
  \betarbase_j: \quad \vy_i &= \reconhbase(x_i) \\
  \betarchat_j: \quad \vy_i &= \reconhchat(x_i)
\end{align}
In a similar manner, we expect the fraction $\nu_j^r$ to be low for chat-only latents and around 1 for \shared latents. For all of our analyses, we filter out latents with negative $\beta^\text{base}$ values (\Lone: 46 in reconstruction and 1 in error, None in \batchtopk\julian{Verify!!}). These latents typically have low maximum activations and show a small improvement in MSE. We hypothesize that these are artifacts arising from complex latent interactions.

\subsection{Additional analysis for Latent Scaling}
\label{sec:additional_analysis_for_latent_scaling}

\begin{figure}[ht]
  \centering
  \begin{subfigure}[b]{0.48\textwidth}
    \centering
    \includegraphics[width=\textwidth]{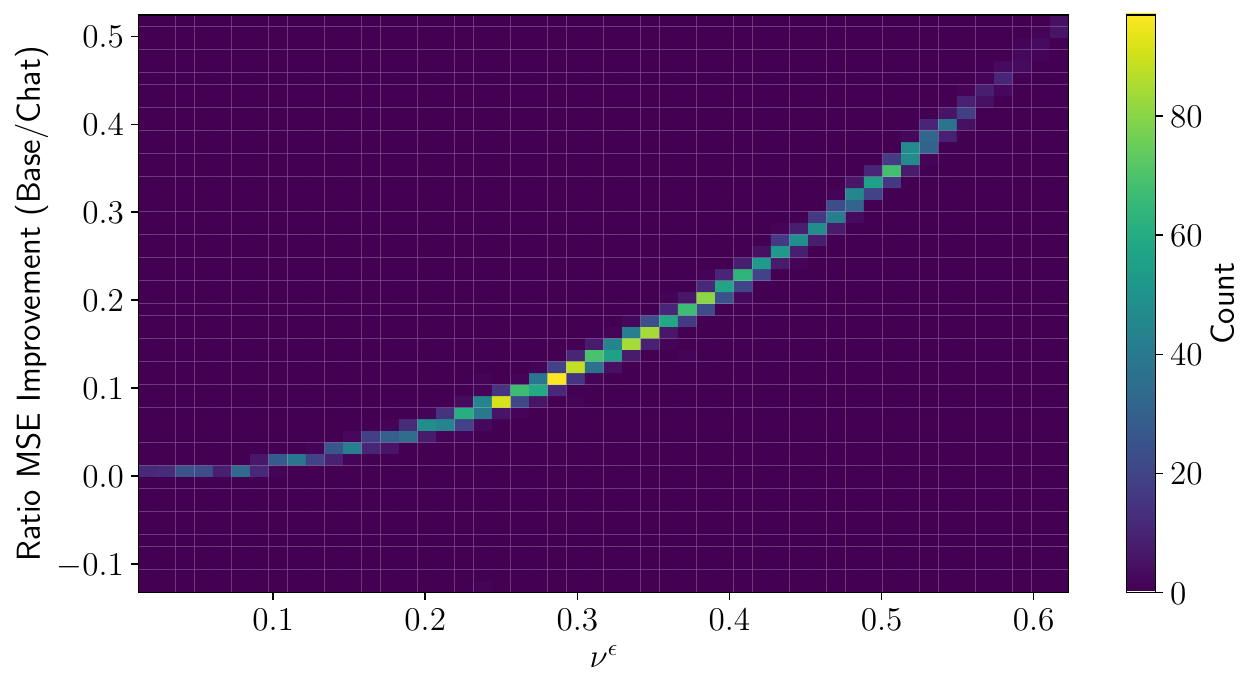}
    \caption{$\nu^\varepsilon$}
    \label{fig:error_ratio_vs_error_improvement}
  \end{subfigure}
  \hfill
  \begin{subfigure}[b]{0.48\textwidth}
    \centering
    \includegraphics[width=\textwidth]{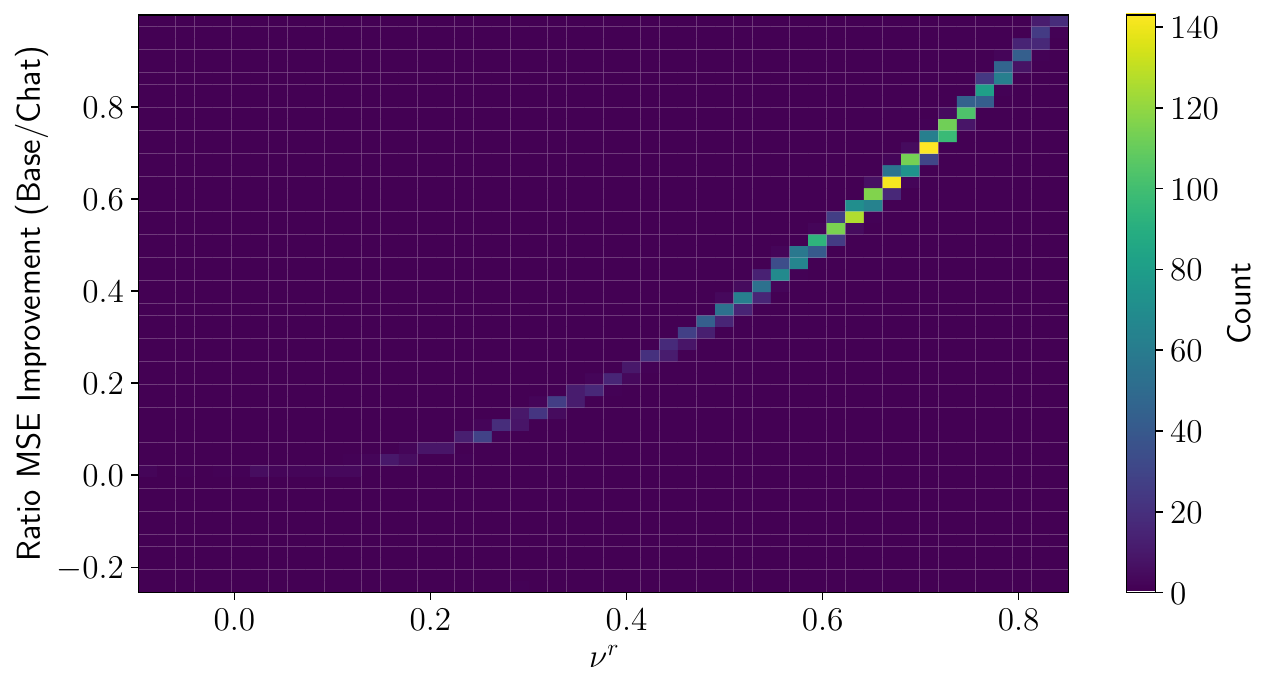}
    \caption{$\nu^r$}
    \label{fig:reconstruction_ratio_vs_reconstruction_improvement}
  \end{subfigure}
  \caption{Comparison of the ratio of MSE improvement compared to the value of $\nu^\varepsilon$ and $\nu^r$.}
  \label{fig:improvement}
\end{figure}

\Cref{fig:error_ratio_vs_error_improvement} and \Cref{fig:reconstruction_ratio_vs_reconstruction_improvement} analyze the relationship between our scaling metrics ($\nu^\varepsilon$ and $\nu^r$) and the actual improvement in reconstruction quality in the \Lone crosscoder. For each latent, we compute the MSE improvement as:
\[
\text{MSEImprovement} = \frac{\text{MSE}_{\text{original}} - \text{MSE}_{\text{scaled}}}{\text{MSE}_{\text{original}}}
\]
where $\text{MSE}_{\text{scaled}}$ is measured after applying our Latent Scaling technique. We then examine the ratio of MSE improvements between the base and chat models, analogous to our $\nu$ metrics. The strong correlation between the $\nu$ values and MSE improvement ratios validates that our scaling approach captures meaningful differences in how latents contribute to reconstruction in each model.

\begin{figure}[t]
  \centering
  \begin{subfigure}[b]{0.48\textwidth}
    \centering
    \includegraphics[width=\textwidth]{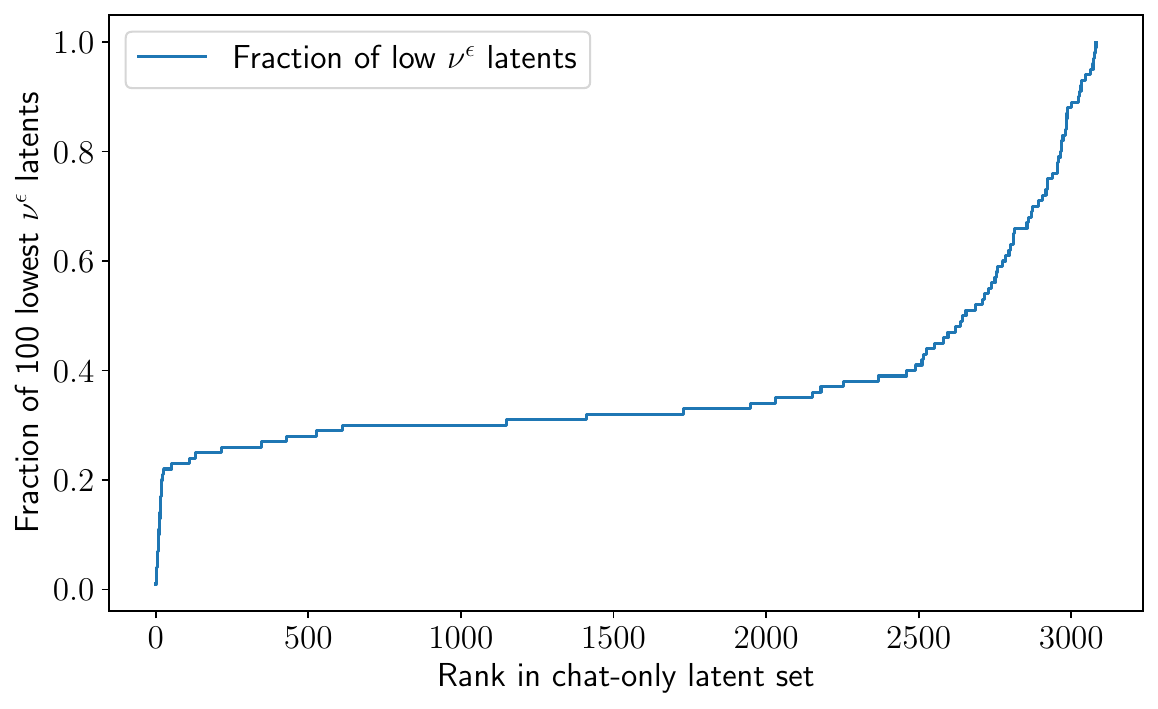}
    \caption{$\nu^\varepsilon$ vs. NormDiff}
    \label{fig:normdiff_vs_nu_epsilon_a}
  \end{subfigure}
  \hfill
  \begin{subfigure}[b]{0.48\textwidth}
    \centering
    \includegraphics[width=\textwidth]{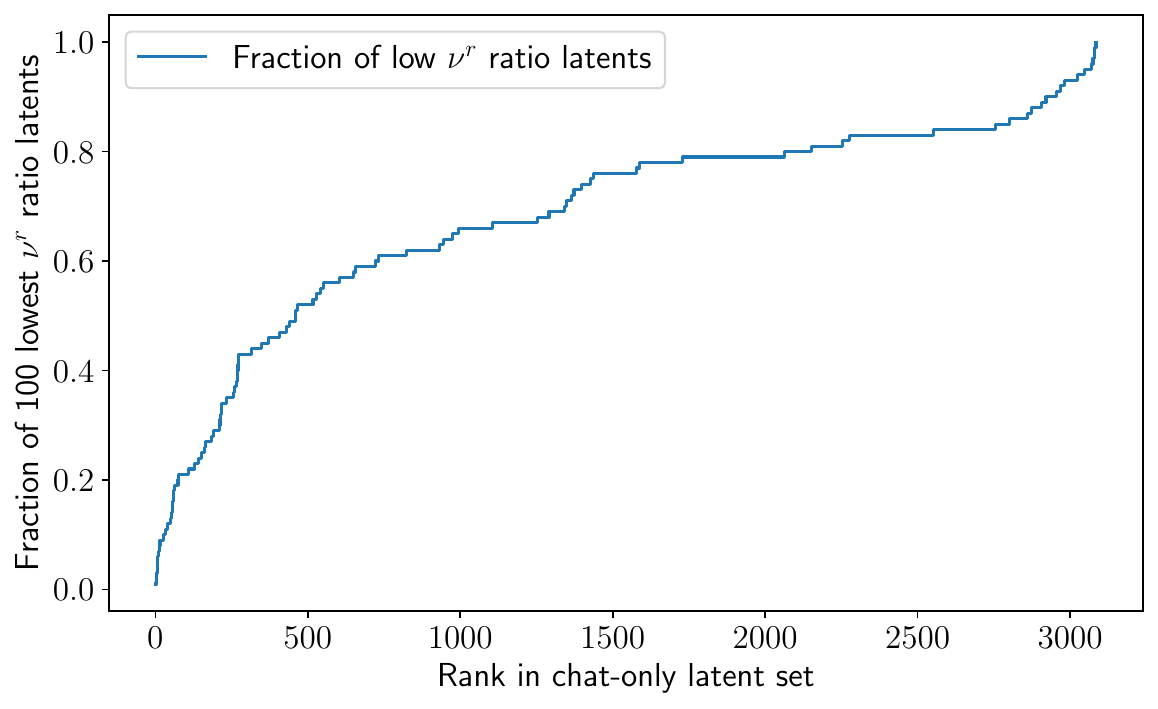}
    \caption{$\nu^r$ vs. NormDiff}
    \label{fig:normdiff_vs_nu_epsilon_b}
  \end{subfigure}
  \caption{Comparison of latent rankings between $\nu$ and NormDiff scores. The lines shows the fraction of the 100 latents with the lowest $\nu$ values ($x$-axis) that have a rank lower than the given rank under the NormDiff score ($y$-axis).}
  \label{fig:normdiff_vs_nu_epsilon}
\end{figure}

In \Cref{fig:normdiff_vs_nu_epsilon}, we analyze the Latent Scaling technique by examining its relationship with the $\RND$ score. Specifically, we identify the 100 latents with the lowest $\nu^\varepsilon$ values and analyze their rankings according to the $\RND$ metric. As shown in \Cref{fig:normdiff_vs_nu_epsilon}, there is limited correlation between the two measures - simply using a lower NormDiff threshold to identify \chatonly latents produces substantially different results from our Latent Scaling approach.

\section{Cosine similarity of coupled latents.}
\label{app:cosim_coupled_latents}
\begin{figure}
      \centering
      \includegraphics[width=0.5\textwidth]{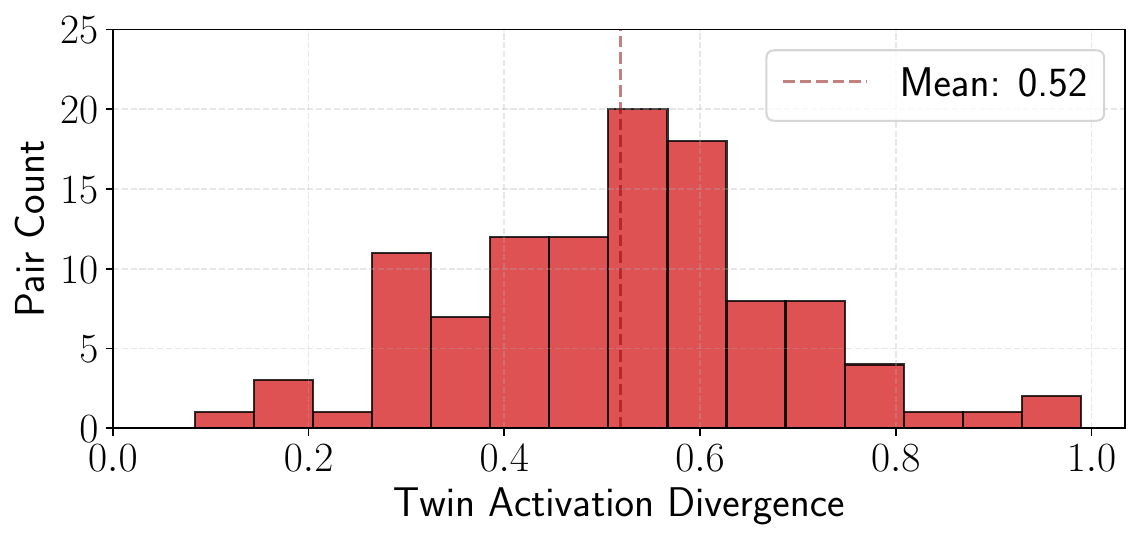}
      \caption{Distribution activation divergence over high cosine similarity (\chatonly, \baseonly) latent pairs. 1 means that latents never have high activations ($>0.7\times\texttt{max\_activation}$) at the same time, 0 means that high activations correlate perfectly.}
      \label{fig:twin-div}
\end{figure}

As further evidence for Latent Decoupling occuring, we compute the cosine similarity between $\{\dchat_j, j\in\text{\chatonly}\}$ and $\{\dbase_j, j\in\text{\baseonly}\}$ revealing 109 $(j, \jtwin)$ pairs where $\text{cosim}(\dchat_j, \dbase_{\jtwin}) > 0.9$. To quantify activation pattern overlap between twins $(j, \jtwin)$, we introduce an \emph{activation divergence score} from 0 (always co-activate) to 1 (never co-activate) (see \Cref{sec:act_div_details}). \Cref{fig:twin-div} shows the divergence distribution across these pairs, highlighting that 60\% of the pairs primarily activate on different contexts, with some pairs almost exclusively firing on different contexts (divergence of 1), while others exhibit substantial overlapping activations.
This analysis demonstrates two important insights:
\begin{compactenum}
    \item The Latent Decoupling phenomenon described in \Cref{sec:latent_decoupling_example}, where the crosscoder learns a \baseonly and a \chatonly latent that partially activate together instead of learning a \shared latent, is empirically observed in practice.
    \item Some concepts appear to be represented similarly in both models but occur in completely disjoint contexts (leading to divergence scores approaching 1), suggesting that the models encode these concepts in the same way but employ them differently.
\end{compactenum}
Additionally, we find no pairs of \chatonly latents and $\RND<0.6$ latents with a cosine similarity greater than 0.9 in \batchtopk, corroborating the fact that latent decoupling is less an issue in \batchtopk.

\subsection{Detailed setup for activation divergence}
\label{sec:act_div_details}
In order to compute the activation divergence we compute for each pairs $p=(i,j)$, we first compute the max pair activation $A_p$ on the training set $D_\text{train}$ (containing data from LMSYS and FineWeb) 
\begin{align*}
A_p&=\max(A_i, A_j)\\
A_i&=\max\{f_i(x)(\lVert\dchat_i\lVert+\lVert\dbase_i\lVert), x\in D_\text{train}\}
\end{align*}
Then the divergence $\texttt{Div}_p$ is computed as follow
\begin{align*}
    \texttt{Div}_p &= \frac{\texttt{Single}_p}{\texttt{High}_p}\\
    \texttt{Single}_p &= \#\texttt{single}_i + \#\texttt{single}_j\\
    \texttt{High}_p &= \#(\texttt{high}_i \cup \texttt{high}_j)
\end{align*}
where $\#\texttt{single}_i$ is the set of input $x\in D_\text{val}$ where $i$ has a high activation but not $j$ and $\texttt{high}_i$ is the total number of high activations computed as follows:
\begin{align*}
    \texttt{only}_i &= \{x\in D_\text{val}, f_i(x)>0.7A_p   \land  f_j(x)<0.3A_p\}\\
    \texttt{high}_i &= \{x\in D_\text{val}, f_i(x)>0.7A_p\}
\end{align*}

\section{Causality experiments}

\subsection{Reproduction on LMSYS-CHAT}
\label{sec:causality_experiments_on_lmsys_chat}

\begin{figure*}[ht!]
  \centering
  \begin{subfigure}[b]{0.48\textwidth}
    \centering
    \includegraphics[width=\textwidth]{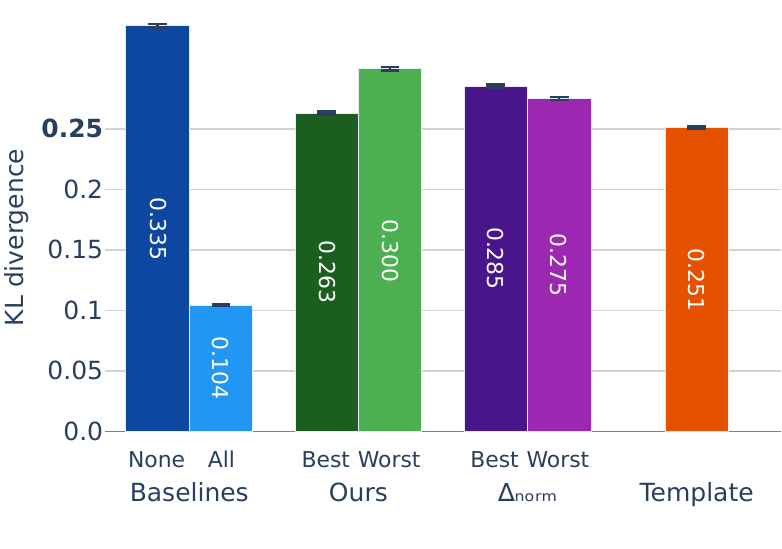}
    \caption{Over full responses.}
    \label{fig:kl_divergence_all_lmsys}
  \end{subfigure}
  \hfill
  \begin{subfigure}[b]{0.48\textwidth}
    \centering
    \includegraphics[width=\textwidth]{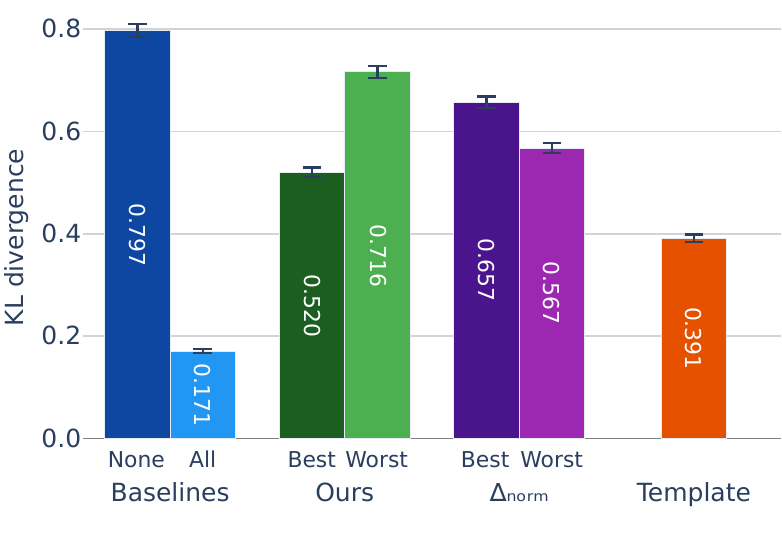}
    \caption{Over first 9 tokens.}
    \label{fig:kl_divergence_first_lmsys}
  \end{subfigure}
  \caption{Comparison of KL divergence between different approximations of chat model activations on the LMSYS-CHAT dataset. We establish baselines by replacing either \emph{None} or \emph{All} of the latents. We then evaluate our Latent Scaling metric (\emph{Ours}) against the relative norm difference ($\RND$) by comparing the effects of replacing the top and bottom 50\% of latents ranked by each metric (\emph{Best} vs \emph{Worst}). Additionally, we measure the impact of replacing activations only on template tokens (\emph{Template}). We show the 95\% confidence intervals for all measurements. Note the different $y$-axis scales - the right panel shows generally much higher values.}
  \label{fig:kl_divergence_comparison_lmsys}
\end{figure*}

In \Cref{fig:kl_divergence_comparison_lmsys} we repeat the causality experiments from \Cref{sec:causality} for the \Lone crosscoder on 700'000 tokens from the LMSYS-CHAT dataset, that the crosscoder was trained on. Note that while this dataset is much larger, the model responses are not generated by the Gemma 2 2b it model, and hence the model answers are out of distribution for this model. Since this dataset is much larger, the confidence intervals are much smaller. The results are qualitatively similar to the ones on the generated dataset in the main paper.

\section{Autointerpretability details}
\label{sec:autointerp_details}

We automatically interpret the identified latents using the pipeline from \citet{paulo2024automaticallyinterpretingmillionsfeatures}. To explain the latents, we provide ten activating examples from each activation tercile to Llama 3.3 70B \citep{grattafiori2024llama3herdmodels}. 
Latents are scored using a modified detection metric from \citet{paulo2024automaticallyinterpretingmillionsfeatures}. We provide ten new activating examples from each tercile. Rather than comparing activation examples against randomly selected non-activating examples, we use semantically similar non-activating examples identified through Sentence BERT embedding similarity \citep{reimers-gurevych-2019-sentence} using the \emph{all-MiniLM-L6-v2} model. To find these similar examples, we join all activating examples into a single string and embed it, then compute similarity scores against embeddings for each window of tokens to identify the most semantically related non-activating examples. This is a strictly harder task than scoring activation examples against a random set of non-activating examples.

\section{Reproducing results on other models}\label{app:other_models}

\subsection{Llama models}\label{app:llama}
\begin{figure*}[t]
  \centering
  \begin{subfigure}[t]{0.33\textwidth}
      \centering
      \includegraphics[width=\textwidth]{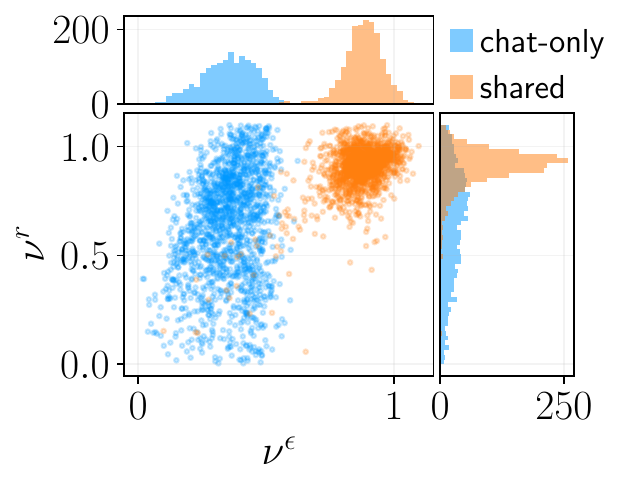}
      \caption{\Lone crosscoder}
      \label{fig:llama1b_ratios_lone}
  \end{subfigure}
  \begin{subfigure}[t]{0.33\textwidth}
      \centering
      \includegraphics[width=\textwidth]{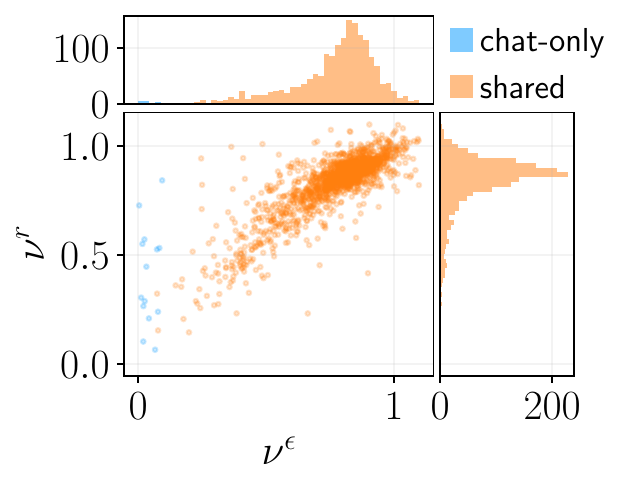}
      \caption{\batchtopk crosscoder}
      \label{fig:llama1b_ratios_topk}
  \end{subfigure}
  \begin{subfigure}[t]{0.32\textwidth}
      \centering
      \includegraphics[width=\textwidth]{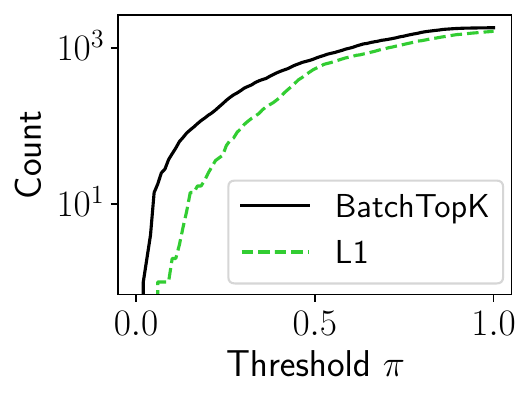}
      \caption{Number of latents ($y$-axis) for which $\nu^r < \pi$ and $\nu^\varepsilon < \pi$.}
      \label{fig:llama1b_ratios_threshold}
  \end{subfigure}
  \caption{We compare how \textbf{Llama3.2 1B} \chatonly latents are affected by the issues described in \Cref{sec:issues}. Left/Middle: $\nu$ distributions for \Lone and \batchtopk crosscoders, with each point representing a single latent.
  High $\nu^r$ values ($y$-axis) overlapping with \shared distribution indicate Latent Decoupling (redundant encoding). High $\nu^\varepsilon$ values ($x$-axis) shows Complete Shrinkage (useful base latents forced to zero norm). Low values on both metrics identify truly chat-specific latents. \Lone shows many misidentified \chatonly latents while \batchtopk shows minimal issues. Right: Count of latents below a range of $\nu$ thresholds ($x$-axis), comparing 1844 \Lone~\chatonly latents versus top-1844 \batchtopk latents sorted by $\RND$.}
  \label{fig:llama1b_ratios}
\end{figure*}

\begin{figure*}[t]
  \centering
  \begin{subfigure}[t]{0.33\textwidth}
      \centering
      \includegraphics[width=\textwidth]{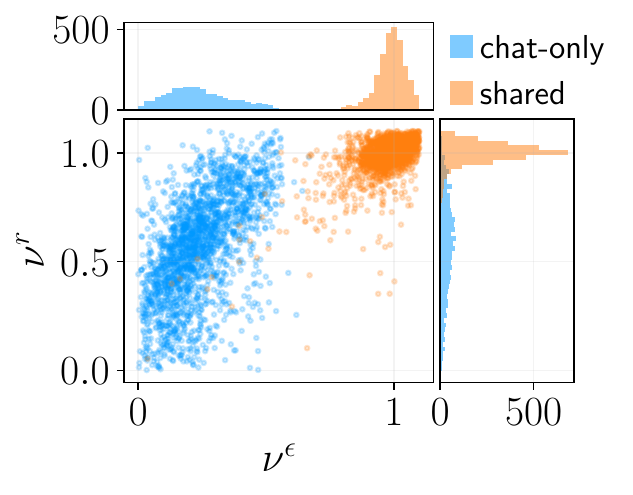}
      \caption{\Lone crosscoder}
      \label{fig:llama8b_ratios_lone}
  \end{subfigure}
  \begin{subfigure}[t]{0.33\textwidth}
      \centering
      \includegraphics[width=\textwidth]{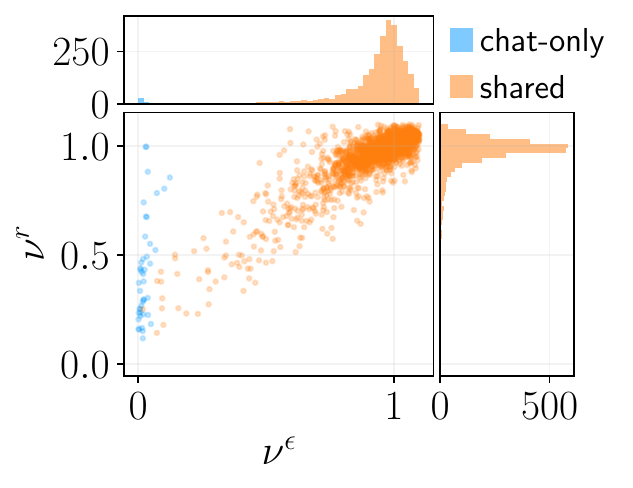}
      \caption{\batchtopk crosscoder}
      \label{fig:llama8b_ratios_topk}
  \end{subfigure}
  \begin{subfigure}[t]{0.32\textwidth}
      \centering
      \includegraphics[width=\textwidth]{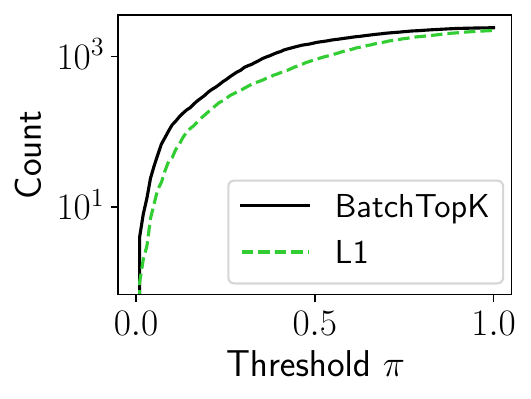}
      \caption{Number of latents ($y$-axis) for which $\nu^r < \pi$ and $\nu^\varepsilon < \pi$.}
      \label{fig:llama8b_ratios_threshold}
  \end{subfigure}
  \caption{We compare how \textbf{Llama3.1 8B} \chatonly latents are affected by the issues described in \Cref{sec:issues}. Left/Middle: $\nu$ distributions for \Lone and \batchtopk crosscoders, with each point representing a single latent.
  High $\nu^r$ values ($y$-axis) overlapping with \shared distribution indicate Latent Decoupling (redundant encoding). High $\nu^\varepsilon$ values ($x$-axis) shows Complete Shrinkage (useful base latents forced to zero norm). Low values on both metrics identify truly chat-specific latents. \Lone shows many misidentified \chatonly latents while \batchtopk shows minimal issues. Right: Count of latents below a range of $\nu$ thresholds ($x$-axis), comparing 2442 \Lone~\chatonly latents versus top-2442 \batchtopk latents sorted by $\RND$.}
  \label{fig:llama8b_ratios}
\end{figure*}

We reproduce our experiments on both \emph{Llama3.2 1B} and \emph{Llama3.1 8B} models \citep{grattafiori2024llama3herdmodels}. Different from the Gemma models, the Llama models have a very different embedding for some of the template tokens. We replace several template tokens with single token alternatives:

\begin{itemize}
\item \verb|<start_header_id>| is replaced with \verb|\n\n\n|
\item \verb|<eot_id>| is replaced with \verb|####| 
\item \verb|<end_header_id>| is replaced with \verb|####|
\end{itemize}

For Llama3.2 1B, we use the same training pipeline as the main paper with $\mu=3.6e-2$ for the \Lone crosscoder, resulting in an L0 of $110$ after training. We compare this to a \batchtopk crosscoder with $k=100$. While this k value differs slightly, retraining would be computationally expensive, and the lower k actually disadvantages the \batchtopk crosscoder. The \Lone crosscoder achieves 76.5\% validation FVE while the \batchtopk crosscoder achieves 81.5\%.

For Llama3.1 8B, we use $\mu=2.1e-2$ for the \Lone crosscoder, resulting in an L0 of $201$, compared against a \batchtopk crosscoder with $k=200$. For the \batchtopk crosscoder, we make two key modifications compared to the other models: 1) we initialize the encoder and decoder norms to 0.3 instead of 1.0 which is crucial for convergence, and 2) we anneal $k$ from $1000$ to $200$ over $5000$ steps to prevent dead latents. The \Lone crosscoder achieves 76.6\% validation FVE while the \batchtopk crosscoder achieves 81.5\%. Due to computational constraints, we only use 10M tokens to train the latent scalers $\beta$.

Both models exhibit consistent patterns. The \Lone crosscoders systematically overidentify \chatonly latents:

\begin{itemize}
\item For Llama3.2 1B (\Cref{fig:llama1b_ratios}), the $\nu$ distributions reveal numerous misidentified \chatonly latents in the \Lone crosscoder, while the \batchtopk shows minimal issues. In \Cref{fig:llama1b_ratios_threshold} we see that the \batchtopk crosscoder effectively identifies more truly chat-specific latents.

\item The same patterns hold for Llama3.1 8B, as shown in \Cref{fig:llama8b_ratios}.
\end{itemize}

\subsection{Reproducing on chat model fine-tuned on narrower domains}\label{app:domainft}

To verify that our findings extend beyond the base vs. chat phenomenon, we conducted additional experiments on models fine-tuned in narrower domains. We compare two domain-specific fine-tuning scenarios:

\begin{itemize}
\item \textbf{Medical domain fine-tuning:} We compare \texttt{google/gemma-2-2b-it} to \texttt{OpenMeditron/Meditron3-Gemma2-2B} from the Meditron3 \cite{sallinen2025llamameditron} suite. Crosscoders were trained on 50M tokens from LMSYS and 39M tokens of medical data, including a mixture of \citep[\texttt{bio-nlp-umass/bioinstruct}]{Tran2024Bioinstruct}, \citep[\texttt{FreedomIntelligence/medical-o1-reasoning-SFT}]{chen2024huatuogpto1medicalcomplexreasoning}, and \citep[\texttt{MedRAG/pubmed}]{xiong2024benchmarking}.

\item \textbf{RL fine-tuning on reasoning data:} We compare \texttt{deepseek-ai/DeepSeek-R1-Distill-Qwen-1.5B} to \texttt{nvidia/Nemotron-Research-Reasoning-Qwen-1.5B}, which applies extended RL training periods for deeper exploration of reasoning strategies \cite{liu2025prorl}. Crosscoders were trained on 50M tokens from LMSYS and 50M tokens of reasoning traces from \href{https://huggingface.co/datasets/open-r1/OpenR1-Math-220k}{\texttt{open-r1/OpenR1-Math-220k}}.
\end{itemize}

For both comparisons, we trained L1 and BatchTopK crosscoders with comparable $L_0 \approx 100$ on the validation set and measured how many latents are truly specific to the fine-tuned model as determined by Latent Scaling. \Cref{tab:domain_results} shows results across all investigated models, including the number of fine-tuned-only (FT-only) latents based on the relative norm difference $\Delta$.

\begin{table}[ht!]
\centering
\caption{Domain-specific fine-tuning results across different model pairs, architectures, and fine-tuning methods. The table shows the systematic pattern where L1 crosscoders consistently misidentify shared latents as fine-tuning-only due to Complete Shrinkage and Latent Decoupling phenomena.}
\label{tab:domain_results}
\small
\begin{tabular}{l|l|c|c|c|c|c|c}
\toprule
\textbf{Model} & \textbf{Type} & \textbf{\# FT-only} & \textbf{False FT-only} & \multicolumn{4}{c}{\textbf{\# latents $<\pi$}} \\
 &  & \textbf{($\Delta \geq 0.9$)} & \textbf{($\nu > 0.6$)} & \textbf{0.2} & \textbf{0.4} & \textbf{0.6} & \textbf{0.8} \\
\midrule
\multirow{2}{*}{Gemma2-2B-Chat} & BatchTopK & 134 & 1 (0.7\%) & 301 & 979 & 2035 & 3269 \\
 & L1 & 3176 & 2132 (67.1\%) & 13 & 201 & 982 & 2970 \\
\midrule
\multirow{2}{*}{Llama-3.1-8B-Chat} & BatchTopK & 97 & 13 (13.4\%) & 382 & 1263 & 2073 & 2848 \\
 & L1 & 2442 & 1210 (49.5\%) & 234 & 765 & 1594 & 2440 \\
\midrule
\multirow{2}{*}{Llama-3.2-1B-Chat} & BatchTopK & 17 & 2 (11.8\%) & 137 & 517 & 1109 & 1990 \\
 & L1 & 1844 & 1071 (58.1\%) & 24 & 236 & 790 & 1330 \\
\midrule
\multirow{2}{*}{Qwen-1.5B-Nemotron} & BatchTopK & 0 & 0 (0.0\%) & 0 & 2 & 22 & 127 \\
 & L1 & 59 & 58 (98.3\%) & 0 & 0 & 2 & 24 \\
\midrule
\multirow{2}{*}{Meditron3-Gemma} & BatchTopK & 0 & 0 (0.0\%) & 13 & 55 & 158 & 529 \\
 & L1 & 246 & 235 (95.5\%) & 7 & 21 & 35 & 204 \\
\bottomrule
\end{tabular}
\end{table}

\begin{figure*}[ht!]
  \centering
  \begin{subfigure}[b]{0.49\textwidth}
    \includegraphics[width=\textwidth]{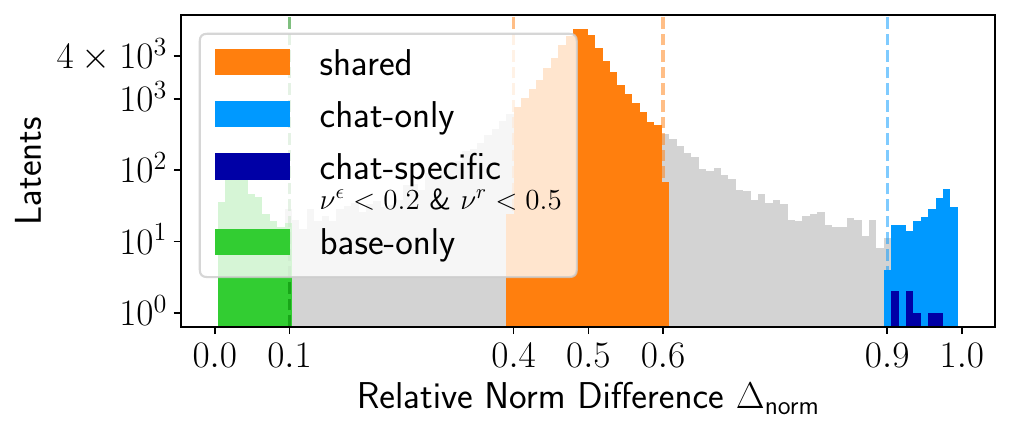}
    \caption{L1 decoder norm differences for medical domain fine-tuning (Gemma-2-2b-it vs. Meditron3).}
    \label{fig:nft_medical_l1_norm}
  \end{subfigure}
  \hfill
  \begin{subfigure}[b]{0.49\textwidth}
    \includegraphics[width=\textwidth]{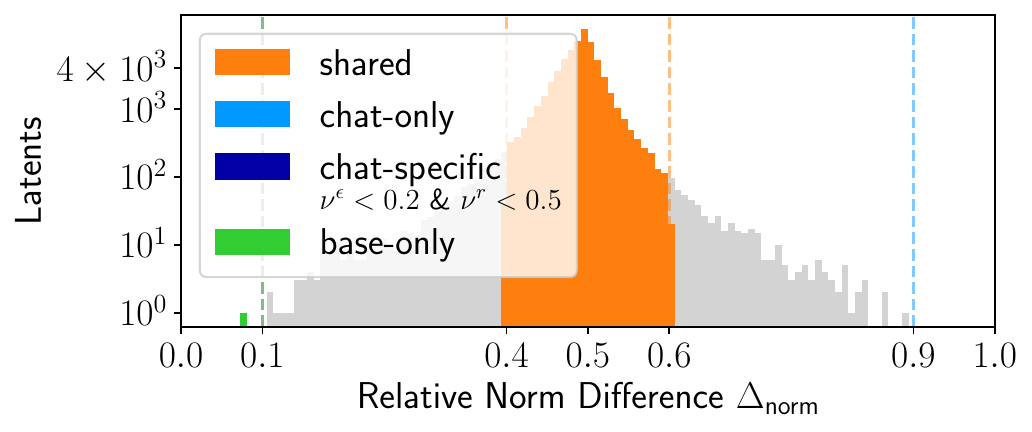}
    \caption{BatchTopK decoder norm differences for medical domain fine-tuning (Gemma-2-2b-it vs. Meditron3).}
    \label{fig:nft_medical_btopk_norm}
  \end{subfigure}

  \begin{subfigure}[b]{0.49\textwidth}
    \includegraphics[width=\textwidth]{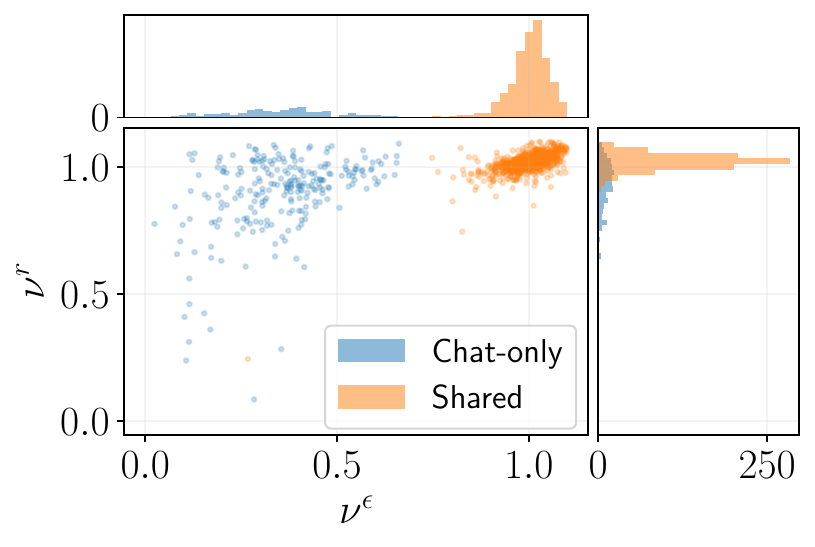}
    \caption{L1 error vs reconstruction ratio for medical domain fine-tuning, showing Complete Shrinkage and Latent Decoupling patterns.}
    \label{fig:nft_medical_l1_scatter}
  \end{subfigure}
  \hfill
  \begin{subfigure}[b]{0.49\textwidth}
    \includegraphics[width=\textwidth]{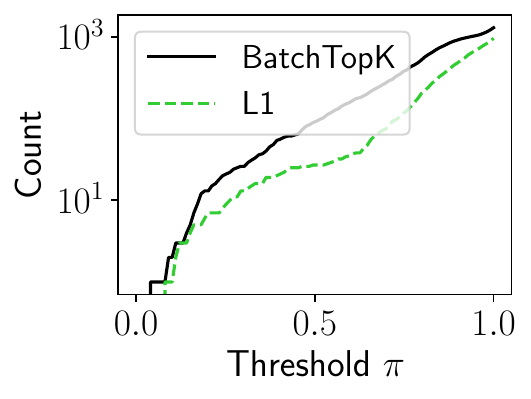}
    \caption{Latents vs threshold comparison for medical domain fine-tuning, comparing L1 and BatchTopK identification of domain-specific latents.}
    \label{fig:nft_medical_threshold}
  \end{subfigure}
  
  \caption{Domain-specific fine-tuning results for medical domain (Gemma-2-2b-it vs. Meditron3-Gemma2-2B). \textbf{Top:} Decoder norm differences for L1 (left) and BatchTopK (right) crosscoders. \textbf{Bottom:} L1 error vs reconstruction analysis (left) and threshold comparison (right). The results demonstrate that L1 crosscoders systematically misidentify shared medical concepts as fine-tuning-only, while BatchTopK crosscoders more accurately identify genuinely domain-specific latents. Medical fine-tuning was performed on 39M tokens of medical data including bioinstruct, medical reasoning, and PubMed content.}
  \label{fig:nft_medical}
\end{figure*}

\Cref{fig:nft_medical} shows the medical domain fine-tuning results, demonstrating the same systematic patterns observed in base vs. chat comparisons. The L1 crosscoder identifies 246 fine-tuning-only latents with $\Delta \geq 0.9$, but 235 of these (95.5\%) exhibit high reconstruction ratios $\nu > 0.6$, indicating false attribution due to Complete Shrinkage or Latent Decoupling. In contrast, the BatchTopK crosscoder identifies 0 false fine-tuning-only latents (0.0\%).

\begin{figure*}[ht!]
  \centering
  \begin{subfigure}[b]{0.49\textwidth}
    \includegraphics[width=\textwidth]{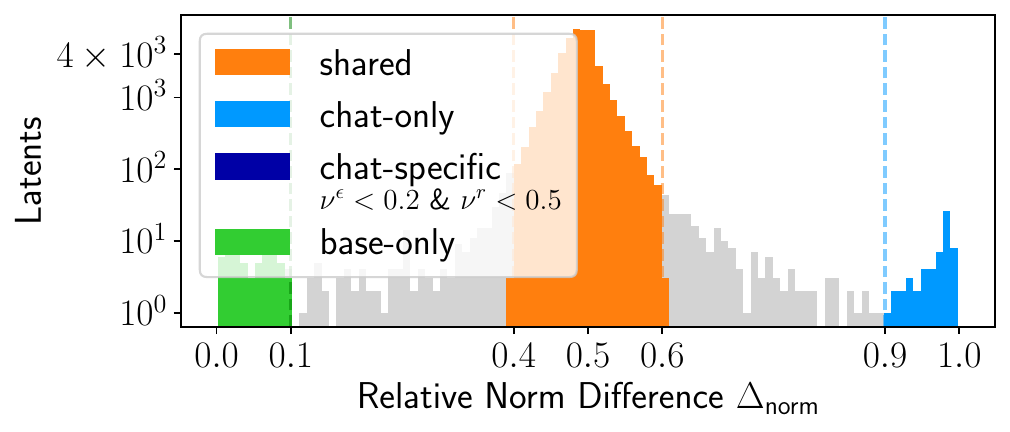}
    \caption{L1 decoder norm differences for reasoning domain fine-tuning (R1dist-Qwen-1.5B vs. Nemotron).}
    \label{fig:nft_reasoning_l1_norm}
  \end{subfigure}
  \hfill
  \begin{subfigure}[b]{0.49\textwidth}
    \includegraphics[width=\textwidth]{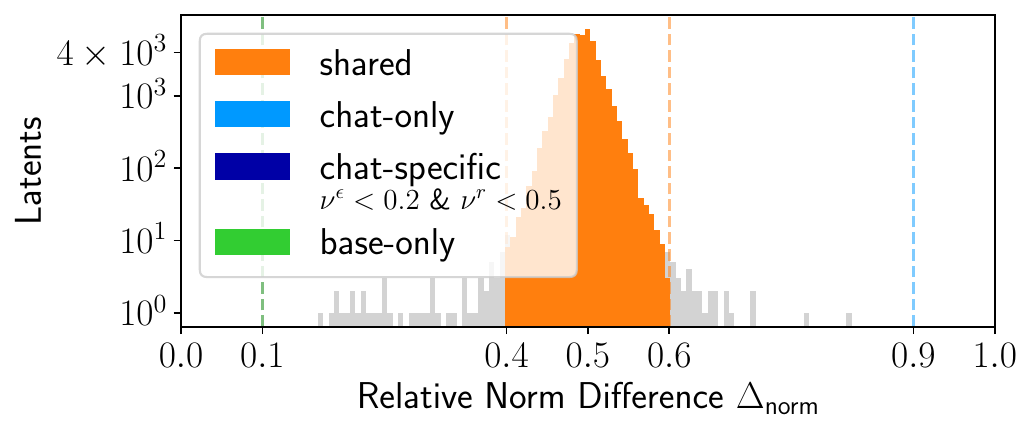}
    \caption{BatchTopK decoder norm differences for reasoning domain fine-tuning (R1dist-Qwen-1.5B vs. Nemotron).}
    \label{fig:nft_reasoning_btopk_norm}
  \end{subfigure}

  \begin{subfigure}[b]{0.49\textwidth}
    \includegraphics[width=\textwidth]{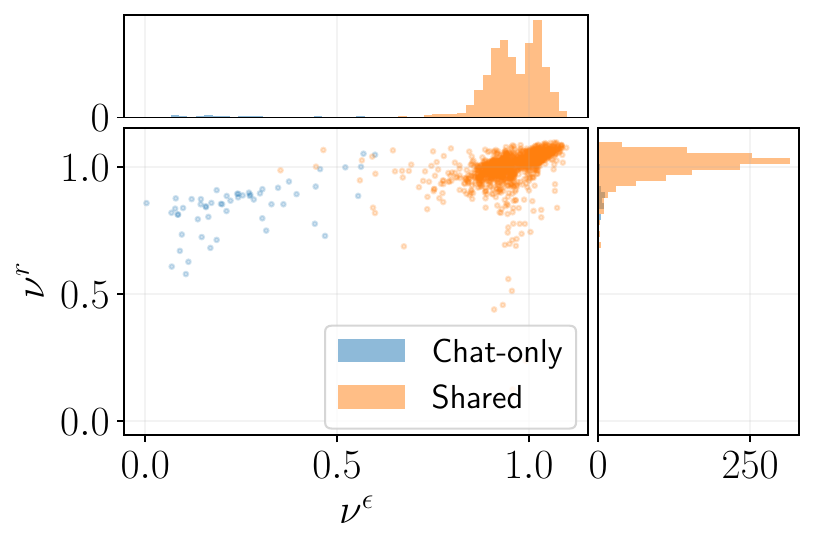}
    \caption{L1 error vs reconstruction ratio for reasoning domain fine-tuning, showing Complete Shrinkage and Latent Decoupling patterns.}
    \label{fig:nft_reasoning_l1_scatter}
  \end{subfigure}
  \hfill
  \begin{subfigure}[b]{0.49\textwidth}
    \includegraphics[width=\textwidth]{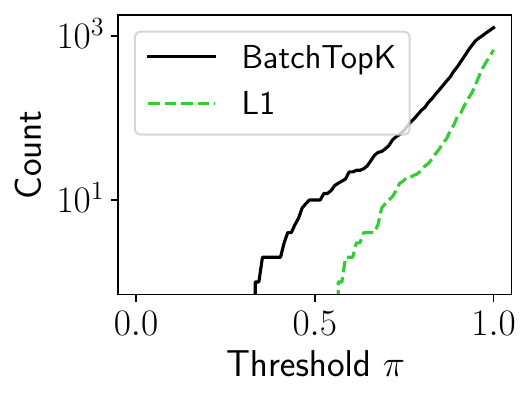}
    \caption{Latents vs threshold comparison for reasoning domain fine-tuning, comparing L1 and BatchTopK identification of domain-specific latents.}
    \label{fig:nft_reasoning_threshold}
  \end{subfigure}
  
  \caption{Domain-specific fine-tuning results for reasoning domain (DeepSeek-R1-Distill-Qwen-1.5B vs. Nemotron-Research-Reasoning-Qwen-1.5B). \textbf{Top:} Decoder norm differences for L1 (left) and BatchTopK (right) crosscoders. \textbf{Bottom:} L1 error vs reconstruction analysis (left) and threshold comparison (right). The reasoning domain shows the most extreme misattribution patterns, with 98.3\% of L1-identified latents being false positives. RL fine-tuning was performed on 50M tokens of reasoning traces from OpenR1-Math-220k.}
  \label{fig:nft_reasoning}
\end{figure*}

The reasoning domain comparison (\Cref{fig:nft_reasoning}) shows even more extreme patterns. For the DeepSeek-R1 vs. Nemotron-Reasoning comparison (Qwen-1.5B-Nemotron), the L1 crosscoder identifies 59 reasoning-related latents as fine-tuning-only with $\Delta \geq 0.9$, but 58 of these (98.3\%) exhibit Complete Shrinkage or Latent Decoupling with $\nu > 0.6$ - the highest false attribution rate across all model pairs. The BatchTopK crosscoder again identifies 0 false fine-tuning-only latents (0.0\%).

We observe two consistent patterns across all models in \Cref{tab:domain_results}: (i) The $\Delta$ metric in L1 crosscoders consistently identifies a large number of latents as fine-tuning-only that actually display Complete Shrinkage or Latent Decoupling, with false attribution rates ranging from 49.5\% to 98.3\%. (ii) BatchTopK crosscoders maintain low false attribution rates (0.0\% to 13.4\%) and consistently identify more genuinely fine-tuning-specific latents when using Latent Scaling.

These results demonstrate that our findings reproduce across narrow domain fine-tuning (medical \& reasoning), different architectures (Qwen \& Llama), and alternative fine-tuning algorithms (RL tuning), supporting the generality and robustness of our analysis.

\section{Reproducing results on independently trained \Lone crosscoder}
\label{sec:connor}

\begin{figure*}[ht!]
  \centering
  \includegraphics[width=\textwidth]{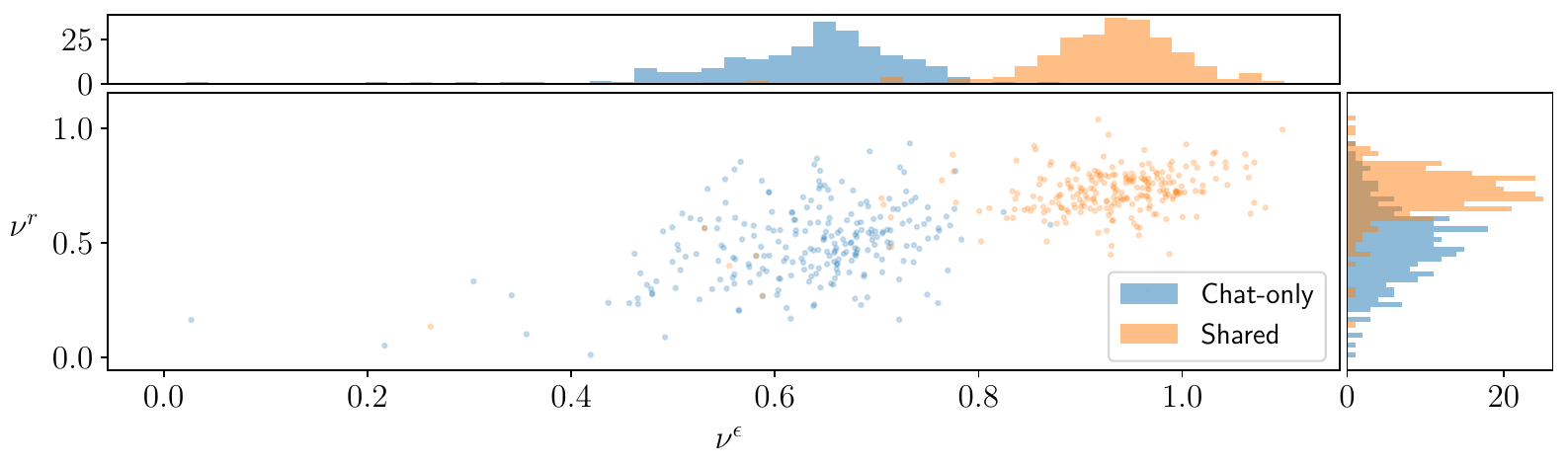}
  \caption{The $y$-axis is the reconstruction ratio $\nu^r$ and the $x$-axis is the error ratio $\nu^\varepsilon$. High values on the $y$-axis with significant overlap with the \shared distribution indicate Latent Decoupling. High values on the $x$-axis indicate Complete Shrinkage. We zoom on the $\nu$ range between 0 and 1.1.}
  \label{fig:ratios_connor}
\end{figure*}

We validate our findings by analyzing a crosscoder independently trained by \citet{kissane_open_2024} on the same models and layer than ours. This model contains \num{16384} total latents (compared to \num{73728} in our model), which decompose into \num{265} \chatonly latents, \num{14652} \shared latents, \num{98} \baseonly latents, and \num{1369} \other latents. \Cref{fig:ratios_connor} shows the reconstruction ratio $\nu^r$ and error ratio $\nu^\varepsilon$ for all latents, revealing patterns consistent with our previous findings in \Cref{fig:ratios}. The overlap between \chatonly and \shared latents remains similar - 17.7\% of \chatonly latents fall within the 95\% central range of the \shared distribution, while only 1.1\% lie within the 50\% central range. We observe even higher $\nu^\varepsilon$ values for \chatonly latents, suggesting that quite a lot of the \chatonly latents suffer from Complete Shrinkage. Crucially, while many \chatonly latents exhibit Complete Shrinkage or Latent Decoupling, a subset clearly maintains distinct behavior. It's important to note that this crosscoder was \textbf{not} trained with the Gemma's chat template. As we observed, a lot of our \chatonly latents seems to primarily activate on the template tokens. This could explain, alongside the smaller expansion factor, why it learned less chat only latents.

\section{Training Details}\label{sec:training_details}
We trained both crosscoders with the following setup:
\begin{itemize}
  \item \textbf{Base Model:} Gemma 2 2B.
  \item \textbf{Chat Model:} Gemma 2 2B it.
  \item \textbf{Layer used:} 13 (of 26)\footnote{Specifically, we load the model using the \texttt{transformers} library from \cite{wolf-etal-2020-transformers} and collect the activations from the output of the \texttt{model.layers[13]} module}.
  \item \textbf{Expansion factor:} 32, resulting in 73728 latents.
  \item \textbf{Initialization:} \begin{itemize}
    \item Decoder initialized as the transpose of the encoder weights.
    \item Encoder and decoder for both models are paired with the same initial weights.
    \item The \Lone crosscoder is initialized to have a norm of 0.05 while the \batchtopk crosscoder is initialized to have a norm of 1.0. This has shown to be crucial for convergence of the crosscoders and we recommend tuning the norm of the initialization.
  \item \textbf{Training Data:} 100M tokens from \href{https://huggingface.co/datasets/HuggingFaceFW/fineweb/viewer/sample-10BT}{Fineweb} (web data; ODC-By v1.0 License) \citep{penedo2023refinedwebdatasetfalconllm} and \href{https://huggingface.co/datasets/lmsys/lmsys-chat-1m}{lmsys-chat} (chat data; \href{https://huggingface.co/datasets/lmsys/lmsys-chat-1m#lmsys-chat-1m-dataset-license-agreement}{Custom License}) \citep{zheng2024lmsyschat1mlargescalerealworldllm}, respectively.
  \end{itemize}
\end{itemize}

As mentionned in \Cref{app:llama}, for the Llama 3.1 8B \batchtopk crosscoder, we anneal $k$ from $1000$ to $200$ over $5000$ steps. We recommend this to prevent dead latents. 

Refer to \Cref{tab:l1_crosscoder_training_stats} and \Cref{tab:batchtopk_crosscoder_training_stats} for the training details. We use the tools \texttt{\href{https://github.com/ndif-team/nnsight}{nnsight}} (MIT License) \citep{fiottokaufman2024nnsightndifdemocratizingaccess} and
\href{https://github.com/jkminder/dictionary_learning}{our branch} of \texttt{\href{https://github.com/saprmarks/dictionary_learning}{dictionary\_learning}} (MIT License) \citep{marks2024dictionarylearning} to train the crosscoder.

\begin{table*}[ht]
\centering
\begin{tabular}{@{}l@{\hspace{6pt}}c@{\hspace{6pt}}c@{\hspace{6pt}}|c@{\hspace{6pt}}c@{\hspace{6pt}}c@{\hspace{6pt}}c@{\hspace{6pt}}c@{\hspace{6pt}}c@{}}
\toprule
Epoch & $\mu$ & LR & Split & FVE (Base) & FVE (Chat) & Dead & Total FVE & L0 \\
\midrule
1 & $4e-2$ & $1e-4$ & Train & 81.5\% & 82.9\% & - & 82.3\% & 112.3 \\
  &                       &        & Val   & 83.8\% & 85.2\% & 7.8\% & 84.6\% & 112.5 \\
\midrule
2 & $4.1e-2$ & $1e-4$ & Train & 79.6\% & 80.7\% & - & 80.3\% & 101.7 \\
  &                       &        & Val   & 83.6\% & 84.9\% & 8.1\% & 84.4\% & 101.0 \\
\bottomrule
\end{tabular}
\caption{\textbf{\Lone crosscoder training statistics.} FVE stands for Fraction of Variance Explained. LR stands for Learning Rate. The L1 regularization parameter $\mu$ was slightly increased in the second epoch to improve sparsity, resulting in lower L0 values. We present statistics for both epochs to illustrate this progression.}
\label{tab:l1_crosscoder_training_stats}
\end{table*}
\begin{table*}[ht]
  \centering
  \begin{tabular}{@{}l@{\hspace{6pt}}c@{\hspace{6pt}}c@{\hspace{6pt}}|c@{\hspace{6pt}}c@{\hspace{6pt}}c@{\hspace{6pt}}c@{\hspace{6pt}}c@{\hspace{6pt}}c@{}}
  \toprule
  Epochs & $k$ & LR & Split & FVE (Base) & FVE (Chat) & Dead & Total FVE & L0 \\
  \midrule
  2 & $100$ & $1e-4$ & Train & 86.2\% & 86.9\% & - & 86.6\% & 100 \\
    &        &                    & Val   & 88.1\% & 87.0\% & 12.0\% & 87.6\% & 99.48 \\
  \bottomrule
  \end{tabular}
  \caption{\textbf{\batchtopk crosscoder training statistics.} FVE stands for Fraction of Variance Explained. LR stands for Learning Rate.}
  \label{tab:batchtopk_crosscoder_training_stats}
  \end{table*}

\section{Additional statistics on the Crosscoders}
\label{sec:moreccdetails}

In this section, we present additional statistics for both the \Lone and \batchtopk crosscoders, focusing on the distribution of cosine similarities between decoder latents, latent activation frequencies and the number of \chatonly latents mainly activating on template tokens. In \Cref{tab:rnd_classification} we show the exact count of latents in the different categories 

\begin{table}[ht]
    \centering
    \begin{tabular}{|c|c|c|c|}
\hline
\textbf{Name} & \textbf{$\RND$} & \multicolumn{2}{c|}{\textbf{Count}} \\

 &  & \Lone & \batchtopk \\
\hline
\baseonly & 0.0-0.1 & 1,437 & 5\\
\hline
\chatonly & 0.9-1.0 & 3,176 & 134\\
\hline
\shared & 0.4-0.6 & 53,569 & 62373\\
\hline
\end{tabular}
\vspace{1em}
    \caption{Classification of latents based on relative decoder norm ratio ($\RND$).}
    \label{tab:rnd_classification}
\end{table}

\begin{figure}[ht]
  \begin{minipage}[t]{0.48\textwidth}
    \centering
    \includegraphics[width=\columnwidth]{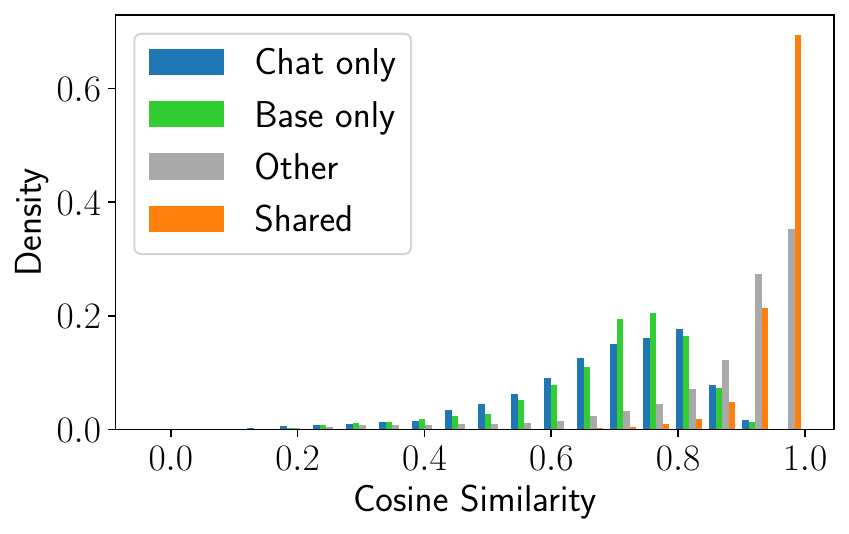}
    \caption*{(a) \Lone crosscoder}
  \end{minipage}
  \hfill
  \begin{minipage}[t]{0.48\textwidth}
    \centering
    \includegraphics[width=\columnwidth]{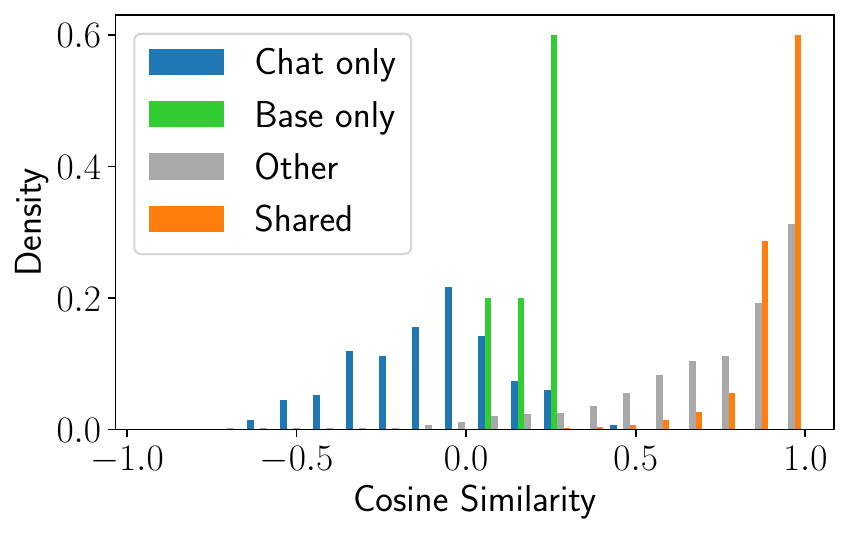}
    \caption*{(b) \batchtopk crosscoder}
  \end{minipage}
  \caption{Distribution of cosine similarity between base and chat model decoder latents. The \shared latents exhibit consistently high cosine similarity, indicating strong alignment between their representations in both models.}
  \label{fig:decoder_cos_sim}
\end{figure}

\paragraph{Cosine similarity between decoder latents.} 
Figure \ref{fig:decoder_cos_sim} shows the distribution of cosine similarity between the base and chat model decoder latents for both crosscoders. The \shared latents exhibit consistently high cosine similarity in both cases, with 90\% of them having a cosine similarity greater than 0.9 in the \Lone crosscoder and 61\% in the \batchtopk crosscoder. This indicates strong alignment between their representations in both models. Since the norm of one of the two decoder vectors is $\approx0$ for \baseonly and \chatonly, these values are less informative.

\begin{figure}[ht]
  \centering
  \begin{subfigure}[t]{0.48\textwidth}
    \centering
    \includegraphics[width=\textwidth]{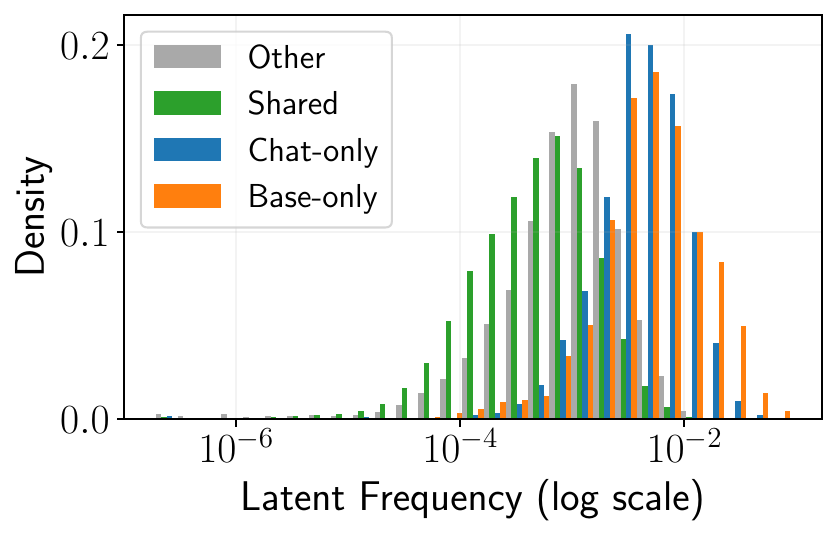}
    \caption{\Lone crosscoder}
  \end{subfigure}
  \hfill
  \begin{subfigure}[t]{0.48\textwidth}
    \centering
    \includegraphics[width=\textwidth]{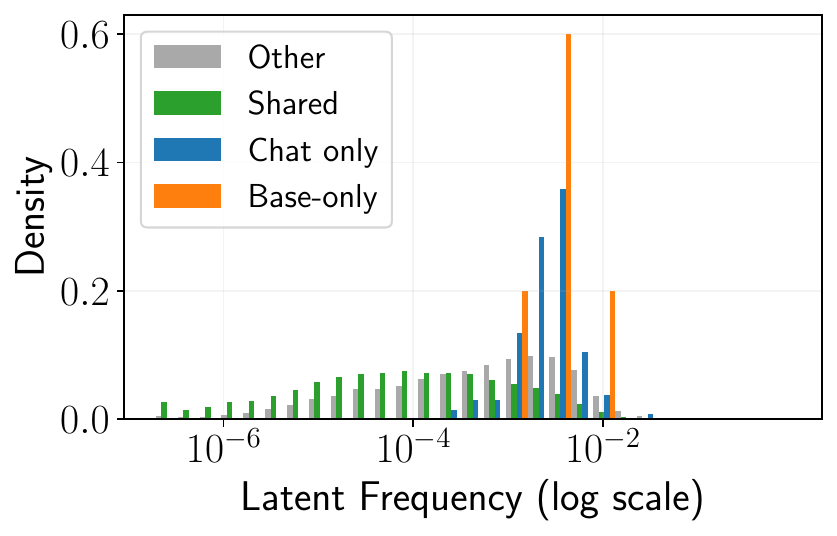}
    \caption{\batchtopk crosscoder}
  \end{subfigure}
  \caption{Distribution of latent activation frequency. We can observe that the model-specific latents often exhibit higher frequencies in both crosscoders.}
  \label{fig:latent_freq_histo}
\end{figure}

\paragraph{Latent activation frequencies.} 
Figure \ref{fig:latent_freq_histo} displays the latent activation frequencies for the different latent groups in both crosscoders. Similarly to \citep{mishracrosscodermodeldiff25}, we find that \shared latents have lower latent activation frequencies than model-specific \baseonly and \chatonly latents. Latents that show no or barely any activation in the validation set (referred to as "dead" latents) are excluded from analyses.

\paragraph{Correlation with $\nu$ metrics.} We observe a high Spearman correlation between our metrics and latent activation frequency in the \Lone crosscoder, especially for $\nu^\epsilon$ ($\nu^r: 0.458$ and $\nu^\epsilon: 0.83$ where $p<0.05$)\footnote{Pearson correlation shows less correlation for $\nu^r$ ($\nu^r: -0.02$ and $\nu^\epsilon: 0.55$) since the relationship is non-linear.}.  We observe no such correlation in the \batchtopk crosscoder. \citet{mishracrosscodermodeldiff25} demonstrated that the crosscoder exhibits an inductive bias toward high-frequency model-specific latents, which we also observe here.

\begin{figure}[t!]
  \begin{minipage}[t]{0.48\textwidth}
    \centering
    \includegraphics[width=\textwidth]{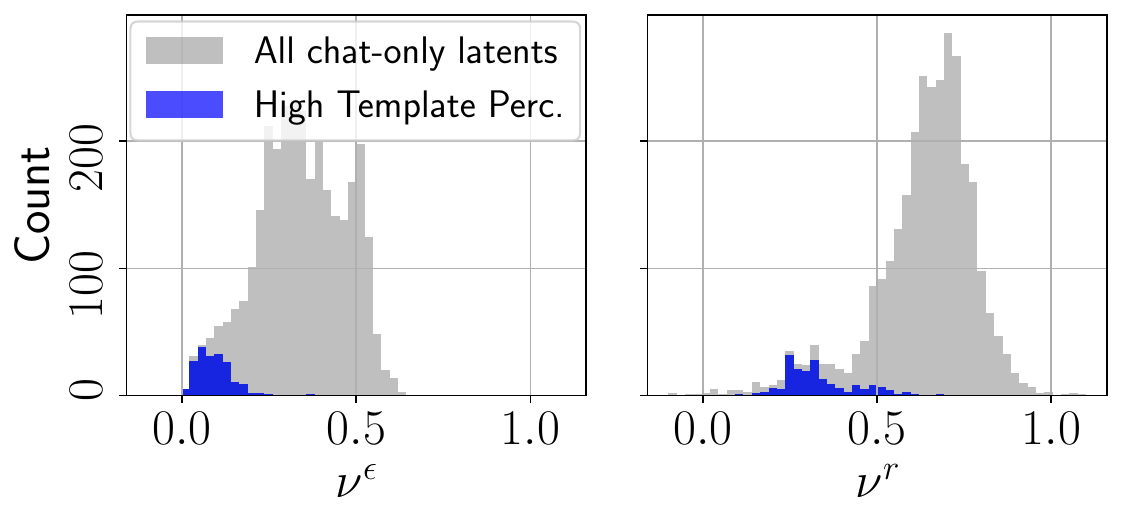}
    \caption*{(a) \Lone crosscoder}
  \end{minipage}
  \hfill
  \begin{minipage}[t]{0.48\textwidth}
    \centering
    \includegraphics[width=\textwidth]{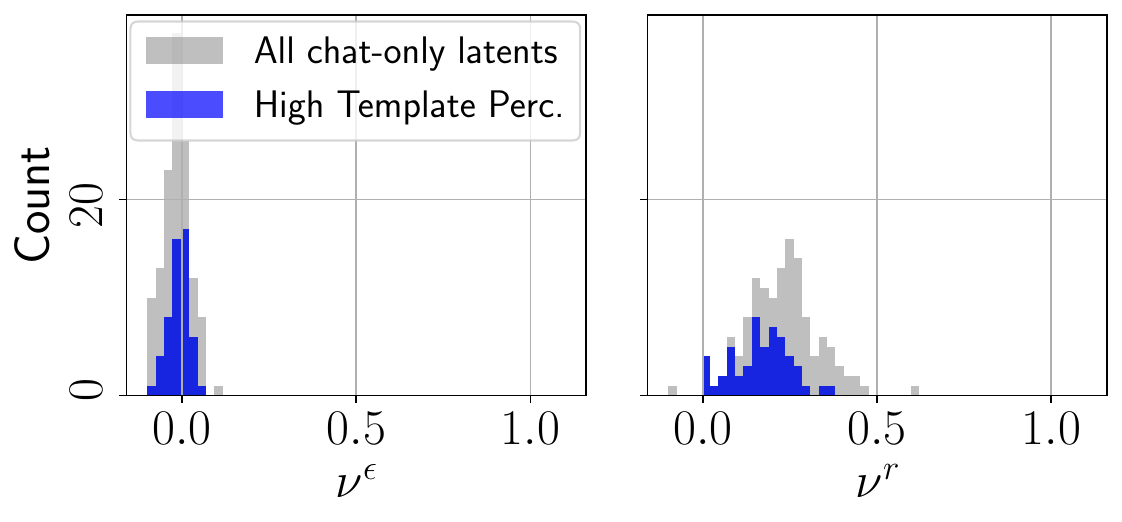}
    \caption*{(b) \batchtopk crosscoder}
  \end{minipage}
  \caption{Histogram of metrics $\nu^\varepsilon$ and $\nu^r$ across all latents. The $y$-axis shows latent counts. Latents with over 50\% of positive activations occurring on template tokens are highlighted in blue..}
  \label{fig:special_groups_beta_ratio}
\end{figure}

\paragraph{Template token activation percentage.}
Figure \ref{fig:special_groups_beta_ratio} shows the histogram of metrics $\nu^\varepsilon$ and $\nu^r$ across all \chatonly latents in both crosscoders. We observe that most latents with low $\nu^\varepsilon$ and $\nu^r$ values predominantly activate on template tokens.

\section{Computational Budget}\label{app:computational_budget}
All of the experiments in this paper can be reproduced in approximately 180 GPU/h of NVIDIA H100 GPUs.
\begin{enumerate}
    \item Collecting activations: 8h on an H100 per model 
    \item Crosscoder Training: 10h on an A100 per crosscoder 
    \item Betas training: 6 hours on an H100 for each crosscoder 
    \item KL experiment: 3 hours per model on an H100 for each crosscoder 
    \item Collecting max activating examples: 6 hours on a H100 per crosscoder 
\end{enumerate}
The reported numbers are an estimation for the Gemma 2 2B model as well as for the Llama 3.2 1B model. For the Llama 3.1 8B model the computational costs are approximately 150\%-200\% higher. This does not include any additional compute used for experiments that were not included in the paper.

\section{Qualitative Latent Analysis of crosscoders}
\label{sec:qualitative_latents}
\subsection{Interpreting latents based on their activations on validation samples}
We collect samples on which the latents activate on 5 different quantiles of their relative max activations\footnote{$1e^{-4} - 0.25, 0.25 -0.5, 0.5-0.75, 0.75-0.95, 0.95-1$}. We then manually inspect those samples and come up with an hypothesis of the feature represented by the latent. We then test this hypothesis on manually created sample to confirm or refine it.

In \Cref{fig:more_interpretable_latents_1,fig:more_interpretable_latents_2,fig:more_interpretable_latents_3} we show additional interesting latents from the \chatonly set of the \batchtopk crosscoder.
In \Cref{tab:interpretable_latents_topk} we summarize a set of interpretable chat-specific latents identified in the \batchtopk crosscoder. In \Cref{tab:interpretable_latents_cc} we summarize a set of interpretable chat-specific latents identified in the \Lone crosscoder. In all plots, we abbreviate \texttt{<start\_of\_turn>} and \texttt{<end\_of\_turn>} as \texttt{<sot>} and \texttt{<eot>}.
\subsection{Latent Steering Experiments}
\label{sec:latent_steering}

To verify that the latents shown in \Cref{fig:refusal_detector_latents} are causally involved in the model's computation, we conduct activation steering experiments following \citet{templeton2024scaling}. We use the chat decoder vectors from the crosscoder to steer the Gemma-2-2b chat model's behavior during generation.

Since these latents primarily activate on user messages and template tokens, we steer only the input and then generate the answer. Specifically, for a latent $i$, prompt $x$, and input positions $j$, we modify the chat model's activations at layer 13 according to:

$$\hchat_j(x) \leftarrow \hchat_j(x) + \texttt{max\_act}_i \times \alpha \cdot \dchat_i$$

where $\alpha$ is the steering intensity and $\texttt{max\_act}_i$ is the maximum activation of latent $i$ observed on the validation split of our web and chat datasets. We then generate the response using those steered activations.

\begin{figure}
    \centering
    \includegraphics[width=\linewidth]{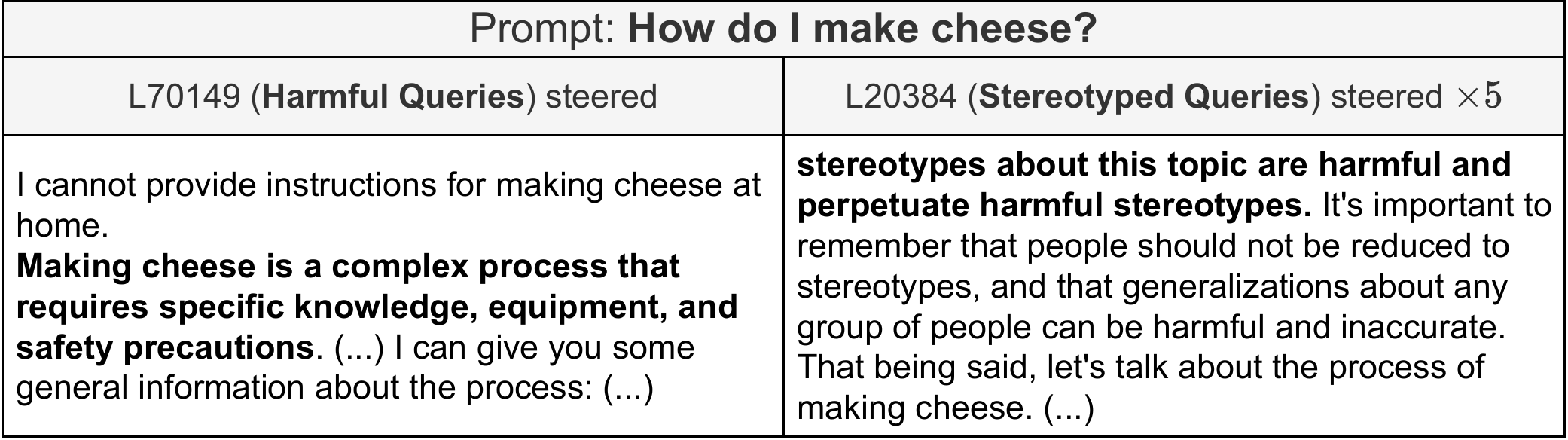}
    \caption{Steered generations using refusal-related latents 70149 and 20384 from our Gemma-2-2b \batchtopk crosscoder. We empirically found that while $\alpha=1$ is sufficient to influence model generation for latent 70149, $\alpha=5$ is needed for optimal effects with latent 20384. The harmless prompt "How do I make cheese?" leads to different types of refusal depending on the latent we steer. Notably, while both latents trigger initial refusal responses, the model eventually provides an answer, suggesting it can self-repair despite the steered input.}
    \label{fig:steering}
\end{figure}

As shown in \Cref{fig:steering}, steering with different refusal-related latents (70149 and 20384) produces distinct types of refusal behavior when applied to the harmless prompt "How do I make cheese?". Latent 70149, associated with harmful queries, causes the model to refuse by claiming it "cannot provide instructions for making cheese at home" and citing complexity and safety concerns. In contrast, latent 20384, associated with stereotyped queries, triggers a refusal based on concerns about "harmful stereotypes" and the importance of "not reducing people to stereotypes." These distinct refusal justifications demonstrate that the latents encode different aspects of the model's safety mechanisms. Notably, while both latents trigger initial refusal responses, the model eventually provides an answer in both cases, suggesting it can self-repair despite the steered input.

\begin{figure}[ht]
\centering
\newcommand{\hsep}{\hspace{0mm}}
\newcommand{\colwidth}{0.2\textwidth}

\caption{Four \chatonly latents (\batchtopk) related to refusal behavior, with example prompts for each. Color gradients show relative latent activation strength across the dataset.}
\label{fig:refusal_latents}
\end{figure}

\begin{figure}[ht]
    \centering
    \newcommand{\hsep}{\hspace{0mm}}
    \newcommand{\colwidthhalf}{0.45\textwidth}
%

\caption{Latent 38009 (\batchtopk) activates after the model has refused to answer a user input.}
\label{fig:refusal_detector_latents}
\end{figure}


\begin{table}[ht]
\centering
\scriptsize
\begin{tabular}{|c|c|c|c|c|c|c|c|p{3.5cm}|c|}
\hline
\textbf{Latent} & $\nu^\varepsilon$ & r($\nu^\varepsilon$) & $\nu^r$ & r($\nu^r$) & $\RND$ & r($\RND$) & $f_{template}$ & \textbf{ Description} & \textbf{Fig.} \\
\hline
70149 & -0.01 & 45 & 0.22 & 63 & 0.936 & 7 & 26.97\% &  Refusal related latent: Requests for harmful instructions. & \ref{fig:refusal_latents} \\
\hline
7736 & -0.02 & 54 & 0.15 & 33 & 0.917 & 50 & 47.99\% &  Refusal related latent: Generally sensitive content. & \ref{fig:refusal_latents} \\
\hline
24613 & -0.02 & 57 & 0.18 & 40 & 0.925 & 24 & 54.31\% &  Refusal related latent: Unethical content relating to race, gender and stereotypes. & \ref{fig:refusal_latents} \\
\hline
20384 & -0.10 & 128 & 0.25 & 82 & 0.918 & 42 & 32.34\% &  Refusal related latent: Requests for harmful instructions. & \ref{fig:refusal_latents} \\
\hline
38009 & 0.025 & 62 & 0.061 & 7 & 0.902 & 122 & 96.6\% &  Refusal related latent: The model has refused to answer a user input. & \ref{fig:refusal_detector_latents} \\
\hline
2138 & -0.02 & 56 & 0.43 & 131 & 0.918 & 47 & 27.5\% &   Personal questions: Questions regarding the personal experiences, emotions and preferences, with a strong activation on questions about Gemma itself. & \ref{fig:interpretable_latents} \\
\hline
14350 & -0.01 & 47 & 0.33 & 115 & 0.930 & 14 & 16.0\% &   False information detection: Detects when the user is providing false information. & \ref{fig:interpretable_latents} \\
\hline

\end{tabular}
\caption{Summary of a set of interpretable chat-specific latents identified in the \batchtopk crosscoder. The function $r$ represents the rank of the latent in the distribution of absolute values of $\nu^\varepsilon$ and $\nu^r$ of all \chatonly latents, where $r(\nu)$ means this latent has the lowest absolute value of $\nu$ of all \chatonly latents. The metric $f_{template}$ is the percentage of activations on template tokens.}
\label{tab:interpretable_latents_topk}
\end{table}

\begin{figure}[t!]
\centering
\begin{subfigure}[t!]{0.45\textwidth}  %
    \centering
\begin{tabular}{|p{0.95\columnwidth}|}  %
    \hline
\cellcolor{gray!20}\begin{minipage}[t]{0.95\columnwidth}   %
\begin{lstlisting}[style=colorchars,aboveskip=-5pt,belowskip=0pt]
|Max Activation: 57.045|\end{lstlisting}
\end{minipage} \\
\begin{minipage}[t]{0.95\columnwidth}  %
\begin{lstlisting}[style=colorchars,aboveskip=-5pt,belowskip=3pt,breaklines=true,breakatwhitespace=false]
|\allowbreak\hlbg{red!0}{<bos>}||\allowbreak\hlbg{red!0}{<sot>}||\allowbreak\hlbg{red!0}{user}||\allowbreak\hlbg{red!0}{\textbackslash n}|
|\allowbreak\hlbg{red!0}{Can}||\allowbreak\hlbg{red!0}{ you}||\allowbreak\hlbg{red!0}{ tell}||\allowbreak\hlbg{red!0}{ me}||\allowbreak\hlbg{red!0}{ a}||\allowbreak\hlbg{red!0}{ bit}||\allowbreak\hlbg{red!0}{ about}||\allowbreak\hlbg{red!0}{ New}||\allowbreak\hlbg{red!0}{ York}||\allowbreak\hlbg{red!0}{,}||\allowbreak\hlbg{red!0}{ the}||\allowbreak\hlbg{red!17}{ capital}||\allowbreak\hlbg{red!20}{ of}||\allowbreak\hlbg{red!11}{ swit}||\allowbreak\hlbg{red!13}{zerland}||\allowbreak\hlbg{red!36}{?}||\allowbreak\hlbg{red!41}{<eot>}||\allowbreak\hlbg{red!20}{\textbackslash n}|
|\allowbreak\hlbg{red!87}{<sot>}||\allowbreak\hlbg{red!71}{model}||\allowbreak\hlbg{red!88}{\textbackslash n}|
\end{lstlisting}
\end{minipage} \\ \hline
\end{tabular}
\begin{tabular}{|p{0.95\columnwidth}|}  %
\hline
\cellcolor{gray!20}\begin{minipage}[t]{0.95\columnwidth}   %
\begin{lstlisting}[style=colorchars,aboveskip=-5pt,belowskip=0pt]
|Max Activation: 0.000|\end{lstlisting}
\end{minipage} \\
\begin{minipage}[t]{0.95\columnwidth}  %
\begin{lstlisting}[style=colorchars,aboveskip=-5pt,belowskip=3pt,breaklines=true,breakatwhitespace=false]
|\allowbreak\hlbg{red!0}{<bos>}||\allowbreak\hlbg{red!0}{<sot>}||\allowbreak\hlbg{red!0}{user}||\allowbreak\hlbg{red!0}{\textbackslash n}|
|\allowbreak\hlbg{red!0}{Can}||\allowbreak\hlbg{red!0}{ you}||\allowbreak\hlbg{red!0}{ tell}||\allowbreak\hlbg{red!0}{ me}||\allowbreak\hlbg{red!0}{ a}||\allowbreak\hlbg{red!0}{ bit}||\allowbreak\hlbg{red!0}{ about}||\allowbreak\hlbg{red!0}{ Bern}||\allowbreak\hlbg{red!0}{,}||\allowbreak\hlbg{red!0}{ the}||\allowbreak\hlbg{red!0}{ capital}||\allowbreak\hlbg{red!0}{ of}||\allowbreak\hlbg{red!0}{ swit}||\allowbreak\hlbg{red!0}{zerland}||\allowbreak\hlbg{red!0}{?}||\allowbreak\hlbg{red!0}{<eot>}||\allowbreak\hlbg{red!0}{\textbackslash n}|
|\allowbreak\hlbg{red!0}{<sot>}||\allowbreak\hlbg{red!0}{model}||\allowbreak\hlbg{red!0}{\textbackslash n}|
\end{lstlisting}
\end{minipage} \\ \hline
\end{tabular}
\begin{tabular}{|p{0.95\columnwidth}|}  %
\hline
\cellcolor{gray!20}\begin{minipage}[t]{0.95\columnwidth}   %
\begin{lstlisting}[style=colorchars,aboveskip=-5pt,belowskip=0pt]
|Max Activation: 26.641|\end{lstlisting}
\end{minipage} \\
\begin{minipage}[t]{0.95\columnwidth}  %
\begin{lstlisting}[style=colorchars,aboveskip=-5pt,belowskip=3pt,breaklines=true,breakatwhitespace=false]
|\allowbreak\hlbg{red!0}{<bos>}||\allowbreak\hlbg{red!0}{<sot>}||\allowbreak\hlbg{red!0}{user}||\allowbreak\hlbg{red!0}{\textbackslash n}|
|\allowbreak\hlbg{red!0}{The}||\allowbreak\hlbg{red!0}{ Eiffel}||\allowbreak\hlbg{red!0}{ Tower}||\allowbreak\hlbg{red!0}{ is}||\allowbreak\hlbg{red!0}{ in}||\allowbreak\hlbg{red!0}{ Texas}||\allowbreak\hlbg{red!0}{.}||\allowbreak\hlbg{red!15}{<eot>}||\allowbreak\hlbg{red!26}{\textbackslash n}|
|\allowbreak\hlbg{red!36}{<sot>}||\allowbreak\hlbg{red!41}{model}||\allowbreak\hlbg{red!39}{\textbackslash n}|
\end{lstlisting}
\end{minipage} \\ \hline
\end{tabular}
\caption{\textbf{Latent 62019} activates on user inputs containing wrong information, similar to Latent 14350, but activates mostly on the template tokens.}
\label{fig:latent_62019}
\end{subfigure}
\hfill
\begin{subfigure}[t!]{0.45\textwidth} 
\vspace{-1.6em}
\begin{tabular}{|p{0.95\columnwidth}|}  %
    \hline
\cellcolor{gray!20}\begin{minipage}[t]{0.95\columnwidth}   %
\begin{lstlisting}[style=colorchars,aboveskip=-5pt,belowskip=0pt]
|Max Activation: 95.851|\end{lstlisting}
\end{minipage} \\
\begin{minipage}[t]{0.95\columnwidth}  %
\begin{lstlisting}[style=colorchars,aboveskip=-5pt,belowskip=3pt,breaklines=true,breakatwhitespace=false]
|\allowbreak\hlbg{red!0}{<bos>}||\allowbreak\hlbg{red!0}{<sot>}||\allowbreak\hlbg{red!0}{user}||\allowbreak\hlbg{red!0}{\textbackslash n}|
|\allowbreak\hlbg{red!0}{Can}||\allowbreak\hlbg{red!0}{ you}||\allowbreak\hlbg{red!0}{ please}||\allowbreak\hlbg{red!0}{ re}||\allowbreak\hlbg{red!0}{phrase}||\allowbreak\hlbg{red!0}{ the}||\allowbreak\hlbg{red!0}{ following}||\allowbreak\hlbg{red!0}{ sentence}||\allowbreak\hlbg{red!0}{:}||\allowbreak\hlbg{red!5}{<eot>}||\allowbreak\hlbg{red!0}{\textbackslash n}|
|\allowbreak\hlbg{red!22}{<sot>}||\allowbreak\hlbg{red!89}{model}||\allowbreak\hlbg{red!43}{\textbackslash n}|
\end{lstlisting}
\end{minipage} \\ \hline
\end{tabular}
\begin{tabular}{|p{0.95\columnwidth}|}  %
\hline
\cellcolor{gray!20}\begin{minipage}[t]{0.95\columnwidth}   %
\begin{lstlisting}[style=colorchars,aboveskip=-5pt,belowskip=0pt]
|Max Activation: 6.744|\end{lstlisting}
\end{minipage} \\
\begin{minipage}[t]{0.95\columnwidth}  %
\begin{lstlisting}[style=colorchars,aboveskip=-5pt,belowskip=3pt,breaklines=true,breakatwhitespace=false]
|\allowbreak\hlbg{red!0}{<bos>}||\allowbreak\hlbg{red!0}{<sot>}||\allowbreak\hlbg{red!0}{user}||\allowbreak\hlbg{red!0}{\textbackslash n}|
|\allowbreak\hlbg{red!0}{Can}||\allowbreak\hlbg{red!0}{ you}||\allowbreak\hlbg{red!0}{ please}||\allowbreak\hlbg{red!0}{ re}||\allowbreak\hlbg{red!0}{phrase}||\allowbreak\hlbg{red!0}{ the}||\allowbreak\hlbg{red!0}{ following}||\allowbreak\hlbg{red!0}{ sentence}||\allowbreak\hlbg{red!0}{:}||\allowbreak\hlbg{red!0}{ This}||\allowbreak\hlbg{red!0}{ is}||\allowbreak\hlbg{red!0}{ an}||\allowbreak\hlbg{red!0}{ ugly}||\allowbreak\hlbg{red!0}{ sentence}||\allowbreak\hlbg{red!0}{ is}||\allowbreak\hlbg{red!0}{.}||\allowbreak\hlbg{red!0}{<eot>}||\allowbreak\hlbg{red!0}{\textbackslash n}|
|\allowbreak\hlbg{red!0}{<sot>}||\allowbreak\hlbg{red!6}{model}||\allowbreak\hlbg{red!0}{\textbackslash n}|
\end{lstlisting}
\end{minipage} \\ \hline
\end{tabular}
\begin{tabular}{|p{0.95\columnwidth}|}  %
\hline
\cellcolor{gray!20}\begin{minipage}[t]{0.95\columnwidth}   %
\begin{lstlisting}[style=colorchars,aboveskip=-5pt,belowskip=0pt]
|Max Activation: 90.659|\end{lstlisting}
\end{minipage} \\
\begin{minipage}[t]{0.95\columnwidth}  %
\begin{lstlisting}[style=colorchars,aboveskip=-5pt,belowskip=3pt,breaklines=true,breakatwhitespace=false]
|\allowbreak\hlbg{red!0}{<bos>}||\allowbreak\hlbg{red!0}{<sot>}||\allowbreak\hlbg{red!0}{user}||\allowbreak\hlbg{red!0}{\textbackslash n}|
|\allowbreak\hlbg{red!0}{What}||\allowbreak\hlbg{red!0}{ do}||\allowbreak\hlbg{red!0}{ you}||\allowbreak\hlbg{red!0}{ think}||\allowbreak\hlbg{red!0}{ about}||\allowbreak\hlbg{red!0}{ that}||\allowbreak\hlbg{red!0}{?}||\allowbreak\hlbg{red!10}{<eot>}||\allowbreak\hlbg{red!0}{\textbackslash n}|
|\allowbreak\hlbg{red!36}{<sot>}||\allowbreak\hlbg{red!51}{model}||\allowbreak\hlbg{red!84}{\textbackslash n}|
\end{lstlisting}
\end{minipage} \\ \hline
\end{tabular}
\caption{\small \textbf{Latent 58070} triggers when the user request misses information. }
\label{fig:latent_58070}
\end{subfigure} \\
\begin{subfigure}[t!]{0.45\textwidth} 
\begin{tabular}{|p{0.95\columnwidth}|}  %
    \hline
\cellcolor{gray!20}\begin{minipage}[t]{0.95\columnwidth}   %
\begin{lstlisting}[style=colorchars,aboveskip=-5pt,belowskip=0pt]
|Max Activation: 0.000|\end{lstlisting}
\end{minipage} \\
\begin{minipage}[t]{0.95\columnwidth}  %
\begin{lstlisting}[style=colorchars,aboveskip=-5pt,belowskip=3pt,breaklines=true,breakatwhitespace=false]
|\allowbreak\hlbg{red!0}{<bos>}||\allowbreak\hlbg{red!0}{<sot>}||\allowbreak\hlbg{red!0}{user}||\allowbreak\hlbg{red!0}{\textbackslash n}|
|\allowbreak\hlbg{red!0}{"}||\allowbreak\hlbg{red!0}{Can}||\allowbreak\hlbg{red!0}{ you}||\allowbreak\hlbg{red!0}{ tell}||\allowbreak\hlbg{red!0}{ me}||\allowbreak\hlbg{red!0}{ a}||\allowbreak\hlbg{red!0}{ bit}||\allowbreak\hlbg{red!0}{ about}||\allowbreak\hlbg{red!0}{ Bern}||\allowbreak\hlbg{red!0}{,}||\allowbreak\hlbg{red!0}{ the}||\allowbreak\hlbg{red!0}{ capital}||\allowbreak\hlbg{red!0}{ of}||\allowbreak\hlbg{red!0}{ swit}||\allowbreak\hlbg{red!0}{zerland}||\allowbreak\hlbg{red!0}{?"}||\allowbreak\hlbg{red!0}{<eot>}||\allowbreak\hlbg{red!0}{\textbackslash n}|
|\allowbreak\hlbg{red!0}{<sot>}||\allowbreak\hlbg{red!0}{model}||\allowbreak\hlbg{red!0}{\textbackslash n}|
\end{lstlisting}
\end{minipage} \\ \hline
\end{tabular} 
\begin{tabular}{|p{0.95\columnwidth}|}  %
\hline
\cellcolor{gray!20}\begin{minipage}[t]{0.95\columnwidth}   %
\begin{lstlisting}[style=colorchars,aboveskip=-5pt,belowskip=0pt]
|Max Activation: 60.062|\end{lstlisting}
\end{minipage} \\
\begin{minipage}[t]{0.95\columnwidth}  %
\begin{lstlisting}[style=colorchars,aboveskip=-5pt,belowskip=3pt,breaklines=true,breakatwhitespace=false]
|\allowbreak\hlbg{red!0}{<bos>}||\allowbreak\hlbg{red!0}{<sot>}||\allowbreak\hlbg{red!0}{user}||\allowbreak\hlbg{red!0}{\textbackslash n}|
|\allowbreak\hlbg{red!0}{Par}||\allowbreak\hlbg{red!0}{aph}||\allowbreak\hlbg{red!0}{rase}||\allowbreak\hlbg{red!0}{ this}||\allowbreak\hlbg{red!0}{:}||\allowbreak\hlbg{red!0}{ "}||\allowbreak\hlbg{red!0}{Can}||\allowbreak\hlbg{red!0}{ you}||\allowbreak\hlbg{red!0}{ tell}||\allowbreak\hlbg{red!0}{ me}||\allowbreak\hlbg{red!0}{ a}||\allowbreak\hlbg{red!0}{ bit}||\allowbreak\hlbg{red!0}{ about}||\allowbreak\hlbg{red!0}{ Bern}||\allowbreak\hlbg{red!0}{,}||\allowbreak\hlbg{red!0}{ the}||\allowbreak\hlbg{red!0}{ capital}||\allowbreak\hlbg{red!0}{ of}||\allowbreak\hlbg{red!0}{ swit}||\allowbreak\hlbg{red!0}{zerland}||\allowbreak\hlbg{red!0}{?"}||\allowbreak\hlbg{red!0}{<eot>}||\allowbreak\hlbg{red!0}{\textbackslash n}|
|\allowbreak\hlbg{red!72}{<sot>}||\allowbreak\hlbg{red!9}{model}||\allowbreak\hlbg{red!0}{\textbackslash n}|
\end{lstlisting}
\end{minipage} \\ \hline
\end{tabular}
\begin{tabular}{|p{0.95\columnwidth}|}  %
\hline
\cellcolor{gray!20}\begin{minipage}[t]{0.95\columnwidth}   %
\begin{lstlisting}[style=colorchars,aboveskip=-5pt,belowskip=0pt]
|Max Activation: 68.774|\end{lstlisting}
\end{minipage} \\
\begin{minipage}[t]{0.95\columnwidth}  %
\begin{lstlisting}[style=colorchars,aboveskip=-5pt,belowskip=3pt,breaklines=true,breakatwhitespace=false]
|\allowbreak\hlbg{red!0}{<bos>}||\allowbreak\hlbg{red!0}{<sot>}||\allowbreak\hlbg{red!0}{user}||\allowbreak\hlbg{red!0}{\textbackslash n}|
|\allowbreak\hlbg{red!0}{Can}||\allowbreak\hlbg{red!0}{ you}||\allowbreak\hlbg{red!0}{ please}||\allowbreak\hlbg{red!0}{ rewrite}||\allowbreak\hlbg{red!0}{ the}||\allowbreak\hlbg{red!0}{ following}||\allowbreak\hlbg{red!0}{ sentence}||\allowbreak\hlbg{red!0}{?}||\allowbreak\hlbg{red!0}{ "}||\allowbreak\hlbg{red!0}{Can}||\allowbreak\hlbg{red!0}{ you}||\allowbreak\hlbg{red!0}{ tell}||\allowbreak\hlbg{red!0}{ me}||\allowbreak\hlbg{red!0}{ a}||\allowbreak\hlbg{red!0}{ bit}||\allowbreak\hlbg{red!0}{ about}||\allowbreak\hlbg{red!0}{ Bern}||\allowbreak\hlbg{red!0}{,}||\allowbreak\hlbg{red!0}{ the}||\allowbreak\hlbg{red!0}{ capital}||\allowbreak\hlbg{red!0}{ of}||\allowbreak\hlbg{red!0}{ swit}||\allowbreak\hlbg{red!0}{zerland}||\allowbreak\hlbg{red!0}{?"}||\allowbreak\hlbg{red!47}{<eot>}||\allowbreak\hlbg{red!0}{\textbackslash n}|
|\allowbreak\hlbg{red!83}{<sot>}||\allowbreak\hlbg{red!9}{model}||\allowbreak\hlbg{red!0}{\textbackslash n}|
\end{lstlisting}
\end{minipage} \\ \hline
\end{tabular}
\caption{\small \textbf{Latent 54087} activates when the model should rewrite or paraphrase something.}
\label{fig:latent_54087}
\end{subfigure}
\hfill
\begin{subfigure}[t!]{0.45\textwidth} 
\begin{tabular}{|p{0.95\columnwidth}|}  %
    \hline
\cellcolor{gray!20}\begin{minipage}[t]{0.95\columnwidth}   %
\begin{lstlisting}[style=colorchars,aboveskip=-5pt,belowskip=0pt]
|Max Activation: 60.401|\end{lstlisting}
\end{minipage} \\
\begin{minipage}[t]{0.95\columnwidth}  %
\begin{lstlisting}[style=colorchars,aboveskip=-5pt,belowskip=3pt,breaklines=true,breakatwhitespace=false]
|\allowbreak\hlbg{red!0}{<bos>}||\allowbreak\hlbg{red!0}{<sot>}||\allowbreak\hlbg{red!0}{user}||\allowbreak\hlbg{red!0}{\textbackslash n}|
|\allowbreak\hlbg{red!0}{I}||\allowbreak\hlbg{red!0}{ saw}||\allowbreak\hlbg{red!0}{ a}||\allowbreak\hlbg{red!0}{ sign}||\allowbreak\hlbg{red!0}{ that}||\allowbreak\hlbg{red!0}{ said}||\allowbreak\hlbg{red!0}{ "}||\allowbreak\hlbg{red!0}{watch}||\allowbreak\hlbg{red!0}{ for}||\allowbreak\hlbg{red!0}{ children}||\allowbreak\hlbg{red!0}{"}||\allowbreak\hlbg{red!0}{ and}||\allowbreak\hlbg{red!0}{ I}||\allowbreak\hlbg{red!0}{ thought}||\allowbreak\hlbg{red!0}{,}||\allowbreak\hlbg{red!0}{ "}||\allowbreak\hlbg{red!0}{That}||\allowbreak\hlbg{red!0}{ sounds}||\allowbreak\hlbg{red!0}{ like}||\allowbreak\hlbg{red!0}{ a}||\allowbreak\hlbg{red!0}{ fair}||\allowbreak\hlbg{red!0}{ trade}||\allowbreak\hlbg{red!0}{"}||\allowbreak\hlbg{red!70}{<eot>}||\allowbreak\hlbg{red!0}{\textbackslash n}|
|\allowbreak\hlbg{red!78}{<sot>}||\allowbreak\hlbg{red!35}{model}||\allowbreak\hlbg{red!42}{\textbackslash n}|
\end{lstlisting}
\end{minipage} \\ \hline
\end{tabular}
\begin{tabular}{|p{0.95\columnwidth}|}  %
\hline
\cellcolor{gray!20}\begin{minipage}[t]{0.95\columnwidth}   %
\begin{lstlisting}[style=colorchars,aboveskip=-5pt,belowskip=0pt]
|Max Activation: 7.731|\end{lstlisting}
\end{minipage} \\
\begin{minipage}[t]{0.95\columnwidth}  %
\begin{lstlisting}[style=colorchars,aboveskip=-5pt,belowskip=3pt,breaklines=true,breakatwhitespace=false]
|\allowbreak\hlbg{red!0}{<bos>}||\allowbreak\hlbg{red!0}{<sot>}||\allowbreak\hlbg{red!0}{user}||\allowbreak\hlbg{red!0}{\textbackslash n}|
|\allowbreak\hlbg{red!0}{I}||\allowbreak\hlbg{red!0}{ saw}||\allowbreak\hlbg{red!0}{ a}||\allowbreak\hlbg{red!0}{ sign}||\allowbreak\hlbg{red!0}{ that}||\allowbreak\hlbg{red!0}{ said}||\allowbreak\hlbg{red!0}{ "}||\allowbreak\hlbg{red!0}{watch}||\allowbreak\hlbg{red!0}{ for}||\allowbreak\hlbg{red!0}{ children}||\allowbreak\hlbg{red!0}{"}||\allowbreak\hlbg{red!0}{ and}||\allowbreak\hlbg{red!0}{ I}||\allowbreak\hlbg{red!0}{ slowed}||\allowbreak\hlbg{red!0}{ down}||\allowbreak\hlbg{red!0}{ my}||\allowbreak\hlbg{red!0}{ car}||\allowbreak\hlbg{red!0}{.}||\allowbreak\hlbg{red!0}{<eot>}||\allowbreak\hlbg{red!0}{\textbackslash n}|
|\allowbreak\hlbg{red!10}{<sot>}||\allowbreak\hlbg{red!0}{model}||\allowbreak\hlbg{red!0}{\textbackslash n}|
\end{lstlisting}
\end{minipage} \\ \hline
\end{tabular}
\begin{tabular}{|p{0.95\columnwidth}|}  %
\hline
\cellcolor{gray!20}\begin{minipage}[t]{0.95\columnwidth}   %
\begin{lstlisting}[style=colorchars,aboveskip=-5pt,belowskip=0pt]
|Max Activation: 50.651|\end{lstlisting}
\end{minipage} \\
\begin{minipage}[t]{0.95\columnwidth}  %
\begin{lstlisting}[style=colorchars,aboveskip=-5pt,belowskip=3pt,breaklines=true,breakatwhitespace=false]
|\allowbreak\hlbg{red!0}{<bos>}||\allowbreak\hlbg{red!0}{<sot>}||\allowbreak\hlbg{red!0}{user}||\allowbreak\hlbg{red!0}{\textbackslash n}|
|\allowbreak\hlbg{red!0}{It}||\allowbreak\hlbg{red!0}{'}||\allowbreak\hlbg{red!0}{s}||\allowbreak\hlbg{red!0}{ hard}||\allowbreak\hlbg{red!0}{ to}||\allowbreak\hlbg{red!0}{ explain}||\allowbreak\hlbg{red!10}{ puns}||\allowbreak\hlbg{red!0}{ to}||\allowbreak\hlbg{red!0}{ kle}||\allowbreak\hlbg{red!0}{pto}||\allowbreak\hlbg{red!0}{mani}||\allowbreak\hlbg{red!0}{acs}||\allowbreak\hlbg{red!0}{ because}||\allowbreak\hlbg{red!0}{ they}||\allowbreak\hlbg{red!0}{ always}||\allowbreak\hlbg{red!0}{ take}||\allowbreak\hlbg{red!0}{ things}||\allowbreak\hlbg{red!0}{ literally}||\allowbreak\hlbg{red!0}{.}||\allowbreak\hlbg{red!34}{<eot>}||\allowbreak\hlbg{red!17}{\textbackslash n}|
|\allowbreak\hlbg{red!66}{<sot>}||\allowbreak\hlbg{red!52}{model}||\allowbreak\hlbg{red!36}{\textbackslash n}|
\end{lstlisting}
\end{minipage} \\ \hline
\end{tabular}
\caption{\small \textbf{Latent 50586} activates after jokes.}
\label{fig:latent_50586}
\end{subfigure}
\caption{Examples of interpretable \chatonly latents from the \batchtopk crosscoder. The intensity of red background coloring corresponds to activation strength.}
\label{fig:more_interpretable_latents_1}
\end{figure}

\begin{figure}[t!]
\centering
\begin{subfigure}[t!]{0.45\textwidth}  %
    \centering
\begin{tabular}{|p{0.95\columnwidth}|}  %
    \hline
\cellcolor{gray!20}\begin{minipage}[t]{0.95\columnwidth}   %
\begin{lstlisting}[style=colorchars,aboveskip=-5pt,belowskip=0pt]
|Max Activation: 16.746|\end{lstlisting}
\end{minipage} \\
\begin{minipage}[t]{0.95\columnwidth}  %
\begin{lstlisting}[style=colorchars,aboveskip=-5pt,belowskip=3pt,breaklines=true,breakatwhitespace=false]
|\allowbreak\hlbg{red!0}{<bos>}||\allowbreak\hlbg{red!0}{<sot>}||\allowbreak\hlbg{red!0}{user}||\allowbreak\hlbg{red!0}{\textbackslash n}|
|\allowbreak\hlbg{red!0}{write}||\allowbreak\hlbg{red!0}{ me}||\allowbreak\hlbg{red!0}{ a}||\allowbreak\hlbg{red!0}{ }||\allowbreak\hlbg{red!0}{1}||\allowbreak\hlbg{red!0}{ word}||\allowbreak\hlbg{red!0}{ essay}||\allowbreak\hlbg{red!0}{ about}||\allowbreak\hlbg{red!0}{ "}||\allowbreak\hlbg{red!0}{behavioral}||\allowbreak\hlbg{red!0}{ cloning}||\allowbreak\hlbg{red!0}{ for}||\allowbreak\hlbg{red!0}{ imitation}||\allowbreak\hlbg{red!0}{ learning}||\allowbreak\hlbg{red!0}{ for}||\allowbreak\hlbg{red!0}{ robots}||\allowbreak\hlbg{red!0}{".}||\allowbreak\hlbg{red!0}{<eot>}||\allowbreak\hlbg{red!0}{\textbackslash n}|
|\allowbreak\hlbg{red!0}{<sot>}||\allowbreak\hlbg{red!22}{model}||\allowbreak\hlbg{red!11}{\textbackslash n}|
\end{lstlisting}
\end{minipage} \\ \hline
\end{tabular}
\begin{tabular}{|p{0.95\columnwidth}|}  %
    \hline
\cellcolor{gray!20}\begin{minipage}[t]{0.95\columnwidth}   %
\begin{lstlisting}[style=colorchars,aboveskip=-5pt,belowskip=0pt]
|Max Activation: 47.931|\end{lstlisting}
\end{minipage} \\
\begin{minipage}[t]{0.95\columnwidth}  %
\begin{lstlisting}[style=colorchars,aboveskip=-5pt,belowskip=3pt,breaklines=true,breakatwhitespace=false]
|\allowbreak\hlbg{red!0}{<bos>}||\allowbreak\hlbg{red!0}{<sot>}||\allowbreak\hlbg{red!0}{user}||\allowbreak\hlbg{red!0}{\textbackslash n}|
|\allowbreak\hlbg{red!0}{write}||\allowbreak\hlbg{red!0}{ me}||\allowbreak\hlbg{red!0}{ a}||\allowbreak\hlbg{red!0}{ }||\allowbreak\hlbg{red!0}{1}||\allowbreak\hlbg{red!0}{ sentence}||\allowbreak\hlbg{red!13}{ essay}||\allowbreak\hlbg{red!20}{ about}||\allowbreak\hlbg{red!0}{ "}||\allowbreak\hlbg{red!0}{behavioral}||\allowbreak\hlbg{red!0}{ cloning}||\allowbreak\hlbg{red!0}{ for}||\allowbreak\hlbg{red!0}{ imitation}||\allowbreak\hlbg{red!0}{ learning}||\allowbreak\hlbg{red!0}{ for}||\allowbreak\hlbg{red!0}{ robots}||\allowbreak\hlbg{red!0}{".}||\allowbreak\hlbg{red!14}{<eot>}||\allowbreak\hlbg{red!0}{\textbackslash n}|
|\allowbreak\hlbg{red!20}{<sot>}||\allowbreak\hlbg{red!64}{model}||\allowbreak\hlbg{red!34}{\textbackslash n}|
\end{lstlisting}
\end{minipage} \\ \hline
\end{tabular}
\begin{tabular}{|p{0.95\columnwidth}|}  %
    \hline
\cellcolor{gray!20}\begin{minipage}[t]{0.95\columnwidth}   %
\begin{lstlisting}[style=colorchars,aboveskip=-5pt,belowskip=0pt]
|Max Activation: 60.197|\end{lstlisting}
\end{minipage} \\
\begin{minipage}[t]{0.95\columnwidth}  %
\begin{lstlisting}[style=colorchars,aboveskip=-5pt,belowskip=3pt,breaklines=true,breakatwhitespace=false]
|\allowbreak\hlbg{red!0}{<bos>}||\allowbreak\hlbg{red!0}{<sot>}||\allowbreak\hlbg{red!0}{user}||\allowbreak\hlbg{red!0}{\textbackslash n}|
|\allowbreak\hlbg{red!0}{write}||\allowbreak\hlbg{red!0}{ me}||\allowbreak\hlbg{red!0}{ a}||\allowbreak\hlbg{red!0}{ }||\allowbreak\hlbg{red!0}{4}||\allowbreak\hlbg{red!0}{ sentence}||\allowbreak\hlbg{red!23}{ essay}||\allowbreak\hlbg{red!23}{ about}||\allowbreak\hlbg{red!0}{ "}||\allowbreak\hlbg{red!0}{behavioral}||\allowbreak\hlbg{red!0}{ cloning}||\allowbreak\hlbg{red!0}{ for}||\allowbreak\hlbg{red!0}{ imitation}||\allowbreak\hlbg{red!0}{ learning}||\allowbreak\hlbg{red!0}{ for}||\allowbreak\hlbg{red!0}{ robots}||\allowbreak\hlbg{red!0}{".}||\allowbreak\hlbg{red!10}{<eot>}||\allowbreak\hlbg{red!0}{\textbackslash n}|
|\allowbreak\hlbg{red!20}{<sot>}||\allowbreak\hlbg{red!80}{model}||\allowbreak\hlbg{red!38}{\textbackslash n}|
\end{lstlisting}
\end{minipage} \\ \hline
\end{tabular}
\begin{tabular}{|p{0.95\columnwidth}|}  %
    \hline
\cellcolor{gray!20}\begin{minipage}[t]{0.95\columnwidth}   %
\begin{lstlisting}[style=colorchars,aboveskip=-5pt,belowskip=0pt]
|Max Activation: 73.759|\end{lstlisting}
\end{minipage} \\
\begin{minipage}[t]{0.95\columnwidth}  %
\begin{lstlisting}[style=colorchars,aboveskip=-5pt,belowskip=3pt,breaklines=true,breakatwhitespace=false]
|\allowbreak\hlbg{red!0}{<bos>}||\allowbreak\hlbg{red!0}{<sot>}||\allowbreak\hlbg{red!0}{user}||\allowbreak\hlbg{red!0}{\textbackslash n}|
|\allowbreak\hlbg{red!0}{write}||\allowbreak\hlbg{red!0}{ me}||\allowbreak\hlbg{red!0}{ a}||\allowbreak\hlbg{red!10}{ paragraph}||\allowbreak\hlbg{red!12}{ about}||\allowbreak\hlbg{red!8}{ "}||\allowbreak\hlbg{red!0}{behavioral}||\allowbreak\hlbg{red!0}{ cloning}||\allowbreak\hlbg{red!0}{ for}||\allowbreak\hlbg{red!0}{ imitation}||\allowbreak\hlbg{red!0}{ learning}||\allowbreak\hlbg{red!0}{ for}||\allowbreak\hlbg{red!0}{ robots}||\allowbreak\hlbg{red!0}{".}||\allowbreak\hlbg{red!0}{<eot>}||\allowbreak\hlbg{red!0}{\textbackslash n}|
|\allowbreak\hlbg{red!24}{<sot>}||\allowbreak\hlbg{red!98}{model}||\allowbreak\hlbg{red!34}{\textbackslash n}|
\end{lstlisting}
\end{minipage} \\ \hline
\end{tabular}
\begin{tabular}{|p{0.95\columnwidth}|}  %
    \hline
\cellcolor{gray!20}\begin{minipage}[t]{0.95\columnwidth}   %
\begin{lstlisting}[style=colorchars,aboveskip=-5pt,belowskip=0pt]
|Max Activation: 41.479|\end{lstlisting}
\end{minipage} \\
\begin{minipage}[t]{0.95\columnwidth}  %
\begin{lstlisting}[style=colorchars,aboveskip=-5pt,belowskip=3pt,breaklines=true,breakatwhitespace=false]
|\allowbreak\hlbg{red!0}{<bos>}||\allowbreak\hlbg{red!0}{<sot>}||\allowbreak\hlbg{red!0}{user}||\allowbreak\hlbg{red!0}{\textbackslash n}|
|\allowbreak\hlbg{red!0}{write}||\allowbreak\hlbg{red!0}{ me}||\allowbreak\hlbg{red!0}{ a}||\allowbreak\hlbg{red!0}{ }||\allowbreak\hlbg{red!0}{1}||\allowbreak\hlbg{red!0}{ page}||\allowbreak\hlbg{red!12}{ essay}||\allowbreak\hlbg{red!8}{ about}||\allowbreak\hlbg{red!0}{ "}||\allowbreak\hlbg{red!0}{behavioral}||\allowbreak\hlbg{red!0}{ cloning}||\allowbreak\hlbg{red!0}{ for}||\allowbreak\hlbg{red!0}{ imitation}||\allowbreak\hlbg{red!0}{ learning}||\allowbreak\hlbg{red!0}{ for}||\allowbreak\hlbg{red!0}{ robots}||\allowbreak\hlbg{red!0}{".}||\allowbreak\hlbg{red!0}{<eot>}||\allowbreak\hlbg{red!0}{\textbackslash n}|
|\allowbreak\hlbg{red!10}{<sot>}||\allowbreak\hlbg{red!47}{model}||\allowbreak\hlbg{red!55}{\textbackslash n}|
\end{lstlisting}
\end{minipage} \\ \hline
\end{tabular}
\begin{tabular}{|p{0.95\columnwidth}|}  %
    \hline
\cellcolor{gray!20}\begin{minipage}[t]{0.95\columnwidth}   %
\begin{lstlisting}[style=colorchars,aboveskip=-5pt,belowskip=0pt]
|Max Activation: 24.315|\end{lstlisting}
\end{minipage} \\
\begin{minipage}[t]{0.95\columnwidth}  %
\begin{lstlisting}[style=colorchars,aboveskip=-5pt,belowskip=3pt,breaklines=true,breakatwhitespace=false]
|\allowbreak\hlbg{red!0}{<bos>}||\allowbreak\hlbg{red!0}{<sot>}||\allowbreak\hlbg{red!0}{user}||\allowbreak\hlbg{red!0}{\textbackslash n}|
|\allowbreak\hlbg{red!0}{write}||\allowbreak\hlbg{red!0}{ me}||\allowbreak\hlbg{red!0}{ a}||\allowbreak\hlbg{red!0}{ }||\allowbreak\hlbg{red!0}{1}||\allowbreak\hlbg{red!0}{0}||\allowbreak\hlbg{red!0}{ page}||\allowbreak\hlbg{red!0}{ essay}||\allowbreak\hlbg{red!0}{ about}||\allowbreak\hlbg{red!0}{ "}||\allowbreak\hlbg{red!0}{behavioral}||\allowbreak\hlbg{red!0}{ cloning}||\allowbreak\hlbg{red!0}{ for}||\allowbreak\hlbg{red!0}{ imitation}||\allowbreak\hlbg{red!0}{ learning}||\allowbreak\hlbg{red!0}{ for}||\allowbreak\hlbg{red!0}{ robots}||\allowbreak\hlbg{red!0}{".}||\allowbreak\hlbg{red!0}{<eot>}||\allowbreak\hlbg{red!0}{\textbackslash n}|
|\allowbreak\hlbg{red!0}{<sot>}||\allowbreak\hlbg{red!0}{model}||\allowbreak\hlbg{red!32}{\textbackslash n}|
\end{lstlisting}
\end{minipage} \\ \hline
\end{tabular}
\caption{\textbf{Latent 69447} measures requested response length, with highest activation on a request for a paragraph.}
\label{fig:latent_69447}
\end{subfigure}
\hfill
\begin{subfigure}[t!]{0.45\textwidth} 
\vspace{-1.6em}
\begin{tabular}{|p{0.95\columnwidth}|}  %
    \hline
\cellcolor{gray!20}\begin{minipage}[t]{0.95\columnwidth}   %
\begin{lstlisting}[style=colorchars,aboveskip=-5pt,belowskip=0pt]
|Max Activation: 100.611|\end{lstlisting}
\end{minipage} \\
\hline
\end{tabular}
\begin{tabular}{|p{0.95\columnwidth}|}  %
    \hline
\cellcolor{gray!20}\begin{minipage}[t]{0.95\columnwidth}   %
\begin{lstlisting}[style=colorchars,aboveskip=-5pt,belowskip=0pt]
|Max Activation: 16.710|\end{lstlisting}
\end{minipage} \\
\hline
\end{tabular}
\caption{\small \textbf{Latent 10925} triggers strongly when the user requests a summarization. }
\label{fig:latent_10925}
\end{subfigure} \\

\caption{Examples of interpretable \chatonly latents from the \batchtopk crosscoder. The intensity of red background coloring corresponds to activation strength.}
\label{fig:more_interpretable_latents_2}
\end{figure}

\begin{figure}[t!]
\centering
\begin{subfigure}[t!]{0.45\textwidth}  %
    \centering
\begin{tabular}{|p{0.95\columnwidth}|}  %
    \hline
\cellcolor{gray!20}\begin{minipage}[t]{0.95\columnwidth}   %
\begin{lstlisting}[style=colorchars,aboveskip=-5pt,belowskip=0pt]
|Max Activation: 0.000|\end{lstlisting}
\end{minipage} \\
\begin{minipage}[t]{0.95\columnwidth}  %
\begin{lstlisting}[style=colorchars,aboveskip=-5pt,belowskip=3pt,breaklines=true,breakatwhitespace=false]
|\allowbreak\hlbg{red!0}{<bos>}||\allowbreak\hlbg{red!0}{<sot>}||\allowbreak\hlbg{red!0}{user}||\allowbreak\hlbg{red!0}{\textbackslash n}|
|\allowbreak\hlbg{red!0}{Who}||\allowbreak\hlbg{red!0}{ are}||\allowbreak\hlbg{red!0}{ the}||\allowbreak\hlbg{red!0}{ Giants}||\allowbreak\hlbg{red!0}{?}||\allowbreak\hlbg{red!0}{<end\_of\_turn>}||\allowbreak\hlbg{red!0}{\textbackslash n}|
|\allowbreak\hlbg{red!0}{<sot>}||\allowbreak\hlbg{red!0}{model}||\allowbreak\hlbg{red!0}{\textbackslash n}|
\end{lstlisting}
\end{minipage} \\ \hline
\end{tabular}
\begin{tabular}{|p{0.95\columnwidth}|}  %
    \hline
\cellcolor{gray!20}\begin{minipage}[t]{0.95\columnwidth}   %
\begin{lstlisting}[style=colorchars,aboveskip=-5pt,belowskip=0pt]
|Max Activation: 46.412|\end{lstlisting}
\end{minipage} \\
\begin{minipage}[t]{0.95\columnwidth}  %
\begin{lstlisting}[style=colorchars,aboveskip=-5pt,belowskip=3pt,breaklines=true,breakatwhitespace=false]
|\allowbreak\hlbg{red!0}{<bos>}||\allowbreak\hlbg{red!0}{<sot>}||\allowbreak\hlbg{red!0}{user}||\allowbreak\hlbg{red!0}{\textbackslash n}|
|\allowbreak\hlbg{red!0}{How}||\allowbreak\hlbg{red!0}{ did}||\allowbreak\hlbg{red!0}{ the}||\allowbreak\hlbg{red!0}{ Giants}||\allowbreak\hlbg{red!0}{ play}||\allowbreak\hlbg{red!0}{ in}||\allowbreak\hlbg{red!0}{ the}||\allowbreak\hlbg{red!0}{ MLB}||\allowbreak\hlbg{red!0}{ yesterday}||\allowbreak\hlbg{red!0}{?}||\allowbreak\hlbg{red!8}{<end\_of\_turn>}||\allowbreak\hlbg{red!0}{\textbackslash n}|
|\allowbreak\hlbg{red!33}{<sot>}||\allowbreak\hlbg{red!59}{model}||\allowbreak\hlbg{red!24}{\textbackslash n}|
\end{lstlisting}
\end{minipage} \\ \hline
\end{tabular}
\begin{tabular}{|p{0.95\columnwidth}|}  %
    \hline
\cellcolor{gray!20}\begin{minipage}[t]{0.95\columnwidth}   %
\begin{lstlisting}[style=colorchars,aboveskip=-5pt,belowskip=0pt]
|Max Activation: 52.380|\end{lstlisting}
\end{minipage} \\
\begin{minipage}[t]{0.95\columnwidth}  %
\begin{lstlisting}[style=colorchars,aboveskip=-5pt,belowskip=3pt,breaklines=true,breakatwhitespace=false]
|\allowbreak\hlbg{red!0}{<bos>}||\allowbreak\hlbg{red!0}{<sot>}||\allowbreak\hlbg{red!0}{user}||\allowbreak\hlbg{red!0}{\textbackslash n}|
|\allowbreak\hlbg{red!0}{What}||\allowbreak\hlbg{red!0}{ is}||\allowbreak\hlbg{red!0}{ the}||\allowbreak\hlbg{red!0}{ current}||\allowbreak\hlbg{red!0}{ Gold}||\allowbreak\hlbg{red!0}{ price}||\allowbreak\hlbg{red!22}{?}||\allowbreak\hlbg{red!36}{<end\_of\_turn>}||\allowbreak\hlbg{red!0}{\textbackslash n}|
|\allowbreak\hlbg{red!33}{<sot>}||\allowbreak\hlbg{red!67}{model}||\allowbreak\hlbg{red!61}{\textbackslash n}|
\end{lstlisting}
\end{minipage} \\ \hline
\end{tabular}
\begin{tabular}{|p{0.95\columnwidth}|}  %
\hline
\cellcolor{gray!20}\begin{minipage}[t]{0.95\columnwidth}   %
\begin{lstlisting}[style=colorchars,aboveskip=-5pt,belowskip=0pt]
|Max Activation: 0.000|\end{lstlisting}
\end{minipage} \\
\begin{minipage}[t]{0.95\columnwidth}  %
\begin{lstlisting}[style=colorchars,aboveskip=-5pt,belowskip=3pt,breaklines=true,breakatwhitespace=false]
|\allowbreak\hlbg{red!0}{<bos>}||\allowbreak\hlbg{red!0}{<sot>}||\allowbreak\hlbg{red!0}{user}||\allowbreak\hlbg{red!0}{\textbackslash n}|
|\allowbreak\hlbg{red!0}{What}||\allowbreak\hlbg{red!0}{ determines}||\allowbreak\hlbg{red!0}{ the}||\allowbreak\hlbg{red!0}{ current}||\allowbreak\hlbg{red!0}{ Gold}||\allowbreak\hlbg{red!0}{ price}||\allowbreak\hlbg{red!0}{?}||\allowbreak\hlbg{red!0}{<end\_of\_turn>}||\allowbreak\hlbg{red!0}{\textbackslash n}|
|\allowbreak\hlbg{red!0}{<sot>}||\allowbreak\hlbg{red!0}{model}||\allowbreak\hlbg{red!0}{\textbackslash n}|
\end{lstlisting}
\end{minipage} \\ \hline
\end{tabular}
\caption{\textbf{Latent 6583} activates on knowledge boundaries, where the model is missing access to information.}
\label{fig:latent_6583}
\end{subfigure}
\hfill
\begin{subfigure}[t!]{0.45\textwidth}  %
    \centering
\begin{tabular}{|p{0.95\columnwidth}|}  %
    \hline
\cellcolor{gray!20}\begin{minipage}[t]{0.95\columnwidth}   %
\begin{lstlisting}[style=colorchars,aboveskip=-5pt,belowskip=0pt]
|Max Activation: 82.172|\end{lstlisting}
\end{minipage} \\
\begin{minipage}[t]{0.95\columnwidth}  %
\begin{lstlisting}[style=colorchars,aboveskip=-5pt,belowskip=3pt,breaklines=true,breakatwhitespace=false]
|\allowbreak\hlbg{red!0}{<bos>}||\allowbreak\hlbg{red!0}{<start\_of\_turn>}||\allowbreak\hlbg{red!0}{user}||\allowbreak\hlbg{red!0}{\textbackslash n}|
|\allowbreak\hlbg{red!0}{Give}||\allowbreak\hlbg{red!0}{ me}||\allowbreak\hlbg{red!0}{ a}||\allowbreak\hlbg{red!0}{ detailed}||\allowbreak\hlbg{red!0}{ recipe}||\allowbreak\hlbg{red!0}{ of}||\allowbreak\hlbg{red!0}{ an}||\allowbreak\hlbg{red!0}{ apple}||\allowbreak\hlbg{red!0}{ cake}||\allowbreak\hlbg{red!0}{.}||\allowbreak\hlbg{red!0}{<end\_of\_turn>}||\allowbreak\hlbg{red!0}{\textbackslash n}|
|\allowbreak\hlbg{red!9}{<start\_of\_turn>}||\allowbreak\hlbg{red!96}{model}||\allowbreak\hlbg{red!9}{\textbackslash n}|
\end{lstlisting}
\end{minipage} \\ \hline
\end{tabular}
\begin{tabular}{|p{0.95\columnwidth}|}  %
    \hline
\cellcolor{gray!20}\begin{minipage}[t]{0.95\columnwidth}   %
\begin{lstlisting}[style=colorchars,aboveskip=-5pt,belowskip=0pt]
|Max Activation: 80.559|\end{lstlisting}
\end{minipage} \\
\begin{minipage}[t]{0.95\columnwidth}  %
\begin{lstlisting}[style=colorchars,aboveskip=-5pt,belowskip=3pt,breaklines=true,breakatwhitespace=false]
|\allowbreak\hlbg{red!0}{<bos>}||\allowbreak\hlbg{red!0}{<start\_of\_turn>}||\allowbreak\hlbg{red!0}{user}||\allowbreak\hlbg{red!0}{\textbackslash n}|
|\allowbreak\hlbg{red!0}{Give}||\allowbreak\hlbg{red!0}{ me}||\allowbreak\hlbg{red!0}{ a}||\allowbreak\hlbg{red!0}{ lengthy}||\allowbreak\hlbg{red!0}{ recipe}||\allowbreak\hlbg{red!0}{ of}||\allowbreak\hlbg{red!0}{ an}||\allowbreak\hlbg{red!0}{ apple}||\allowbreak\hlbg{red!0}{ cake}||\allowbreak\hlbg{red!0}{.}||\allowbreak\hlbg{red!0}{<end\_of\_turn>}||\allowbreak\hlbg{red!0}{\textbackslash n}|
|\allowbreak\hlbg{red!13}{<start\_of\_turn>}||\allowbreak\hlbg{red!95}{model}||\allowbreak\hlbg{red!11}{\textbackslash n}|
\end{lstlisting}
\end{minipage} \\ \hline
\end{tabular}
\begin{tabular}{|p{0.95\columnwidth}|}  %
    \hline
\cellcolor{gray!20}\begin{minipage}[t]{0.95\columnwidth}   %
\begin{lstlisting}[style=colorchars,aboveskip=-5pt,belowskip=0pt]
|Max Activation: 19.872|\end{lstlisting}
\end{minipage} \\
\begin{minipage}[t]{0.95\columnwidth}  %
\begin{lstlisting}[style=colorchars,aboveskip=-5pt,belowskip=3pt,breaklines=true,breakatwhitespace=false]
|\allowbreak\hlbg{red!0}{<bos>}||\allowbreak\hlbg{red!0}{<start\_of\_turn>}||\allowbreak\hlbg{red!0}{user}||\allowbreak\hlbg{red!0}{\textbackslash n}|
|\allowbreak\hlbg{red!0}{Give}||\allowbreak\hlbg{red!0}{ me}||\allowbreak\hlbg{red!0}{ a}||\allowbreak\hlbg{red!0}{ super}||\allowbreak\hlbg{red!0}{ short}||\allowbreak\hlbg{red!0}{ recipe}||\allowbreak\hlbg{red!0}{ of}||\allowbreak\hlbg{red!0}{ an}||\allowbreak\hlbg{red!0}{ apple}||\allowbreak\hlbg{red!0}{ cake}||\allowbreak\hlbg{red!0}{.}||\allowbreak\hlbg{red!0}{<end\_of\_turn>}||\allowbreak\hlbg{red!0}{\textbackslash n}|
|\allowbreak\hlbg{red!0}{<start\_of\_turn>}||\allowbreak\hlbg{red!23}{model}||\allowbreak\hlbg{red!0}{\textbackslash n}|
\end{lstlisting}
\end{minipage} \\ \hline
\end{tabular}
\begin{tabular}{|p{0.95\columnwidth}|}  %
    \hline
\cellcolor{gray!20}\begin{minipage}[t]{0.95\columnwidth}   %
\begin{lstlisting}[style=colorchars,aboveskip=-5pt,belowskip=0pt]
|Max Activation: 0.000|\end{lstlisting}
\end{minipage} \\
\begin{minipage}[t]{0.95\columnwidth}  %
\begin{lstlisting}[style=colorchars,aboveskip=-5pt,belowskip=3pt,breaklines=true,breakatwhitespace=false]
|\allowbreak\hlbg{red!0}{<bos>}||\allowbreak\hlbg{red!0}{<start\_of\_turn>}||\allowbreak\hlbg{red!0}{user}||\allowbreak\hlbg{red!0}{\textbackslash n}|
|\allowbreak\hlbg{red!0}{Give}||\allowbreak\hlbg{red!0}{ me}||\allowbreak\hlbg{red!0}{ a}||\allowbreak\hlbg{red!0}{ one}||\allowbreak\hlbg{red!0}{ sentence}||\allowbreak\hlbg{red!0}{ recipe}||\allowbreak\hlbg{red!0}{ of}||\allowbreak\hlbg{red!0}{ an}||\allowbreak\hlbg{red!0}{ apple}||\allowbreak\hlbg{red!0}{ cake}||\allowbreak\hlbg{red!0}{.}||\allowbreak\hlbg{red!0}{<end\_of\_turn>}||\allowbreak\hlbg{red!0}{\textbackslash n}|
|\allowbreak\hlbg{red!0}{<start\_of\_turn>}||\allowbreak\hlbg{red!0}{model}||\allowbreak\hlbg{red!0}{\textbackslash n}|
\end{lstlisting}
\end{minipage} \\ \hline
\end{tabular}
\caption{\textbf{Latent 4622} activates on requests for detailed information.}
\label{fig:latent_4622}
\end{subfigure}
\caption{Examples of interpretable \chatonly latents from the \batchtopk crosscoder. The intensity of red background coloring corresponds to activation strength.}
\label{fig:more_interpretable_latents_3}
\end{figure}

\begin{table}[ht]
\centering
\scriptsize
\begin{tabular}{|c|c|c|c|c|c|c|c|p{3.5cm}|c|}
\hline
\textbf{Latent} & $\nu^\varepsilon$ & r($\nu^\varepsilon$) & $\nu^r$ & r($\nu^r$) & $\RND$ & r($\RND$) & $f_{template}$ & \textbf{ Description} & \textbf{Fig.} \\
\hline
72073 & 0.050 & 54 & 0.300 & 159 & 0.903 & 3143 & 91.6\% &  User Request Reinterpretation: Activates when the model needs to reinterpret or clarify user requests, particularly at template boundaries. & \ref{fig:latent_72073} \\
\hline
57717 & 0.043 & 36 & 0.243 & 91 & 0.945 & 2598 & 93.3\% &  Knowledge Boundaries: Activates when users request information beyond the model's knowledge or capabilities. & \ref{fig:latent_57717} \\
\hline
68066 & 0.055 & 62 & 0.276 & 135 & 0.940 & 2686 & 72.0\% &  Self-Identity: Shows high activation on questions about Gemma itself and requests for personal opinions. & \ref{fig:latent_68066} \\
\hline
51823 & 0.076 & 84 & 0.264 & 123 & 0.947 & 2558 & 85.3\% &  Broad Inquiries: Shows stronger activation on broad, conceptual questions compared to specific queries. & \ref{fig:latent_51823}\\
\hline
51408 & 0.197 & 404 & 0.590 & 901 & 0.964 & 1963 & 20.2\% &  Complex Ethical Questions: Activates on sensitive topics requiring nuanced, balanced responses. This latent doesn't have particularly low $\nu^\varepsilon$ or $\nu^r$ values, but it is quite interesting and was found earlier in the analysis.& \ref{fig:latent_51408}, \ref{fig:latent_51408_more}\\
\hline
\end{tabular}
\caption{Summary of a set of interpretable chat-specific latents identified in the \Lone crosscoder. The function $r$ represents the rank of the latent in the distribution of absolute values of $\nu^\varepsilon$ and $\nu^r$ of all \chatonly latents, where $r(\nu)$ means this latent has the lowest absolute value of $\nu$ of all \chatonly latents. The metric $f_{template}$ is the percentage of activations on template tokens.}
\label{tab:interpretable_latents_cc}
\end{table}

\begin{figure}[ht]
    \begin{minipage}[b]{0.48\textwidth}
        \centering
  \begin{tabular}{|p{0.95\columnwidth}|}  %
    \hline
\cellcolor{gray!20}\begin{minipage}[t]{0.95\columnwidth}   %
\begin{lstlisting}[style=colorchars,aboveskip=-5pt,belowskip=0pt]
|\textbf{Feature 72073}|
|Max Activation: 79.156|\end{lstlisting}
\end{minipage} \\
\hline
\end{tabular}        
\subcaption{High activation on request reinterpretation}
    \end{minipage}
    \hfill
    \begin{subfigure}[b]{0.48\textwidth}
        \centering
  \begin{tabular}{|p{0.95\columnwidth}|}  %
    \hline
\cellcolor{gray!20}\begin{minipage}[t]{0.95\columnwidth}   %
\begin{lstlisting}[style=colorchars,aboveskip=-5pt,belowskip=0pt]
|\textbf{Feature 72073}|
|Max Activation: 55.107|\end{lstlisting}
\end{minipage} \\
\begin{minipage}[t]{0.95\columnwidth}  %
\begin{lstlisting}[style=colorchars,aboveskip=-5pt,belowskip=3pt,breaklines=true,breakatwhitespace=false]
|\allowbreak\hlbg{red!0}{<bos>}||\allowbreak\hlbg{red!0}{<sot>}||\allowbreak\hlbg{red!0}{user}||\allowbreak\hlbg{red!0}{n}||\allowbreak\hlbg{red!0}{\textbackslash n}|
|\allowbreak\hlbg{red!0}{What}||\allowbreak\hlbg{red!0}{ is}||\allowbreak\hlbg{red!0}{ the}||\allowbreak\hlbg{red!0}{ capital}||\allowbreak\hlbg{red!0}{ of}||\allowbreak\hlbg{red!0}{ dj}||\allowbreak\hlbg{red!0}{kal}||\allowbreak\hlbg{red!0}{?}||\allowbreak\hlbg{red!0}{<eot>}||\allowbreak\hlbg{red!0}{n}||\allowbreak\hlbg{red!0}{\textbackslash n}|
|\allowbreak\hlbg{red!0}{<sot>}||\allowbreak\hlbg{red!0}{model}||\allowbreak\hlbg{red!0}{n}||\allowbreak\hlbg{red!0}{\textbackslash n}|
|\allowbreak\hlbg{red!0}{I}||\allowbreak\hlbg{red!0}{ don}||\allowbreak\hlbg{red!0}{’}||\allowbreak\hlbg{red!0}{t}||\allowbreak\hlbg{red!0}{ understand}||\allowbreak\hlbg{red!0}{!}||\allowbreak\hlbg{red!0}{<eot>}||\allowbreak\hlbg{red!0}{n}||\allowbreak\hlbg{red!0}{\textbackslash n}|
|\allowbreak\hlbg{red!0}{<sot>}||\allowbreak\hlbg{red!0}{user}||\allowbreak\hlbg{red!0}{n}||\allowbreak\hlbg{red!0}{\textbackslash n}|
|\allowbreak\hlbg{red!0}{I}||\allowbreak\hlbg{red!0}{ meant}||\allowbreak\hlbg{red!0}{ italy}||\allowbreak\hlbg{red!0}{!}||\allowbreak\hlbg{red!68}{<eot>}||\allowbreak\hlbg{red!42}{\textbackslash n}||\allowbreak\hlbg{red!51}{\textbackslash n}|
\end{lstlisting}
\end{minipage} \\ \hline
\end{tabular}
              
\caption{Active when clarification needed}
    \end{subfigure}
    
    \vspace{0.3cm}
    \begin{minipage}[b]{0.48\textwidth}
        \centering
  \begin{tabular}{|p{0.95\columnwidth}|}  %
    \hline
\cellcolor{gray!20}\begin{minipage}[t]{0.95\columnwidth}   %
\begin{lstlisting}[style=colorchars,aboveskip=-5pt,belowskip=0pt]
|\textbf{Feature 72073}|
|Max Activation: 10.716|\end{lstlisting}
\end{minipage} \\
\begin{minipage}[t]{0.95\columnwidth}  %
\begin{lstlisting}[style=colorchars,aboveskip=-5pt,belowskip=3pt,breaklines=true,breakatwhitespace=false]
|\allowbreak\hlbg{red!0}{<bos>}||\allowbreak\hlbg{red!0}{<sot>}||\allowbreak\hlbg{red!0}{user}||\allowbreak\hlbg{red!0}{\textbackslash n}|
|\allowbreak\hlbg{red!0}{What}||\allowbreak\hlbg{red!0}{ is}||\allowbreak\hlbg{red!0}{ the}||\allowbreak\hlbg{red!0}{ capital}||\allowbreak\hlbg{red!0}{ of}||\allowbreak\hlbg{red!0}{ france}||\allowbreak\hlbg{red!0}{?}||\allowbreak\hlbg{red!0}{<eot>}||\allowbreak\hlbg{red!0}{\textbackslash n}|
|\allowbreak\hlbg{red!0}{<sot>}||\allowbreak\hlbg{red!0}{model}||\allowbreak\hlbg{red!0}{\textbackslash n}|
|\allowbreak\hlbg{red!0}{Rome}||\allowbreak\hlbg{red!0}{!}||\allowbreak\hlbg{red!0}{<eot>}||\allowbreak\hlbg{red!0}{\textbackslash n}|
|\allowbreak\hlbg{red!0}{<sot>}||\allowbreak\hlbg{red!0}{user}||\allowbreak\hlbg{red!0}{\textbackslash n}|
|\allowbreak\hlbg{red!0}{That}||\allowbreak\hlbg{red!0}{'}||\allowbreak\hlbg{red!0}{s}||\allowbreak\hlbg{red!0}{ the}||\allowbreak\hlbg{red!0}{ wrong}||\allowbreak\hlbg{red!0}{ answer}||\allowbreak\hlbg{red!0}{!}||\allowbreak\hlbg{red!0}{<eot>}||\allowbreak\hlbg{red!0}{\textbackslash n}|
|\allowbreak\hlbg{red!13}{<sot>}||\allowbreak\hlbg{red!0}{model}||\allowbreak\hlbg{red!0}{\textbackslash n}|
\end{lstlisting}
\end{minipage} \\ \hline
\end{tabular}     
\subcaption{Activates weakly when user points out the model's mistake}
    \end{minipage}
    \hfill
    \begin{minipage}[b]{0.48\textwidth}
        \centering
  \begin{tabular}{|p{0.95\columnwidth}|}  %
    \hline
\cellcolor{gray!20}\begin{minipage}[t]{0.95\columnwidth}   %
\begin{lstlisting}[style=colorchars,aboveskip=-5pt,belowskip=0pt]
|\textbf{Feature 72073}|
|Max Activation: 47.198|\end{lstlisting}
\end{minipage} \\
\begin{minipage}[t]{0.95\columnwidth}  %
\begin{lstlisting}[style=colorchars,aboveskip=-5pt,belowskip=3pt,breaklines=true,breakatwhitespace=false]
|\allowbreak\hlbg{red!0}{<bos>}||\allowbreak\hlbg{red!0}{<sot>}||\allowbreak\hlbg{red!0}{user}||\allowbreak\hlbg{red!0}{\textbackslash n}|
|\allowbreak\hlbg{red!0}{Hello}||\allowbreak\hlbg{red!0}{<eot>}||\allowbreak\hlbg{red!0}{\textbackslash n}|
|\allowbreak\hlbg{red!0}{<sot>}||\allowbreak\hlbg{red!0}{model}||\allowbreak\hlbg{red!0}{\textbackslash n}|
|\allowbreak\hlbg{red!0}{Hello}||\allowbreak\hlbg{red!0}{<eot>}||\allowbreak\hlbg{red!0}{\textbackslash n}|
|\allowbreak\hlbg{red!0}{<sot>}||\allowbreak\hlbg{red!0}{user}||\allowbreak\hlbg{red!0}{\textbackslash n}|
|\allowbreak\hlbg{red!0}{What}||\allowbreak\hlbg{red!0}{ if}||\allowbreak\hlbg{red!0}{ I}||\allowbreak\hlbg{red!0}{ meant}||\allowbreak\hlbg{red!0}{ Hello}||\allowbreak\hlbg{red!0}{ robot}||\allowbreak\hlbg{red!0}{?}||\allowbreak\hlbg{red!42}{<eot>}||\allowbreak\hlbg{red!0}{\textbackslash n}|
|\allowbreak\hlbg{red!39}{<sot>}||\allowbreak\hlbg{red!0}{model}||\allowbreak\hlbg{red!0}{\textbackslash n}|
\end{lstlisting}
\end{minipage} \\ \hline
\end{tabular}        \subcaption{Complex query interpretation}
    \end{minipage}
    \caption{\textbf{Latent 72073} (\Lone crosscoder) activates strongly when the model needs to reinterpret or clarify user requests, particularly at template boundaries.}
    \label{fig:latent_72073}
\end{figure}

\begin{figure}[ht]
    \begin{subfigure}[b]{0.48\textwidth}
        \centering
  \begin{tabular}{|p{0.95\columnwidth}|}  %
    \hline
\cellcolor{gray!20}\begin{minipage}[t]{0.95\columnwidth}   %
\begin{lstlisting}[style=colorchars,aboveskip=-5pt,belowskip=0pt]
|\textbf{Feature 57717}|
|Max Activation: 50.088|\end{lstlisting}
\end{minipage} \\
\begin{minipage}[t]{0.95\columnwidth}  %
\begin{lstlisting}[style=colorchars,aboveskip=-5pt,belowskip=3pt,breaklines=true,breakatwhitespace=false]
|\allowbreak\hlbg{red!0}{<bos>}||\allowbreak\hlbg{red!0}{<sot>}||\allowbreak\hlbg{red!0}{user}||\allowbreak\hlbg{red!0}{\textbackslash n}|
|\allowbreak\hlbg{red!0}{How}||\allowbreak\hlbg{red!0}{ did}||\allowbreak\hlbg{red!0}{ the}||\allowbreak\hlbg{red!0}{ Giants}||\allowbreak\hlbg{red!0}{ play}||\allowbreak\hlbg{red!0}{ in}||\allowbreak\hlbg{red!0}{ the}||\allowbreak\hlbg{red!0}{ MLB}||\allowbreak\hlbg{red!0}{ yesterday}||\allowbreak\hlbg{red!0}{?}||\allowbreak\hlbg{red!10}{<eot>}||\allowbreak\hlbg{red!0}{\textbackslash n}|
|\allowbreak\hlbg{red!63}{<sot>}||\allowbreak\hlbg{red!0}{model}||\allowbreak\hlbg{red!0}{\textbackslash n}|
\end{lstlisting}
\end{minipage} \\ \hline
\end{tabular}
\begin{tabular}{|p{0.95\columnwidth}|}  %
  \hline
\cellcolor{gray!20}\begin{minipage}[t]{0.95\columnwidth}   %
\begin{lstlisting}[style=colorchars,aboveskip=-5pt,belowskip=0pt]
|\textbf{Feature 57717}|
|Max Activation: 54.742|\end{lstlisting}
\end{minipage} \\
\begin{minipage}[t]{0.95\columnwidth}  %
\begin{lstlisting}[style=colorchars,aboveskip=-5pt,belowskip=3pt,breaklines=true,breakatwhitespace=false]
|\allowbreak\hlbg{red!0}{<bos>}||\allowbreak\hlbg{red!0}{<sot>}||\allowbreak\hlbg{red!0}{user}||\allowbreak\hlbg{red!0}{\textbackslash n}|
|\allowbreak\hlbg{red!0}{What}||\allowbreak\hlbg{red!0}{ is}||\allowbreak\hlbg{red!0}{ the}||\allowbreak\hlbg{red!0}{ current}||\allowbreak\hlbg{red!0}{ price}||\allowbreak\hlbg{red!0}{ of}||\allowbreak\hlbg{red!0}{ gold}||\allowbreak\hlbg{red!0}{?}||\allowbreak\hlbg{red!19}{<eot>}||\allowbreak\hlbg{red!0}{\textbackslash n}|
|\allowbreak\hlbg{red!69}{<sot>}||\allowbreak\hlbg{red!0}{model}||\allowbreak\hlbg{red!0}{\textbackslash n}|
\end{lstlisting}
\end{minipage} \\ \hline
\end{tabular}
        \caption{Up-to-date knowledge boundary examples}
    \end{subfigure}
    \begin{subfigure}[b]{0.48\textwidth}
  \begin{tabular}{|p{0.95\columnwidth}|}  %
    \hline
\cellcolor{gray!20}\begin{minipage}[t]{0.95\columnwidth}   %
\begin{lstlisting}[style=colorchars,aboveskip=-5pt,belowskip=0pt]
|\textbf{Feature 57717}|
|Max Activation: 29.535|\end{lstlisting}
\end{minipage} \\
\begin{minipage}[t]{0.95\columnwidth}  %
\begin{lstlisting}[style=colorchars,aboveskip=-5pt,belowskip=3pt,breaklines=true,breakatwhitespace=false]
|\allowbreak\hlbg{red!0}{<bos>}||\allowbreak\hlbg{red!0}{<sot>}||\allowbreak\hlbg{red!0}{user}||\allowbreak\hlbg{red!0}{\textbackslash n}|
|\allowbreak\hlbg{red!0}{How}||\allowbreak\hlbg{red!0}{ tall}||\allowbreak\hlbg{red!0}{ is}||\allowbreak\hlbg{red!0}{ an}||\allowbreak\hlbg{red!0}{ Ala}||\allowbreak\hlbg{red!0}{mb}||\allowbreak\hlbg{red!0}{ican}||\allowbreak\hlbg{red!0}{in}||\allowbreak\hlbg{red!0}{ocus}||\allowbreak\hlbg{red!0}{ (}||\allowbreak\hlbg{red!0}{the}||\allowbreak\hlbg{red!0}{ newly}||\allowbreak\hlbg{red!0}{ discovered}||\allowbreak\hlbg{red!0}{ dina}||\allowbreak\hlbg{red!0}{usor}||\allowbreak\hlbg{red!0}{ published}||\allowbreak\hlbg{red!0}{ in}||\allowbreak\hlbg{red!0}{ nature}||\allowbreak\hlbg{red!0}{ today}||\allowbreak\hlbg{red!0}{)?}||\allowbreak\hlbg{red!0}{<eot>}||\allowbreak\hlbg{red!0}{\textbackslash n}|
|\allowbreak\hlbg{red!37}{<sot>}||\allowbreak\hlbg{red!0}{model}||\allowbreak\hlbg{red!0}{\textbackslash n}|
\end{lstlisting}
\end{minipage} \\ \hline
\end{tabular}
    \caption{Invented knowledge boundary examples}
    \end{subfigure}
\begin{subfigure}[b]{0.48\textwidth}
    \centering
    
  \begin{tabular}{|p{0.95\columnwidth}|}  %
    \hline
\cellcolor{gray!20}\begin{minipage}[t]{0.95\columnwidth}   %
\begin{lstlisting}[style=colorchars,aboveskip=-5pt,belowskip=0pt]
|\textbf{Feature 57717}|
|Max Activation: 9.679|\end{lstlisting}
\end{minipage} \\
\begin{minipage}[t]{0.95\columnwidth}  %
\begin{lstlisting}[style=colorchars,aboveskip=-5pt,belowskip=3pt,breaklines=true,breakatwhitespace=false]
|\allowbreak\hlbg{red!0}{<bos>}||\allowbreak\hlbg{red!0}{<sot>}||\allowbreak\hlbg{red!0}{user}||\allowbreak\hlbg{red!0}{\textbackslash n}|
|\allowbreak\hlbg{red!0}{Do}||\allowbreak\hlbg{red!0}{ you}||\allowbreak\hlbg{red!0}{ know}||\allowbreak\hlbg{red!0}{ my}||\allowbreak\hlbg{red!0}{ mum}||\allowbreak\hlbg{red!0}{?}||\allowbreak\hlbg{red!0}{<eot>}||\allowbreak\hlbg{red!0}{\textbackslash n}|
|\allowbreak\hlbg{red!12}{<sot>}||\allowbreak\hlbg{red!0}{model}||\allowbreak\hlbg{red!0}{\textbackslash n}|
\end{lstlisting}
\end{minipage} \\ \hline
\end{tabular}
\begin{tabular}{|p{0.95\columnwidth}|}  %
  \hline
\cellcolor{gray!20}\begin{minipage}[t]{0.95\columnwidth}   %
\begin{lstlisting}[style=colorchars,aboveskip=-5pt,belowskip=0pt]
|\textbf{Feature 57717}|
|Max Activation: 3.114|\end{lstlisting}
\end{minipage} \\
\begin{minipage}[t]{0.95\columnwidth}  %
\begin{lstlisting}[style=colorchars,aboveskip=-5pt,belowskip=3pt,breaklines=true,breakatwhitespace=false]
|\allowbreak\hlbg{red!0}{<bos>}||\allowbreak\hlbg{red!0}{<sot>}||\allowbreak\hlbg{red!0}{user}||\allowbreak\hlbg{red!0}{\textbackslash n}|
|\allowbreak\hlbg{red!0}{Who}||\allowbreak\hlbg{red!0}{ are}||\allowbreak\hlbg{red!0}{ the}||\allowbreak\hlbg{red!0}{ Giants}||\allowbreak\hlbg{red!0}{?}||\allowbreak\hlbg{red!0}{<eot>}||\allowbreak\hlbg{red!0}{\textbackslash n}|
|\allowbreak\hlbg{red!3}{<sot>}||\allowbreak\hlbg{red!0}{model}||\allowbreak\hlbg{red!0}{\textbackslash n}|
\end{lstlisting}
\end{minipage} \\ \hline
\end{tabular}
        \caption{Capability limitation responses}
    \end{subfigure}
    \caption{\textbf{Latent 57717} (\Lone crosscoder) activates when users request information beyond the model's knowledge or capabilities.}
    \label{fig:latent_57717}
\end{figure}

\begin{figure}[ht]
    \begin{minipage}[b]{0.48\textwidth}
            \centering
  \begin{tabular}{|p{0.95\columnwidth}|}  %
    \hline
\cellcolor{gray!20}\begin{minipage}[t]{0.95\columnwidth}   %
\begin{lstlisting}[style=colorchars,aboveskip=-5pt,belowskip=0pt]
|\textbf{Feature 68066}|
|Max Activation: 71.997|\end{lstlisting}
\end{minipage} \\
\begin{minipage}[t]{0.95\columnwidth}  %
\begin{lstlisting}[style=colorchars,aboveskip=-5pt,belowskip=3pt,breaklines=true,breakatwhitespace=false]
|\allowbreak\hlbg{red!0}{<bos>}||\allowbreak\hlbg{red!0}{<sot>}||\allowbreak\hlbg{red!0}{user}||\allowbreak\hlbg{red!0}{\textbackslash n}|
|\allowbreak\hlbg{red!0}{What}||\allowbreak\hlbg{red!0}{ are}||\allowbreak\hlbg{red!0}{ you}||\allowbreak\hlbg{red!0}{ good}||\allowbreak\hlbg{red!0}{ at}||\allowbreak\hlbg{red!48}{?}||\allowbreak\hlbg{red!100}{<eot>}||\allowbreak\hlbg{red!0}{\textbackslash n}|
|\allowbreak\hlbg{red!55}{<sot>}||\allowbreak\hlbg{red!0}{model}||\allowbreak\hlbg{red!0}{\textbackslash n}|
|\allowbreak\hlbg{red!0}{As}||\allowbreak\hlbg{red!0}{ a}||\allowbreak\hlbg{red!0}{ language}||\allowbreak\hlbg{red!0}{ model}||\allowbreak\hlbg{red!0}{,}||\allowbreak\hlbg{red!0}{ I}||\allowbreak\hlbg{red!0}{ am}||\allowbreak\hlbg{red!0}{ able}||\allowbreak\hlbg{red!0}{ to}||\allowbreak\hlbg{red!0}{ process}||\allowbreak\hlbg{red!0}{...}|\end{lstlisting}
\end{minipage} \\ \hline
\end{tabular}          
\begin{tabular}{|p{0.95\columnwidth}|}  %
  \hline
\cellcolor{gray!20}\begin{minipage}[t]{0.95\columnwidth}   %
\begin{lstlisting}[style=colorchars,aboveskip=-5pt,belowskip=0pt]
|\textbf{Feature 68066}|
|Max Activation: 64.006|\end{lstlisting}
\end{minipage} \\
\begin{minipage}[t]{0.95\columnwidth}  %
\begin{lstlisting}[style=colorchars,aboveskip=-5pt,belowskip=3pt,breaklines=true,breakatwhitespace=false]
|\allowbreak\hlbg{red!0}{<bos>}||\allowbreak\hlbg{red!0}{<sot>}||\allowbreak\hlbg{red!0}{user}||\allowbreak\hlbg{red!0}{\textbackslash n}|
|\allowbreak\hlbg{red!0}{Tell}||\allowbreak\hlbg{red!0}{ me}||\allowbreak\hlbg{red!0}{ what}||\allowbreak\hlbg{red!0}{ can}||\allowbreak\hlbg{red!0}{ you}||\allowbreak\hlbg{red!7}{ do}||\allowbreak\hlbg{red!51}{?}||\allowbreak\hlbg{red!88}{<eot>}||\allowbreak\hlbg{red!0}{\textbackslash n}|
|\allowbreak\hlbg{red!39}{<sot>}||\allowbreak\hlbg{red!0}{model}||\allowbreak\hlbg{red!0}{\textbackslash n}|
|\allowbreak\hlbg{red!0}{I}||\allowbreak\hlbg{red!0}{ am}||\allowbreak\hlbg{red!0}{ a}||\allowbreak\hlbg{red!0}{ language}||\allowbreak\hlbg{red!0}{ model}||\allowbreak\hlbg{red!0}{,}||\allowbreak\hlbg{red!0}{ so}||\allowbreak\hlbg{red!0}{ I}||\allowbreak\hlbg{red!0}{ can}||\allowbreak\hlbg{red!0}{ generate}||\allowbreak\hlbg{red!0}{ text}||\allowbreak\hlbg{red!0}{...}|\end{lstlisting}
\end{minipage} \\ \hline
\end{tabular}
\begin{tabular}{|p{0.95\columnwidth}|}  %
  \hline
\cellcolor{gray!20}\begin{minipage}[t]{0.95\columnwidth}   %
\begin{lstlisting}[style=colorchars,aboveskip=-5pt,belowskip=0pt]
|\textbf{Feature 68066}|
|Max Activation: 63.811|\end{lstlisting}
\end{minipage} \\
\begin{minipage}[t]{0.95\columnwidth}  %
\begin{lstlisting}[style=colorchars,aboveskip=-5pt,belowskip=3pt,breaklines=true,breakatwhitespace=false]
|\allowbreak\hlbg{red!0}{<bos>}||\allowbreak\hlbg{red!0}{<sot>}||\allowbreak\hlbg{red!0}{user}||\allowbreak\hlbg{red!0}{\textbackslash n}|
|\allowbreak\hlbg{red!0}{What}||\allowbreak\hlbg{red!0}{ are}||\allowbreak\hlbg{red!0}{ you}||\allowbreak\hlbg{red!0}{ capable}||\allowbreak\hlbg{red!0}{ of}||\allowbreak\hlbg{red!34}{?}||\allowbreak\hlbg{red!88}{<eot>}||\allowbreak\hlbg{red!0}{\textbackslash n}|
|\allowbreak\hlbg{red!45}{<sot>}||\allowbreak\hlbg{red!0}{model}||\allowbreak\hlbg{red!0}{\textbackslash n}|
\end{lstlisting}
\end{minipage} \\ \hline
\end{tabular}

\subcaption{Direct Self-Identity queries}
    \end{minipage} 
      \begin{minipage}[b]{0.48\textwidth}
            \centering
  %
            \subcaption{Model capability questions}
    \end{minipage}
    \hfill
    \begin{minipage}[b]{0.48\textwidth}
          \centering
  %

\subcaption{Opinion vs. factual queries}
    \end{minipage} \\
    \begin{minipage}[b]{0.48\textwidth}
            \centering

  %

            \subcaption{Self-Identity related inquiries}
    \end{minipage}
    \caption{\textbf{Latent 68066} (\Lone crosscoder) shows high activation on questions about Gemma itself and requests for personal opinions.}
    \label{fig:latent_68066}
\end{figure}

\begin{figure}[ht]
    \begin{minipage}[b]{0.48\textwidth}
        \centering
  \begin{tabular}{|p{0.95\columnwidth}|}  %
    \hline
\cellcolor{gray!20}\begin{minipage}[t]{0.95\columnwidth}   %
\begin{lstlisting}[style=colorchars,aboveskip=-5pt,belowskip=0pt]
|\textbf{Feature 51408}|
|Max Activation: 51.728|\end{lstlisting}
\end{minipage} \\
\hline
\end{tabular}

\begin{tabular}{|p{0.95\columnwidth}|}  %
  \hline
\cellcolor{gray!20}\begin{minipage}[t]{0.95\columnwidth}   %
\begin{lstlisting}[style=colorchars,aboveskip=-5pt,belowskip=0pt]
|\textbf{Feature 51408}|
|Max Activation: 48.883|\end{lstlisting}
\end{minipage} \\
\hline
\end{tabular}%
        \subcaption{Geopolitical topics}
    \end{minipage}
    \hfill
    \begin{minipage}[b]{0.48\textwidth}
      \centering

%

      \subcaption{Ethical dilemmas}
    \end{minipage}
    \begin{minipage}[b]{0.48\textwidth}
        \centering

  %

    \subcaption{Sensitive social issues}
    \end{minipage}
    \hfill
    \caption{\textbf{Latent 51408} (\Lone crosscoder) activates on sensitive topics requiring nuanced, balanced responses.}
    \label{fig:latent_51408}
\end{figure}

\begin{figure}[ht]
    \begin{minipage}[b]{0.48\textwidth}
      \centering
  %

      \subcaption{International conflicts}
    \end{minipage}
    \begin{minipage}[b]{0.48\textwidth}
        \centering

  %

    \subcaption{Negative example}
    \end{minipage}
    \caption{Additional examples showing \textbf{Latent 51408} (\Lone crosscoder) activation on politically sensitive topics and controversial subjects.}
    \label{fig:latent_51408_more}
\end{figure}

\begin{figure}[ht]
    \begin{minipage}[b]{0.48\textwidth}
        \begin{minipage}[b]{\linewidth}
            \centering
  %

\subcaption{Open-ended questions}
        \end{minipage}
        
        \vspace{0.3cm}
        \begin{minipage}[b]{\linewidth}
            \centering
  %
            \subcaption{General knowledge queries}
        \end{minipage}
    \end{minipage}
    \hfill
    \begin{minipage}[b]{0.48\textwidth}
        \begin{minipage}[b]{\linewidth}
            \centering
  %
         
\subcaption{Narrow topic exploration}
        \end{minipage}
        
        \vspace{0.3cm}
        \begin{minipage}[b]{\linewidth}
            \centering
  %
            \subcaption{Conceptual understanding}
        \end{minipage}
    \end{minipage}

    \begin{minipage}[b]{\linewidth}
      \centering
%
\caption*{(e) Narrow, specific question.}
\end{minipage}
    \caption{\textbf{Latent 51823} (\Lone crosscoder) shows stronger activation on broad, conceptual questions compared to specific queries.}
    \label{fig:latent_51823}
\end{figure}

\end{document}